\documentclass{article}
\pdfoutput=1 

\usepackage{geometry}
\usepackage[pdftex]{graphicx}

\usepackage{amssymb}
\usepackage{amsmath}

\newtheorem{corollary}{Corollary}
\newtheorem{lemma}{Lemma}
\newtheorem{proposition}[lemma]{Proposition}

\newtheorem{theorem}{Theorem}

\newenvironment{proof}{\par\noindent{\em Proof. }}{$\Box$\\[2mm]}

\newcommand{\lebesguevol}{\mathcal{L}}

\newcommand{\indicator}{1}

\newcommand{\capintegral}{F_C}
\newcommand{\ballintegral}{F_B}

\newcommand{\R}{\mathbb{R}}
\newcommand{\N}{\mathbb{N}}
\newcommand{\E}{\mathbb{E}}

\newcommand{\eps}{\ensuremath{\varepsilon}}
\newcommand{\norm}[1]{\|#1\|}
\DeclareMathOperator{\Bin}{Bin}
\DeclareMathOperator{\Var}{Var}

\newcommand{\cutlim}{CutLim}
\newcommand{\vollim}{VolLim}
\newcommand{\ncutlim}{NCutLim}
\newcommand{\cheegercutlim}{CheegerCutLim}
\newcommand{\xcutlim}{XCutLim}

\DeclareMathOperator{\vol}{vol}
\DeclareMathOperator{\kNN}{kNN}
\DeclareMathOperator{\cheegercut}{CheegerCut}
\DeclareMathOperator{\ncut}{NCut}
\DeclareMathOperator{\xcut}{XCut}
\DeclareMathOperator{\cut}{cut}
\DeclareMathOperator{\dist}{dist}

\newcommand{\cutrate}{s^{\cut}}
\newcommand{\volrate}{s^{\vol}}

\makeatletter
\let\NAT@parse\undefined
\makeatother

\usepackage[round,comma]{natbib}

\begin{document}

\title{How the result of graph clustering methods depends on the construction of the
  graph}

\author{Markus~Maier\\
Max Planck Institute for Biological Cybernetics,
  T\"{u}bingen, Germany \\
  \texttt{mmaier@tuebingen.mpg.de} \\
\and
  Ulrike von~Luxburg (Corresponding author)\\
  Max Planck Institute for Biological Cybernetics,
  T\"{u}bingen, Germany \\
\texttt{ulrike.luxburg@tuebingen.mpg.de}\\
\and Matthias Hein\\ 
Saarland University, Saarbr\"{u}cken, Germany\\
 \texttt{hein@cs.uni-sb.de}
}

\date{\today}

\maketitle

  \begin{abstract}
We study the scenario of graph-based clustering algorithms such as
spectral clustering. Given a set of data points, one first has to
construct a graph on the data points and then apply a graph clustering
algorithm to find a suitable partition of the graph. Our main question
is if and how the construction of the graph (choice of the graph,
choice of parameters, choice of weights) influences the outcome of the
final clustering result. To this end we study the convergence of
cluster quality measures such as the normalized cut or the Cheeger cut
on various kinds of random geometric graphs as the sample size tends
to infinity. It turns out that the limit values of the same objective
function are systematically different on different types of
graphs. This implies that clustering results systematically depend on
the graph and can be very different for different types of graph.  We
provide examples to illustrate the implications on spectral
clustering. 
 \end{abstract}
  
  \section{Introduction}

  Nowadays it is very popular to represent and analyze statistical
  data using random graph or network models. The vertices in such a graph
  correspond to data points, whereas edges in the graph indicate that the
  adjacent vertices are ``similar'' or ``related'' to each other. 
  In this paper we consider the problem of data clustering in a random
  geometric graph setting. We are given a sample of points drawn
  from some underlying probability distribution on a metric
  space. The goal is to cluster the sample points into ``meaningful
  groups''.  A standard procedure is to first transform the data to
  a neighborhood graph, for example a $k$-nearest neighbor graph. In a second step, the cluster structure is then 
  extracted from the graph: clusters correspond to regions in the
  graph that are tightly connected within themselves and only sparsely
  connected to other clusters.

There already exist  a couple of papers that study statistical
properties of this procedure in a particular setting: when the true 
underlying clusters are defined to be the connected components of
a density level set in the underlying space. In his setting, a test for
detecting cluster structure and outliers is proposed in
\citet{BriChaQuiYuk97}. In \citet{BiaCadPel07} the authors build
a neighborhood graph in such a way that its connected components
converge to the underlying true clusters in the data. 
\citet{MaiHeiLux09} compare the properties of different
random graph models for identifying clusters of the density level
sets. 

While the definition of clusters as connected components of level
sets is appealing from a theoretical point of view, the corresponding
algorithms are often too simplistic and only moderately successful in
practice. %
From a practical point of view, clustering methods based on graph
partitioning algorithms are more robust. Clusters do not have to be
perfectly disconnected in the graph, but are allowed to have a small
number of connecting edges between them. Graph partitioning methods are widely used in
practice. The most prominent algorithm in this class is spectral
clustering, which optimizes the normalized cut ($\ncut$) objective
function (see below for exact definitions, and \citet{Luxburg07} for
a tutorial on spectral clustering). It is already known under what
circumstances spectral clustering is statistically consistent
\citep{LuxBelBou08}.  However, there is one important open
question. When applying graph-based methods to given sets of data
points, one obviously has to build a graph first, and there are
several important choices to be made: the type of the graph (for
example, $k$-nearest neighbor graph, the $r$-neighborhood graph or a
Gaussian similarity graph), the
connectivity parameter ($k$ or $r$ or $\sigma$, respectively) and the weights of
the graph.  Making such choices is not so difficult in the domain of
supervised learning, where parameters can be set using
cross-validation. However, it poses a serious problem in unsupervised
learning.  While different researchers use different heuristics and
their ``gut feeling'' to set these parameters, neither systematic
empirical studies have been conducted (for example, how sensitive 
the results are to the choice of graph parameters), nor do theoretical results
exist which lead to well-justified heuristics.

In this paper we study the question if and how the results of graph-based clustering algorithms  
  are affected by the graph type and the parameters that are chosen for 
  the construction of the neighborhood graph. We focus on the case
  where the best clustering is defined as the partition that minimizes the normalized
  cut (Ncut) or the Cheeger cut.  

Our theoretical setup is as follows.   %
In a first step we ignore the problem of actually {\em finding} the optimal
partition.  Instead we fix some partition of the underlying space
and consider it as the ``true'' partition. For any
finite set of points drawn from the underlying space we consider the
clustering of the points
  that is induced by this underlying partition. Then we
  study the convergence of the $\ncut$ value of this clustering as the sample size tends to
  infinity. We investigate this question on different kinds of
  neighborhood graphs. Our first main result is that 
  depending on the type of graph, the clustering quality measure converges to
  different limit values.   For example, depending on whether we use
  the $\kNN$ graph or the $r$-graph, the limit functional integrates over
  different powers of the density.  %
From a statistical point of view, this is very surprising because in
many other 
respects, the $\kNN$ graph and the $r$-graph behave very similar to
each other. Just consider
the related problem of density estimation. Here, both the $k$-nearest
neighbor density estimate and the estimate based on the degrees in the
$r$-graph converge to the same limit, namely the true underlying
density. So it is far from obvious that the $\ncut$ values would
converge to different limits. 

  In a second step we then relate these results to the setting where
  we optimize over all partitions to find the one that minimizes the
  $\ncut$. We can show that the results from the first part can lead
  to the effect that the minimizer of $\ncut$ on the $\kNN$ graph is
  different from the minimizer of $\ncut$ on the $r$-graph or on the
  complete graph with Gaussian weights. This effect can also be
  studied in practical examples.  First, we give examples of
  well-clustered distributions (mixtures of Gaussians) where the
  optimal limit cut on the $\kNN$ graph is different from the one on
  the $r$-neighborhood graph. The optimal limit cuts in these examples
  can be computed analytically. Next we can demonstrate that this
  effect can already been observed on finite samples from these
  distributions. Given a finite sample, running normalized spectral
  clustering to optimize Ncut leads to systematically different
  results on the $\kNN$ graph than on the $r$-graph. This shows
  that our results are not only of theoretical interest, but that they
  are highly relevant in practice.

  In the following section we formally define the graph clustering quality measures and
  the neighborhood graph types we consider in this paper. Furthermore, we introduce
  the notation and technical assumptions for the rest of the paper.
  In Section~\ref{sec:limits_quality_measures} we present our main
  results on the convergence of  $\ncut$ and the $\cheegercut$ on
  different graphs. 
  In Section~\ref{sec:experiments}  we show that our findings are not
  only of theoretical interest, but that they also influence concrete
  algorithms such as spectral clustering in practice.  All proofs
  are deferred to Section~\ref{sec:proofs}. 
  Note that a small part of the results of this paper has already
  been published in~\citet{OurNipsPaper:2009}.

  \section{Definitions and assumptions} 
  \label{sec:definitions_assumptions_limit_ncut}
  Given a directed graph $G=(V,E)$ with weights $w:E \to \R$ and a partition
  of the nodes $V$ into $(U, V \setminus U)$ we define
  \begin{align*}
    \cut (U, V \setminus U)
    = \sum_{u \in U, v \in V \setminus U} \left( w ( u, v )+ w( v, u ) \right) , 
  \end{align*}  
  and $\vol ( U ) =  \sum_{u \in U, v \in V} w ( u,v )$. 
  If $G$ is an undirected graph we replace the ordered pair $( u, v )$ in the sums
  by the unordered pair $\{u,v\}$. Note that by doing so we count each edge twice 
  in the undirected graph. This introduces a constant of two in the limits but it has
  the advantage that there is no need to distinguish in the formulation of our results 
  between directed and undirected graphs.
  
  Intuitively, the $\cut$ measures how strong the connection between the different 
  clusters in the clustering is, whereas the volume of a subset of the nodes 
  measures the ``weight'' of the subset in terms of the edges that originate in it.
  An ideal clustering would have a low $\cut$ and balanced clusters, that is 
  clusters with similar volume. 
  The graph clustering quality measures that we use in this paper, the normalized cut
  and the Cheeger cut, formalize this trade-off in slightly different ways:
  The normalized cut is defined by
  \begin{align} \label{eq:def_ncut}
    &\ncut ( U, V \setminus U )
    = \cut ( U, V \setminus U ) \bigg( \frac{1}{\vol (U)} +
    \frac{1}{\vol (V \setminus U)} \bigg),
    \intertext{whereas the Cheeger cut is defined by}
    &\cheegercut ( U, V \setminus U )
    = \frac{\cut ( U, V \setminus U )}{\min \{\vol (U), \vol (V \setminus U)\}} .
    \label{eq:def_cheegercut}
  \end{align}
   
  These definitions are useful for general weighted graphs and general partitions. 
  As was said in the beginning we want to study the values of $\ncut$ and 
  $\cheegercut$ on neighborhood graphs on sample points in Euclidean space and
  for partitions of the nodes that are induced by a hyperplane $S$ in $\R^d$. 
  The two halfspaces belonging to $S$ are denoted by $H^+$ and $H^-$.
  Having a neighborhood graph on the sample points $\{x_1, \ldots ,x_n\}$,
  the partition of the nodes induced by $S$ is 
  $(\{x_1, \ldots ,x_n\} \cap H^+,\{x_1, \ldots ,x_n\} \cap H^-)$.
  In the rest of this paper for a given neighborhood graph $G_n$ we set
  $\cut_n = \cut (\{x_1, \ldots ,x_n\} \cap H^+,\{x_1, \ldots ,x_n\} \cap H^-)$.
  Similarly, for $H=H^+$ or $H=H^-$ we set 
  $\vol_n(H) = \vol ( \{x_1, \ldots ,x_n\} \cap H^+)$.
  Accordingly we define $\ncut_n$ and $\cheegercut_{n}$.

  In the following we introduce the different types of neighborhood graphs and 
  weighting schemes that are considered in this paper. 
  The graph types are:
  \begin{itemize}
    \item
      {\em The $k$-nearest neighbor ($\kNN$) graphs}, 
      where the idea is to connect each point to its $k$ nearest 
      neighbors. However, this yields a directed graph, since the $k$-nearest neighbor 
      relationship is not symmetric. If we want to construct an undirected $\kNN$ graph we
      can choose between the mutual $\kNN$ graph, where there is an edge between two points
      if both points are among the $k$ nearest neighbors of the other one, and the 
      symmetric $\kNN$ graph, where there is an edge between two points if only one point
      is among the $k$ nearest neighbors of the other one.
      In our proofs for the limit expressions it will become clear that these do not differ
      between the different types of $\kNN$ graphs. Therefore, we do not distinguish between
      them in the statement of the theorems, but rather speak of ``the $\kNN$ graph''.
      
    \item
      {\em The $r$-neighborhood graph}, where a radius $r$ is fixed and two points are 
      connected if their distance does not exceed the threshold radius $r$. 
      Note that due to the symmetry of the distance we do not have to distinguish between
      directed and undirected graphs.
    \item
      {\em The complete weighted graph}, where there is an edge between each pair of 
      distinct nodes (but no loops). 
      Of course, in general we would not consider this graph a neighborhood graph.  
      However, if the weight function is chosen in such a way that the weights of edges
      between nearby nodes are high and the weights between points far away from each other
      are almost negligible, then the behavior of this graph should be similar to that
      of a neighborhood graph. One such weight function is the Gaussian weight function,
      which we introduce below.
  \end{itemize}

  The weights that are used on neighborhood graphs usually depend on the 
  distance of the end nodes of the edge and are non-increasing.
  That is, the weight $w ( x_i,x_j )$ of an edge $(x_i,x_j)$ is given by
  $w ( x_i,x_j )= f (\dist (x_i,x_j))$ with a non-increasing 
  weight function $f$.
  The weight functions we consider here are the {\em unit weight function} $f \equiv 1$,
  which results in the unweighted graph, and the {\em Gaussian weight function}
  \begin{align*}
    f(u) = \frac{1}{(2 \pi \sigma^2)^{d/2}} \exp \left( - \frac{1}{2} \frac{u^2}{\sigma^2} \right)
  \end{align*}
  with a parameter $\sigma>0$ defining the bandwidth. 

  Of course, not every weighting scheme is suitable for every graph type. For example, 
  as mentioned above, we would hardly consider the complete graph with unit weights 
  a neighborhood graph. Therefore, we only consider the Gaussian weight function for 
  this graph.
  On the other hand, for the $\kNN$ graph and the $r$-neighborhood graph with Gaussian weights
  there are two ``mechanisms'' that reduce the influence of far-away nodes:
  first the fact that far-away nodes are not connected to each other by an edge and 
  second the decay of the weight function. In fact, it turns out that the limit expressions 
  we study depend on the interplay between these two mechanisms. 
  Clearly, the decay of the weight function is governed by the parameter $\sigma$.
  For the $r$-neighborhood graph the radius $r$ limits the length of the edges.
  Asymptotically, given sequences $(\sigma_n)_{n \in \N}$ and $(r_n)_{n \in \N}$ of bandwidths and radii
  we distinguish between the following two cases:  
  \begin{itemize}
  \item
    {\em the bandwidth $\sigma_n$ is dominated by the radius $r_n$}, 
    that is $\sigma_n / r_n \to 0$ for $n \to \infty$,
  \item
    {\em the radius $r_n$ is dominated by the bandwidth $\sigma_n$}, 
    that is $r_n / \sigma_n \to 0$ for $n \to \infty$.
  \end{itemize}
  For the $\kNN$ graph we cannot give a radius up to which points are 
  connected by an edge, since this radius for each point is a random variable that depends on the
  positions of all the sample points. However, it is possible to show that for a point
  in a region of constant density $p$ the $k_n$-nearest neighbor radius is concentrated 
  around $\sqrt[d]{k_n/((n-1) \eta_d p)}$, where $\eta_d$ denotes the volume of the
  unit ball in Euclidean space $\R^d$.  
  That is, the $\kNN$ radius decays to zero with the rate $\sqrt[d]{k_n/n}$.
  In the following it is convenient to set for the $\kNN$ graph $r_n=\sqrt[d]{k_n/n}$, 
  noting that this is not 
  the $k$-nearest neighbor radius of any point but only its decay rate.
  Using this ``radius'' we distinguish between the same two cases of the ratio of
  $r_n$ and $\sigma_n$ as for the $r$-neighborhood graph.

  For the sequences $(r_n)_{n \in \N}$ and $(\sigma_n)_{n \in \N}$ we always assume 
  $r_n \to 0$, $\sigma_n \to 0$  and $n r_n \to \infty$, 
  $n \sigma_n \to \infty$ for $n \to \infty$.
  Furthermore, for the parameter sequence $(k_n)_{n \in \N}$ of the $\kNN$ graph
  we always assume $k_n/n \to 0$, which corresponds to $r_n \to 0$, and 
  $k_n/\log n \to \infty$.

  In the rest of this paper we denote by $\lebesguevol_d$ the Lebesgue measure 
  in $\R^d$.
  Furthermore, let $B(x,r)$ denote the closed ball of radius $r$ around $x$ and  
  $\eta_d = \lebesguevol_d (B(0,1))$, where we set $\eta_0=1$.
  
  We make the following {\bf general assumptions in the whole paper: } 
  {\em 
    \begin{itemize}
    \item
      The data points $x_1, ..., x_n$ are drawn independently from some 
      density $p$ on $\R^d$. The measure on $\R^d$ that is induced by $p$ 
      is denoted by $\mu$; that means, for a measurable set $A \subseteq \R^d$ 
      we set $\mu (A) = \int_A p (x) \ \mathrm{d}x$.
    \item
      The density $p$ is bounded from below and above, that is 
      $0 < p_{\min} \leq p(x) \leq p_{\max}$. 
      In particular, it has compact support $C$.  
    \item
      In the interior of $C$, the density $p$ is twice differentiable
      and $\norm{\nabla p (x) } \leq p_{\max}'$ for a $p_{\max}' \in \R$
      and all $x$ in the interior of $C$. 
    \item
      The cut hyperplane $S$ splits the space $\R^d$ into two halfspaces
      $H^{+}$ and $H^{-}$ (both including the hyperplane $S$) with 
      positive probability masses, that is $\mu (H^+) > 0$, $\mu (H^-)>0$.
      The normal of $S$ pointing towards $H^+$ is denoted by $n_S$.     
    \item
      If $d \geq 2$ the boundary $\partial C$ is a compact, smooth $(d-1)$-dimensional surface
      with minimal curvature radius $\kappa>0$, that is the absolute values of the
      principal curvatures are bounded by $1/\kappa$.
      We denote by $n_x$ the normal to the surface $\partial C$ at the point 
      $x \in \partial C$.
      Furthermore, we can find constants $\gamma > 0$ and $r_{\gamma}>0$ 
      such that for all $r \leq r_{\gamma}$ we have
      $\lebesguevol_d (B(x,r) \cap C) \geq \gamma \lebesguevol_d (B(x,r))$ for all $x \in C$.
    \item
      If $d \geq 2$ we can find an angle $\alpha \in (0,\pi/2)$ such that 
      $| \langle n_S, n_x \rangle | \leq \cos \alpha$
      for all $x \in S \cap \partial C$.
      If $d=1$ we assume that (the point) $S$ is in the interior of $C$.
    \end{itemize}
  }
  The assumptions on the boundary $\partial C$ are necessary in order to bound
  the influence of points that are close to the boundary. The problem with these points
  is that the density is not approximately uniform inside small balls around them.
  Therefore, we cannot find a good estimate of their $\kNN$ radius and on their contribution
  to the $\cut$ and the volume. Under the assumptions above we can neglect these points.

  \section{Main results: Limits of the quality measures $\ncut$ and $\cheegercut$}  
  \label{sec:limits_quality_measures}
  As we can see in Equations~\eqref{eq:def_ncut} and~\eqref{eq:def_cheegercut} 
  the definitions of $\ncut$ and $\cheegercut$ rely on the $\cut$ and the volume.
  Therefore, in order to study the convergence of $\ncut$ and $\cheegercut$
  it seems reasonable to study the convergence of the $\cut$ and the volume first.
  In Section~\ref{sec:proofs} the Corollaries~\ref{cor:cut_knn_unweighted}-\ref{cor:cut_knn_sigma_to_r_to_zero}
  and the Corollaries~\ref{cor:vol_knn_unweighted}-\ref{cor:vol_knn_r_to_sigma_to_infinity}
  state the convergence of the $\cut$ and the volume on the $\kNN$ graphs.
  The Corollaries~\ref{cor:cut_unweighted}-\ref{cor:cut_r_graph_gaussian_weights_r_to_sigma_to_zero} state the convergence of the $\cut$ on the 
  $r$-graph and the complete weighted graph, whereas the 
  Corollaries~\ref{cor:volume_unweighted}-\ref{cor:volume_r_graph_r_to_sigma_to_zero}
  state the convergence of the volume on the same graphs.
 
  These corollaries show that there are scaling sequences 
  $(s_n^{\cut})_{n \in \N}$ and $(s_n^{\vol})_{n \in \N}$ 
  that depend on $n$, $r_n$ and the graph type 
  such that, under certain conditions, almost surely 
  \begin{align*}
    \left( s_n^{\cut} \right)^{-1} \cut_n \to \cutlim 
    \qquad \text{ and } \qquad \left( s_n^{\vol} \right)^{-1} \vol_n (H) \to \vollim (H) 
  \end{align*}
  for $n \to \infty$, where $\cutlim \in \R_{\geq 0}$ and 
  $\vollim (H^+),\vollim (H^-) \in \R_{>0}$ are constants depending only on 
  the density $p$ and the hyperplane $S$.

  \begin{table}
    \begin{tabular}{|l||c|c|c|}
      \hline
      \multicolumn{4}{|l|}{\bf The $\cut$ in the $\kNN$-graph and the $r$-graph}\\
      \hline
      Weighting & $s_n^{\cut}$ & $\cutlim$ $\kNN$-graph & $\cutlim$ $r$-graph \\
      \hline
      \hline
      unweighted & $n^{2} r_n^{d+1}$ & $\frac{2 \eta_{d-1}}{(d+1) \eta_d^{1+1/d}} \int_S p^{1-1/d} (s) \ \mathrm{d} s$ & $\frac{2 \eta_{d-1}}{d+1} \int_S p^2 (s) \ \mathrm{d} s$\\
      \hline
      weighted $r_n / \sigma_n \to \infty$ & $n^2 \sigma_n$ & $\frac{2}{\sqrt{2 \pi}} \int_S p^2 (s) \ \mathrm{d} s$ & $\frac{2}{\sqrt{2 \pi}} \int_S p^2 (s) \ \mathrm{d} s$ \\
      \hline
      weighted $r_n/\sigma_n \to 0$ & $\sigma_n^{-d} n^{2} r_n^{d+1}$ & $\frac{2 \eta_{d-1} \eta_{d}^{-1-1/d}}{(d+1) (2\pi)^{d/2}} \int_S p^{1-1/d} (s) \ \mathrm{d} s$ & $\frac{2 \eta_{d-1}}{(d+1) (2 \pi)^{d/2}} \int_S p^2 (s) \ \mathrm{d} s$ \\
      \hline
      \multicolumn{4}{l}{}\\
      \hline
      \multicolumn{4}{|l|}{\bf The $\cut$ in the complete weighted graph}\\
      \hline
      Weighting & $s_n^{\cut}$ & \multicolumn{2}{|c|}{$\cutlim$ in complete weighted graph} \\
      \hline
      \hline
      weighted & $n^2 \sigma_n$ & \multicolumn{2}{|c|}{$\frac{2}{\sqrt{2 \pi}} \int_S p^2 (s) \ \mathrm{d} s$}  \\
      \hline
      \multicolumn{4}{l}{}\\
      \hline
      \multicolumn{4}{|l|}{\bf The volume in the $\kNN$-graph and the $r$-graph}\\
      \hline
      Weighting & $s_n^{\vol}$ & $\vollim (H)$ $\kNN$-graph & $\vollim (H)$ $r$-graph \\
      \hline
      \hline
      unweighted & $n^2  r_n^d$ & $\int_H p(x) \ \mathrm{d} x$ & $\eta_d \int_H p^2 (x) \ \mathrm{d} x$ \\
      \hline
      weighted, $r_n / \sigma_n \to \infty$ & $n^2$ & $\int_H p^2 (x) \ \mathrm{d} x$ & $\int_H p^2 (x) \ \mathrm{d} x$ \\
      \hline
      weighted, $r_n/\sigma_n \to 0$  & $\sigma_n^{-d} n^2 r_n^d$ & $\frac{1}{(2 \pi)^{d/2}} \int_H p(x) \ \mathrm{d} x$ & $\frac{\eta_d}{(2 \pi)^{d/2}} \int_H p^2 (x) \ \mathrm{d} x$ \\
      \hline 
      \multicolumn{4}{l}{}\\ 
      \hline
      \multicolumn{4}{|l|}{\bf The volume in the complete weighted graph}\\
      \hline
      Weighting & $s_n^{\vol}$ & \multicolumn{2}{|c|}{$\vollim$ in complete weighted graph} \\
      \hline
      \hline
      weighted & $n^2$ & \multicolumn{2}{|c|}{$\int_H p^2 (x) \ \mathrm{d} x$} \\
      \hline
    \end{tabular}
    \caption{The scaling sequences and limit expression for the $\cut$ 
      and the volume in all the considered graph types. 
      In the limit expression for the $\cut$ the integral denotes the $(d-1)$-dimensional 
      surface integral along the hyperplane $S$,
      whereas in the limit expressions for the volume the integral denotes the Lebesgue
      integral over the halfspace $H=H^+$ or $H=H^-$.}
    \label{table:cut_limits}
  \end{table}

  Having defined these limits we define, analogously to the definitions in 
  Equations~\eqref{eq:def_ncut} and~\eqref{eq:def_cheegercut}, the limits 
  of $\ncut$ and $\cheegercut$ as
  \begin{align}
    \label{def:ncutlim}
    \ncutlim &= \frac{\cutlim}{\vollim \left( H^+ \right)} 
    + \frac{\cutlim}{\vollim \left( H^- \right)} 
    \intertext{and}
    \label{def:cheegercutlim}
    \cheegercutlim &= \frac{\cutlim}{\min \left\{ \vollim \left( H^+ \right), \vollim \left( H^- \right) \right\}} .
  \end{align}
  In our following main theorems we show the conditions under which we have
  for $n \to \infty$ almost sure convergence of
  \begin{align*}
    \frac{s_n^{\vol}}{s_n^{\cut}} \ncut_n \to \ncutlim
    \qquad \text{ and } \qquad \frac{s_n^{\vol}}{s_n^{\cut}} \cheegercut_n \to \cheegercutlim .
  \end{align*}
  Furthermore, for the unweighted $r$-graph and $\kNN$-graph and for the 
  complete weighted graph with Gaussian weights we state the optimal convergence rates, 
  where ``optimal'' means the best trade-off between 
  our bounds for different quantities derived in Section~\ref{sec:proofs}. 
  Note that we will not prove the following theorems here. Rather the proof of 
  Theorem~\ref{thm:limit_ncut_cheegercut_knn_graph} can be found in 
  Section~\ref{sec:proof_main_theorem_knn}, whereas the proofs of 
  Theorems~\ref{thm:limit_ncut_cheegercut_r_graph} 
  and~\ref{thm:limit_ncut_cheegercut_complete_graph} can be found in 
  Section~\ref{sec:proofs_main_theorems_r_graph}.

  \begin{theorem}[$\ncut$ and $\cheegercut$ on the $\kNN$-graph] 
    \label{thm:limit_ncut_cheegercut_knn_graph} 
    For a sequence $(k_n)_{n \in \N}$ with $k_n/n \to 0$ for $n \to \infty$
    let $G_n$ be the $k_n$-nearest neighbor graph on the 
    sample $x_1, \ldots ,x_n$.
    Set $\xcut=\ncut$ or $\xcut=\cheegercut$ and let $\xcutlim$ denote the
    corresponding limit as defined in Equations~\eqref{def:ncutlim} 
    and~\eqref{def:cheegercutlim}. 
    Set
    \begin{align*}
      \Delta_n = \left| \frac{s_n^{\vol}}{s_n^{\cut}} \xcut_n - \xcutlim \right| .
    \end{align*}
    \begin{itemize}
    \item
      Let $G_n$ be the unweighted $\kNN$ graph. If $k_n/\sqrt{n \log n} \to \infty$ in 
      the case $d=1$ and $k_n/\log n \to \infty$ in the case $d \geq 2$ we have 
      $\Delta_n \to 0$ for $n \to \infty$ almost surely.
      The optimal convergence rate is achieved for $k_n=k_0 \sqrt[4]{n^3 \log n}$ in 
      the case $d=1$ and $k_n=k_0 n^{2/(d+2)} (\log n)^{d/(d+2)}$ in the case $d \geq 2$.
      For this choice of $k_n$ we have $\Delta_n = O (\sqrt[d+4]{\log n/n})$ 
      in the case $d=1$ and $\Delta_n = O (\sqrt[d+2]{\log n/n})$ for $d \geq 2$.
    \item
      Let $G_n$ be the $\kNN$-graph with Gaussian weights and suppose 
      $r_n \geq \sigma_n^\alpha$ for an $\alpha \in (0,1)$.
      Then we have almost sure convergence of $\Delta_n \to 0$ for $n \to \infty$ 
      if $k_n/\log n \to \infty$ and $n \sigma_n^{d+1}/\log n \to \infty$.
    \item
      Let $G_n$ be the $\kNN$-graph with Gaussian weights and $r_n/\sigma_n \to 0$.
      Then we have almost sure convergence of $\Delta_n \to 0$ for $n \to \infty$ 
      if $k_n/\sqrt{n \log n} \to \infty$ in 
      the case $d=1$ and $k_n/\log n \to \infty$ in the case $d \geq 2$.
    \end{itemize}    
  \end{theorem}

  \begin{theorem}[$\ncut$ and $\cheegercut$ on the $r$-graph] 
    \label{thm:limit_ncut_cheegercut_r_graph} 
    For a sequence $(r_n)_{n \in \N} \subseteq \R_{>0}$ with $r_n \to 0$ for $n \to \infty$
    let $G_n$ be the $r_n$-neighborhood graph on the sample $x_1, \ldots ,x_n$.
    Set $\xcut=\ncut$ or $\xcut=\cheegercut$ and let $\xcutlim$ denote the
    corresponding limit as defined in Equations~\eqref{def:ncutlim} 
    and~\eqref{def:cheegercutlim}. 
    Set
    \begin{align*}
      \Delta_n = \left| \frac{s_n^{\vol}}{s_n^{\cut}} \xcut_n - \xcutlim \right| .
    \end{align*}
    \begin{itemize}
    \item
      Let $G_n$ be unweighted. Then $\Delta_n \to 0$ almost surely for $n \to \infty$
      if $n r_n^{d+1}/\log n \to \infty$. 
      The optimal convergence rate is achieved for $r_n = r_0 \sqrt[d+3]{\log n/n}$
      for a suitable constant $r_0>0$.
      For this choice of $r_n$ we have $\Delta_n = O ( \sqrt[d+3]{\log n/n} )$.
    \item
      Let $G_n$ be weighted with Gaussian weights with bandwidth $\sigma_n \to 0$
      and $r_n/\sigma_n \to \infty$ for $n \to \infty$.
      Then $\Delta_n \to 0$ almost surely for $n \to \infty$ 
      if $n \sigma_n^{d+1} / \log n \to \infty$.
    \item
      Let $G_n$ be weighted with Gaussian weights with bandwidth $\sigma_n \to 0$
      and $r_n/\sigma_n \to 0$ for $n \to \infty$.
      Then $\Delta_n \to 0$ almost surely for $n \to \infty$ 
      if $n r_n^{d+1} / \log n \to \infty$.
    \end{itemize}    
  \end{theorem}

  The following theorem presents the limit results for $\ncut$ and $\cheegercut$ 
  on the complete weighted graph. 
  One result that we need in the proof of this theorem is 
  Corollary~\ref{cor:cut_complete_graph} on the convergence of the $\cut$.
  Note that in~\citet{Narayanan/Belkin/Niyogi:2007} a similar $\cut$ convergence 
  problem is studied for the case of the complete weighted graph,
  and the scaling sequence and the limit differ from ours.
  However, the reason is that in that paper the {\em weighted} $\cut$ is considered,
  which can be written as $f' L_{\text{norm}} f$, where $L_{\text{norm}}$ denotes the
  normalized graph Laplacian matrix and $f$ is an $n$-dimensional vector with $f_i=1$ if 
  $x_i$ is in one cluster and $f_i=0$ if $x_i$ is in the other cluster.
  On the other hand, the standard $\cut$, which we consider in this paper,
  can be written (up to a constant) as $f' L_{\text{unnorm}} f$, 
  where $L_{\text{unnorm}}$ denotes the unnormalized graph Laplacian matrix.
  (For the definitions of the graph Laplacian matrices and their relationship to 
  the $\cut$ we refer the reader to~\citet{Luxburg07}.)
  Therefore, the two results do not contradict each other.
 
  \begin{theorem}[$\ncut$ and $\cheegercut$ on the complete weighted graph] 
    \label{thm:limit_ncut_cheegercut_complete_graph} 
    Let $G_n$ be the complete weighted graph with Gaussian weights and bandwidth $\sigma_n$
    on the sample points $x_1, \ldots ,x_n$.
    Set $\xcut=\ncut$ or $\xcut=\cheegercut$ and let $\xcutlim$ denote the
    corresponding limit as defined in Equations~\eqref{def:ncutlim} 
    and~\eqref{def:cheegercutlim}. 
    Set
    \begin{align*}
      \Delta_n = \left| \frac{s_n^{\vol}}{s_n^{\cut}} \xcut_n - \xcutlim \right| .
    \end{align*}
    Under the conditions $\sigma_n \to 0$ and $n \sigma_n^{d+1}/\log n \to \infty$
    we have almost surely $\Delta_n \to 0$ for $n \to \infty$.
    The optimal convergence rate is achieved setting 
    $\sigma_n = \sigma_0 \sqrt[d+3]{\log n/n}$
    with a suitable $\sigma_0>0$.
    For this choice of $\sigma_n$ the convergence rate is in
    $O ( ((\log n)/n)^{\alpha/(d+3)} )$ for any $\alpha \in (0,1)$.
  \end{theorem}

  Let us decrypt these results and for simplicity focus on the $\cut$
  value. 
  When we compare the limits of the $\cut$
  (cf. Table~\ref{table:cut_limits})  it is striking that, depending on the graph 
  type and the weighting scheme, there are two substantially different limits:
  the limit $\int_S p^2 (s) \ \mathrm{d} s$ for  
  the unweighted $r$-neighborhood graph, and the limit $\int_S p^{1-1/d} (s)  \ \mathrm{d} s$
  for the unweighted $k$-nearest neighbor graph.
  
  The limit of the $\cut$ for the complete weighted graph with Gaussian weights is the
  same as the limit for the unweighted $r$-neighborhood graph. 
  There is a simple reason for that: On both graph types the weight of an edge
  only depends on the distance between its end points, no matter where the points are.
  This is in contrast to the $\kNN$-graph, where the radius up to which a point is connected
  strongly depends on its location: If a point is in a region of high density
  there will be many other points close by, which means that the radius is small. 
  On the other hand, this radius is large for points in low-density regions.
  Furthermore, the Gaussian weights decline very rapidly with the distance, depending 
  on the parameter $\sigma$. That is, $\sigma$ plays a similar role as the radius $r$
  for the $r$-neighborhood graph.
  
  The two types of $r$-neighborhood graphs with Gaussian weights have the same limit as 
  the unweighted $r$-neighborhood graph and the complete weighted graph with Gaussian weights.
  When we compare the scaling sequences $s_n^{\cut}$ it turns out that in the case
  $r_n/\sigma_n \to \infty$ this sequence is the same as for the complete weighted graph,
  whereas in the case $r_n/\sigma_n \to 0$ we have $s_n^{\cut}=n^2 r_n^{d+1}/\sigma_n^d$,
  which is the same sequence as for the unweighted $r$-graph corrected by a factor of 
  $\sigma_n^{-d}$. 
  In fact, these effects are easy to explain: If $r_n/\sigma_n \to \infty$ then the edges
  which we have to remove from the complete weighted graph in order to obtain the $r_n$-neighborhood
  graph have a very small weight and their contribution to the value of the $\cut$ can be 
  neglected. Therefore this graph behaves like the complete weighted graph with Gaussian weights.
  On the other hand, if $r_n/\sigma_n \to 0$ then all the edges that remain in the
  $r_n$-neighborhood graph have approximately the same weight, namely the maximum
  of the Gaussian weight function, which is linear in $\sigma_n^{-d}$.
  
  Similar effects can be observed for the $k$-nearest neighbor graphs. The limits of
  the unweighted graph and the graph with Gaussian weight and $r_n/\sigma_n \to 0$
  are identical (up to constants) and the scaling sequence has to correct for the maximum
  of the Gaussian weight function.
  However, the limit for the $\kNN$-graph with Gaussian weights and $r_n/\sigma_n \to \infty$
  is different: In fact, we have the same limit expression as for the complete weighted graph with 
  Gaussian weights. The reason for this is the following: Since $r_n$ is large compared to 
  $\sigma_n$ at some point all the $k$-nearest neighbor radii of the sample points are very
  large. Therefore, all the edges that are in the complete weighted graph but not in the $\kNN$ graph
  have very low weights and thus the limit of this graph behaves like the limit of the 
  complete weighted graph with Gaussian weights.
  
  Finally, we would like to discuss the difference between the two limit expressions,
  where as examples for the graphs we use only the unweighted $r$-neighborhood graph and 
  the unweighted $\kNN$-graph. Of course, the results can be carried over  to the other 
  graph types.
  For the $\cut$ we have the limits $\int_S p^{1-1/d} (s)  \ \mathrm{d} s$ and 
  $\int_S p^2 (s) \ \mathrm{d} s$. In dimension 1 the difference between these 
  expressions is most pronounced: The limit for the $\kNN$ graph does not depend 
  on the density $p$ at all, whereas in the limit for the $r$-graph the exponent of $p$
  is $2$, independent of the dimension. Generally, the limit for the $r$-graph seems to 
  be more sensitive to the absolute value of the density. This can also be seen 
  for the volume: The limit expression for the $\kNN$ graph is $\int_H p (x)  \ \mathrm{d} x$,
  which does not depend on the absolute value of the density at all, but only on 
  the probability mass in the halfspace $H$. This is different for the unweighted
  $r$-neighborhood graph with the limit expression $\int_H p^2 (x)  \ \mathrm{d} x$.
  
  \section{Examples where different limits of Ncut lead to different
    optimal cuts} \label{sec:experiments}

  \begin{figure}[t]
    \centering
    \begin{tabular}{cc}
      \includegraphics[height= 0.12\textheight]{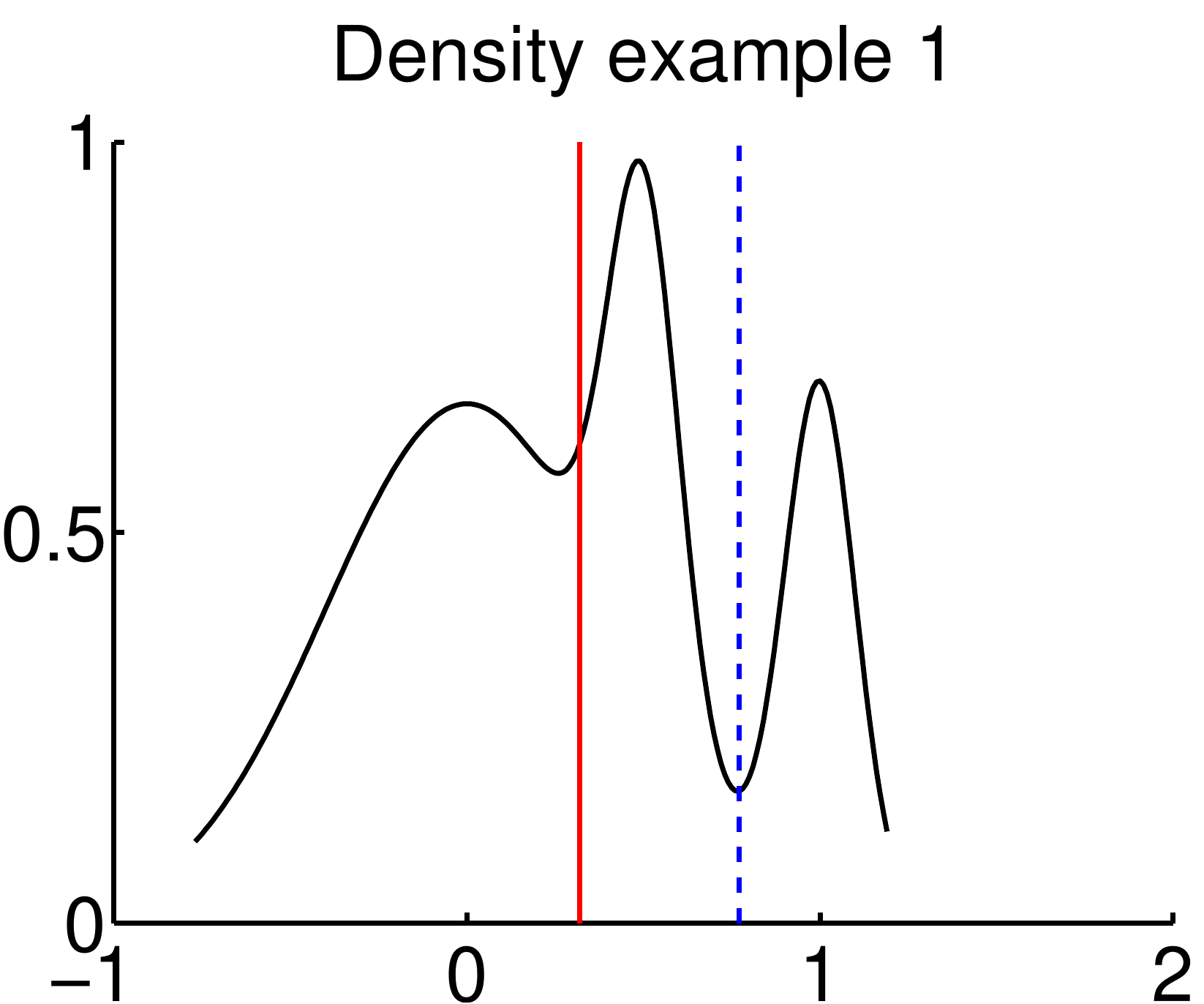} &
      \includegraphics[height= 0.12\textheight]{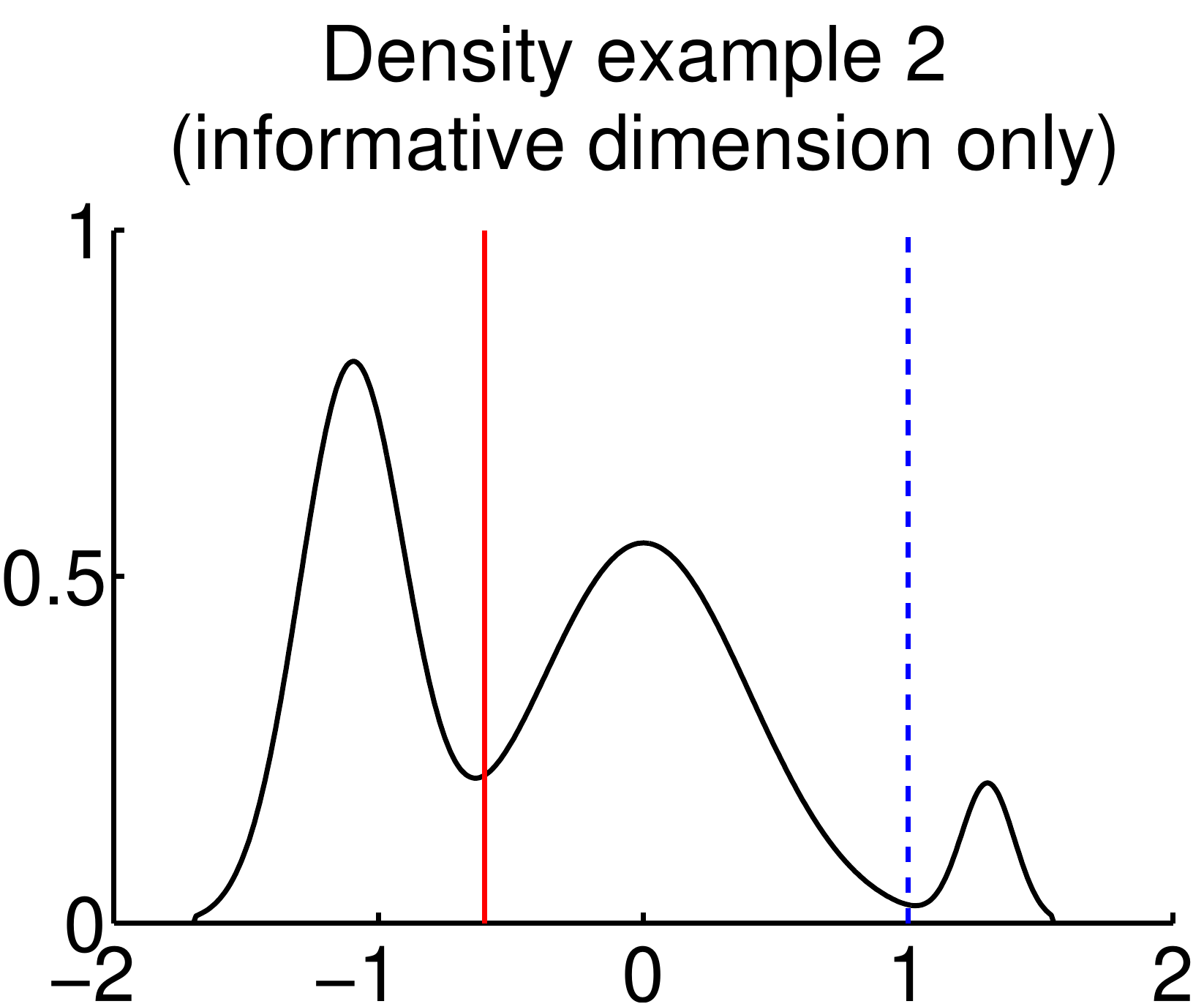}
    \end{tabular}
    \caption{Densities in the examples. In the two-dimensional case, we plot the informative
      dimension (marginal over the other dimensions) only. The dashed blue vertical line depicts the
      optimal limit cut of the $r$-graph, the solid red vertical line
      the optimal limit cut of the $\kNN$ graph. }
    \label{fig-densities}
  \end{figure}

  In Theorems~\ref{thm:limit_ncut_cheegercut_knn_graph}-\ref{thm:limit_ncut_cheegercut_complete_graph} 
  we have proved that the limit expressions for $\ncut$ and $\cheegercut$ 
  are different for different kinds of neighborhood graphs. In fact, 
  apart from constants there are two limit expressions: that of the 
  unweighted $\kNN$-graph, where the exponent of the density $p$ in the 
  limit integral for the $\cut$ is $1-1/d$ and for the volume is $1$,
  and that of the unweighted $r$-neighborhood graph, where the exponent
  in the limit of the $\cut$ is $2$ and in the limit of the $\vol$ is $1$.
  Therefore, we consider here only the unweighted $\kNN$-graph and the 
  unweighted $r$-neighborhood graph.
  
  In this section we show that the difference between the limit expressions 
  is more than a mathematical subtlety without practical relevance: 
  If we select an optimal cut based on the limit criterion for
  the $\kNN$ graph we can obtain a different result than if we use the limit
  criterion based on the $r$-neighborhood graph.

  Consider Gaussian mixture distributions in one (Example~1) and in two dimensions (Example~2)
  of the form $\textstyle \sum_{i=1}^3 \alpha_i N( [\mu_i, 0, \hdots, 0], \sigma_i I)$ 
  which are set to zero where they are below a threshold $\theta$ 
  and properly rescaled. The specific parameters in one and two dimensions are
  \begin{center}
    \begin{tabular}{l||l|l|l||l|l|l||l|l|l||l}
      dim & $\mu_1$ & $\mu_2$ & $\mu_3$ & $\sigma_1$ & $\sigma_2$ &
      $\sigma_3$ & $\alpha_1$ &  $\alpha_2$ &  $\alpha_3$ & $\theta$  \\
      \hline
      \hline
      1 & 0 & 0.5 & 1 & 0.4 & 0.1 & 0.1 & 0.66 & 0.17 & 0.17 & 0.1\\
      \hline
      2 &  $-1.1$ & $0$ & $1.3$ & 0.2 & 0.4 & 0.1 & 0.4 & 0.55 &
      0.05 & 0.01
    \end{tabular}
  \end{center}
  
  Plots of the densities of Example~1 and~2 can be seen in Figure~\ref{fig-densities}. 
  We first investigate the theoretic limit $\cut$ values, for hyperplanes which cut
  perpendicular to the first dimension (which is the ``informative''
  dimension of the data). 
  For the chosen densities, the limit $\ncut$ expressions from 
  Theorems~\ref{thm:limit_ncut_cheegercut_knn_graph} and~\ref{thm:limit_ncut_cheegercut_r_graph}  
  can be computed analytically and optimized over the chosen hyperplanes. 
  The solid red line in Figure~\ref{fig-densities} indicates the position
  of the minimal value for the $\kNN$-graph case, whereas the dashed blue line 
  indicates the the position of the minimal value for the $r$-graph case.

  \begin{figure}[t]
    \begin{center}
      \includegraphics[width=0.3\textwidth]{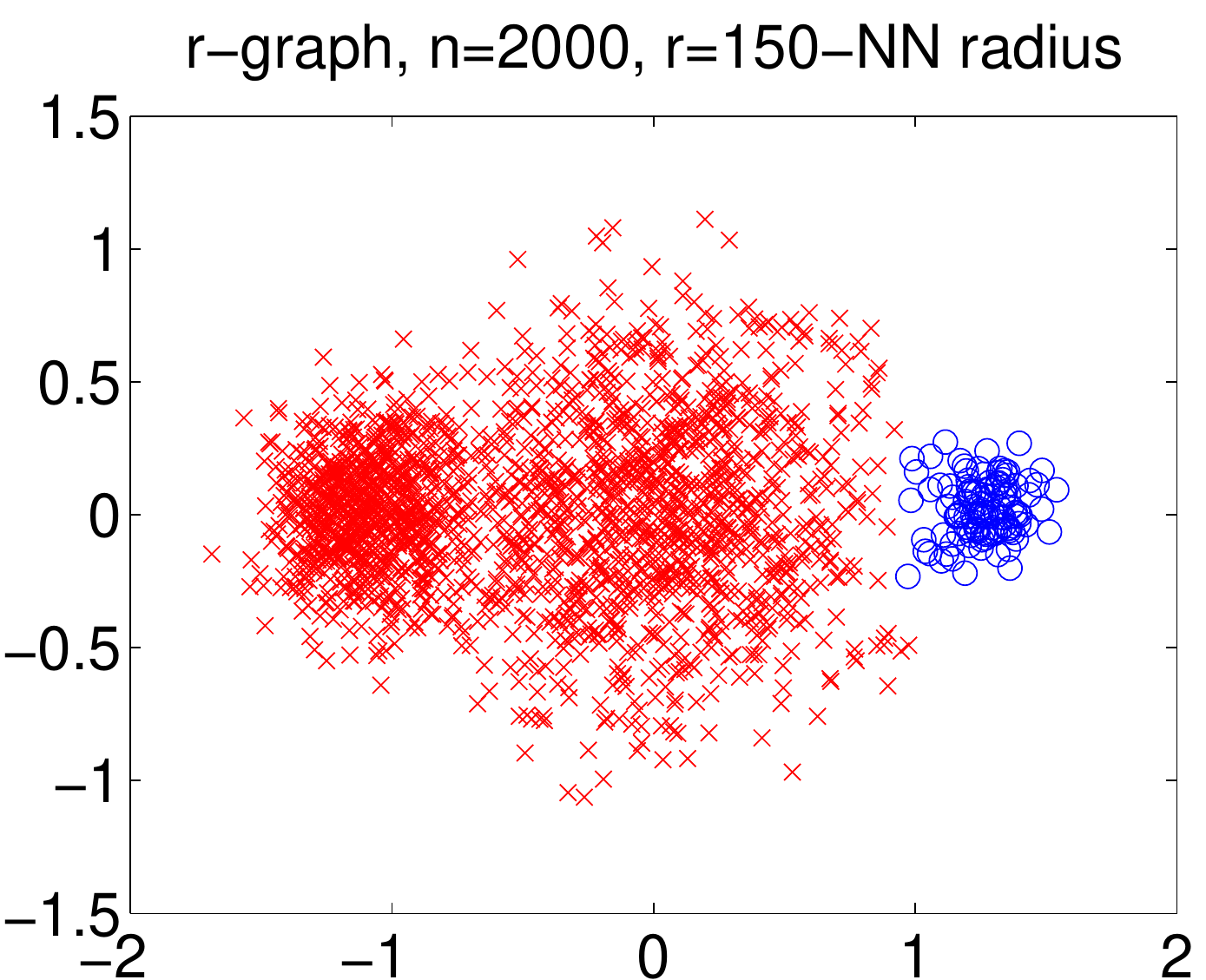} \hspace{2cm}
      \includegraphics[width=0.3\textwidth]{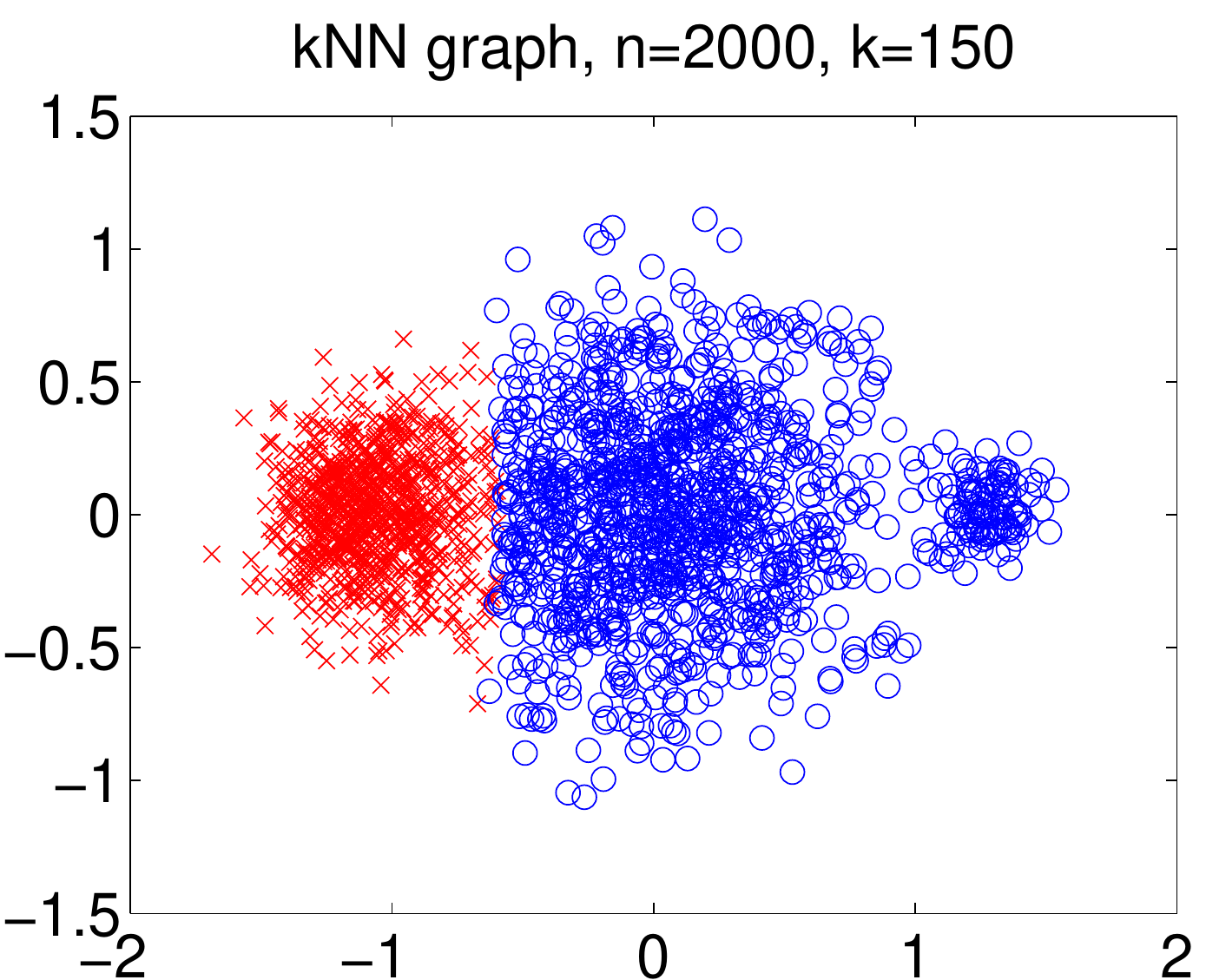}
    \end{center}
    \caption{Results of spectral clustering in two dimensions, for the unweighted 
      $r$-graph (left) and the unweighted $\kNN$ graph (right) }
    \label{fig-spectral}
  \end{figure}  
  Up to now we only compared the limits of different graphs with each other,
  but the question is, whether the effects of these limits can be observed even for 
  finite sample sizes. In order to investigate this question
  we applied normalized spectral clustering (cf.~\citet{Luxburg07})
  to sample data sets of $n=2000$ points from the mixture distribution above. 
  We used the unweighted $r$-graph and the unweighted 
  symmetric $k$-nearest neighbor graph.
  We tried a range of reasonable values for the parameters $k$ and $r$ and 
  the results we obtained were stable over a range of parameters. 
  Here we present the results for the $30$- (for $d=1$) and the 
  $150$-nearest neighbor graphs (for $d=2$) and the $r$-graphs with corresponding 
  parameter $r$, that is $r$ was set to be the mean $30$- and $150$-nearest neighbor 
  radius.
  Different clusterings are compared using the minimal matching distance:
  \begin{align*}
    d_{MM}(\text{Clust}_1, \text{Clust}_2) 
    = \min_{\pi} \frac{1}{n} \sum_{i=1}^n \mathbf{1}_{\text{Clust}_1(x_i) 
      \neq \pi ( \text{Clust}_2(x_i))} 
  \end{align*}
  where the minimum is taken over all permutations $\pi$ of the labels. In the
  case of two clusters, this distance corresponds to the 0-1-loss as
  used in
  classification: a minimal matching distance of $0.35$, say, means that
  $35 \%$ of the data points lie in different clusters.
  In our spectral clustering experiment, we could observe
  that the clusterings obtained by spectral clustering are usually very
  close to the theoretically optimal hyperplane splits predicted by
  theory (the minimal matching distances to the optimal hyperplane splits were
  always in the order of 0.03 or smaller).
  As predicted by theory, the two types of graph
  give different cuts in the data. An illustration of this phenomenon
  for the case of dimension 2 can be found in
  Figure~\ref{fig-spectral}. To give a quantitative evaluation of this
  phenomenon, we computed the mean minimal matching distances
  between clusterings obtained by the
  same type of graph over the different samples (denoted $d_{\kNN}$ and $d_{r}$), 
  and the mean difference $d_{\kNN-r}$ between the clusterings obtained by different
  graph types:
  \begin{center}
    \begin{tabular}{l|l|l|l}
      Example &  $d_{\kNN}$ &  $d_{r}$ &  $d_{\kNN-r}$ \\
      \hline
      \hline
      1 dim & $0.0005 \pm 0.0006$ & $0.0003 \pm 0.0004$  & $0.346 \pm 0.063$  \\
      \hline
      2 dim &$0.005 \pm 0.0023$ & $0.001 \pm 0.001$ & $0.49 \pm 0.01$ 
    \end{tabular}
  \end{center}

  We can see that for the same graph, the clustering results are very
  stable (differences in the order of $10^{-3}$) whereas the differences
  between the $\kNN$ graph and the $r$-neighborhood graph are
  substantial (0.35 and 0.49, respectively). This difference is exactly the
  one induced by assigning the middle mode of the density to different
  clusters, which is the effect predicted by theory.

\begin{figure}[t]
    \centering
    \begin{tabular}{ccc}
      \includegraphics[width= 0.2\textwidth]{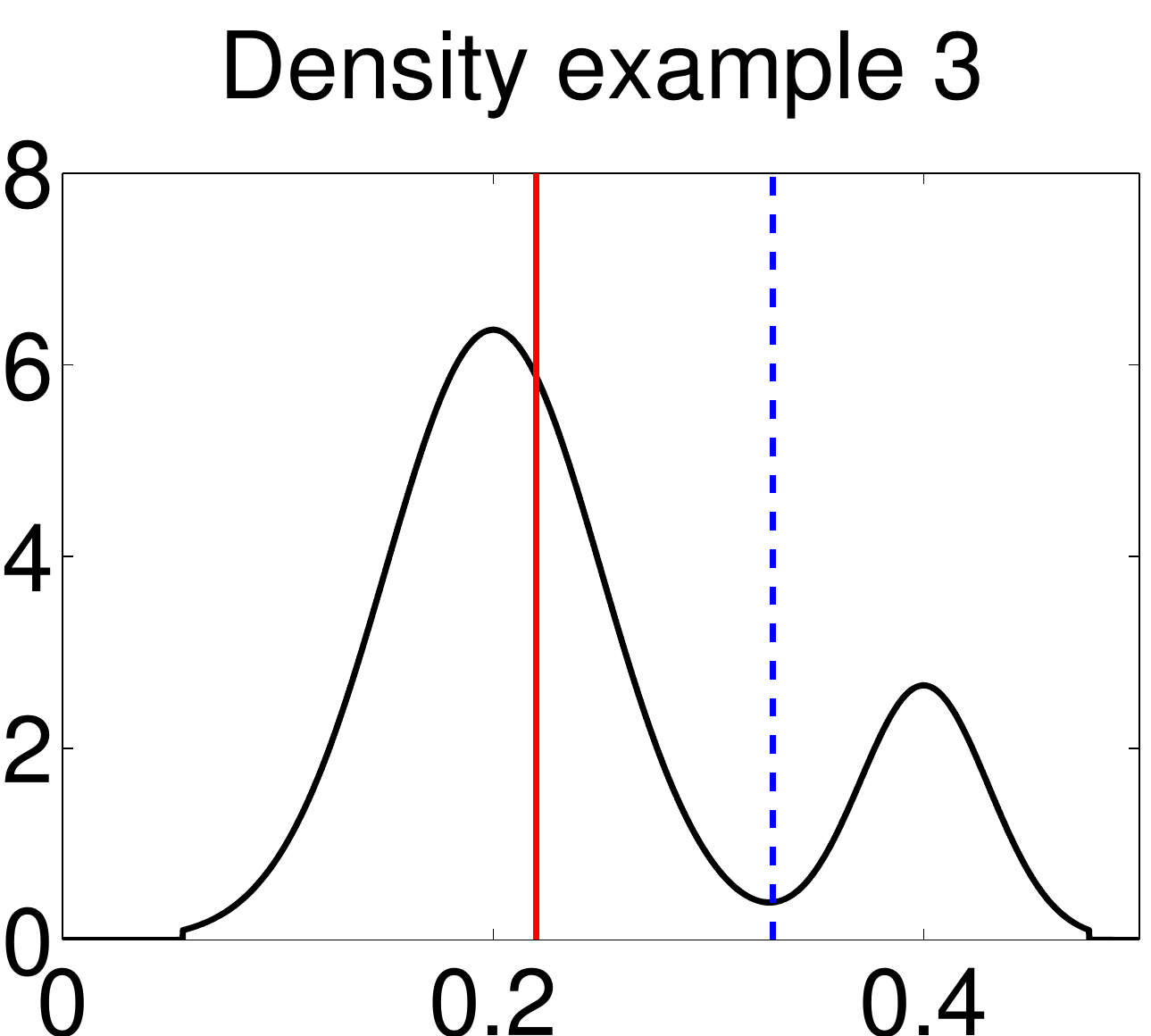} &
      \includegraphics[width= 0.2\textwidth]{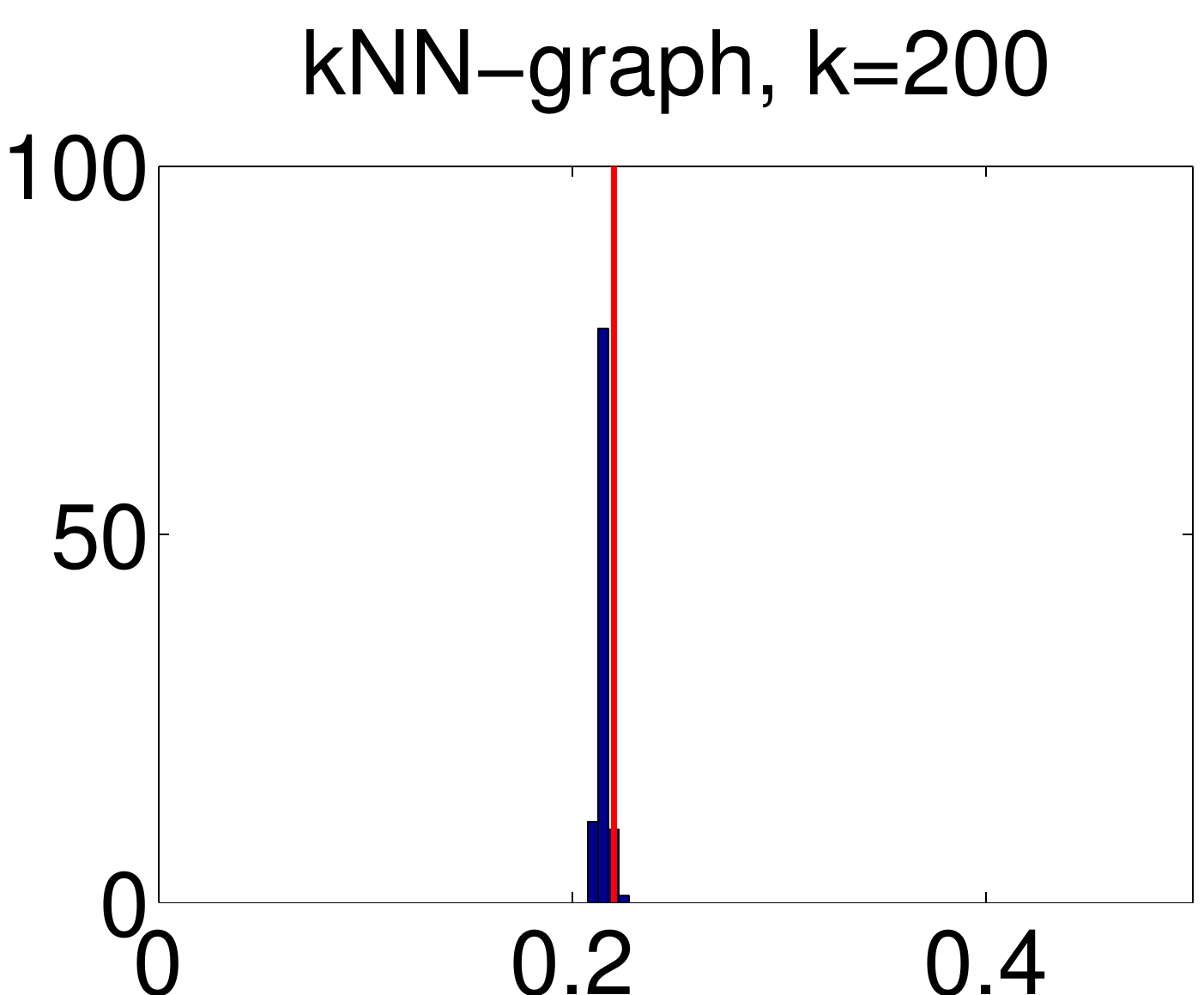} &
      \includegraphics[width= 0.2\textwidth]{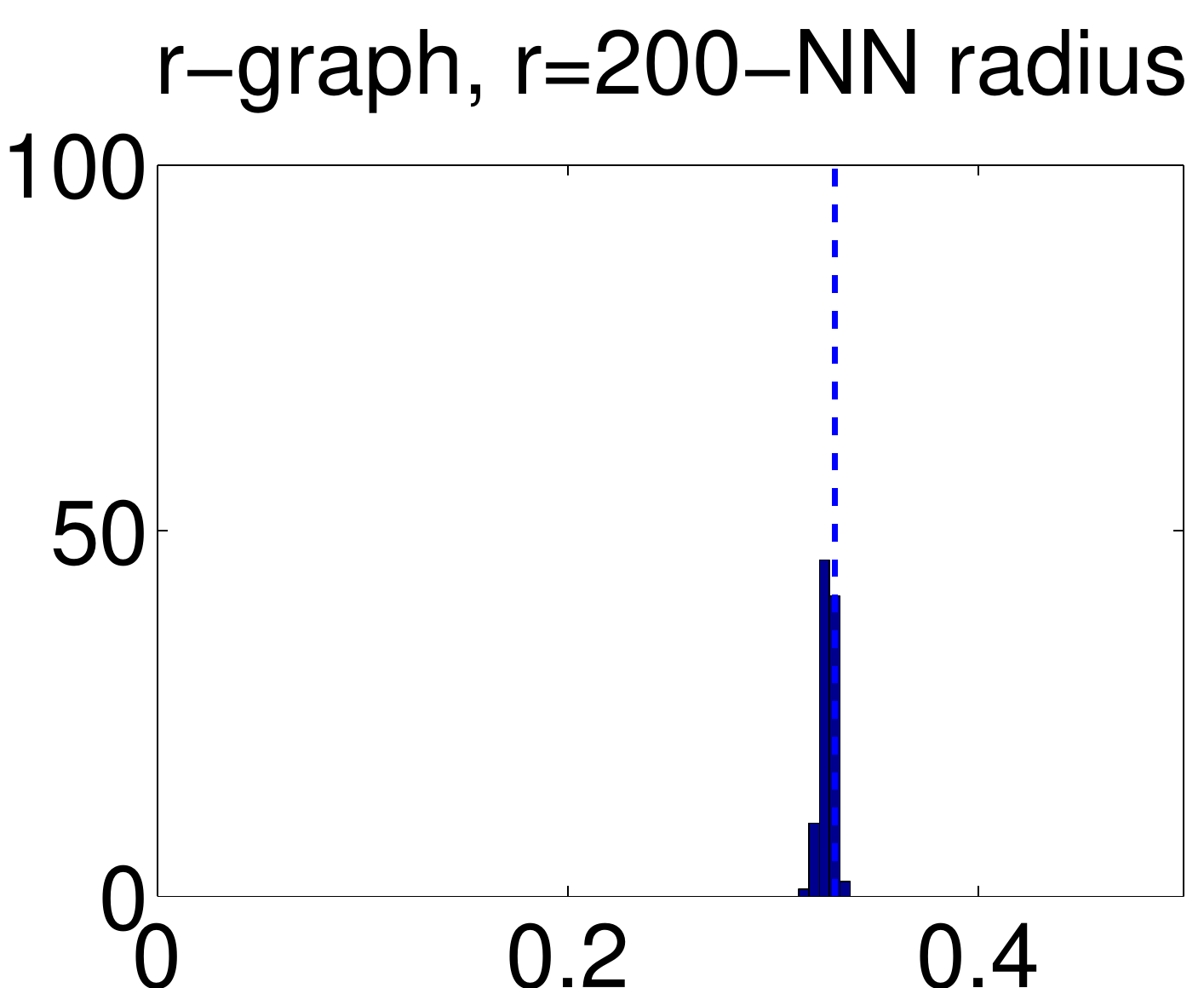}
    \end{tabular}
    \caption{The Example~3 with the sum of two Gaussians, that is two modes of the density. 
      In the left figure the density  with the optimal limit 
      cut of the $r$-graph (dashed blue vertical line) and  the optimal limit cut 
      of the $\kNN$ graph (the solid red vertical line) is depicted. 
      The two figures on the right show the histograms of
      the cluster boundary over $100$ iterations for the unweighted $r$-neighborhood
      and $\kNN$-graphs.
      \label{fig-results-two-clusters}}
  \end{figure}

  It is tempting to conjecture that in Example~1 and~2 the two different limit 
  solutions and their impact on spectral clustering might arise due to the fact that 
  the number of Gaussians and the number of clusters we are looking for do not coincide. 
  Yet the following Example~3 shows that this is not the case: 
  for a density in one dimension as above but with only two Gaussians with parameters 
  \begin{center}
    \begin{tabular}{l|l||l|l||l|l||l}
      $\mu_1$ & $\mu_2$ & $\sigma_1$ & $\sigma_2$ & $\alpha_1$ & $\alpha_2$ & $\theta$ \\
      \hline
      \hline
      0.2 & 0.4 & 0.05 & 0.03 & 0.8 & 0.2 & 0.1\\
    \end{tabular}
  \end{center}
  the same effects can be observed. The density is depicted in the left plot of 
  Figure~\ref{fig-results-two-clusters}.
  
  In this example we draw a sample of $2000$ points from this density and
  compute the spectral clustering of the points, once with the unweighted $\kNN$-graph
  and once with the unweighted $r$-graph.  
  In one dimension we can compute the place of the boundary between two clusters,
  that is the middle between the rightmost point of the left cluster and the leftmost
  point of the right cluster. We did this for $100$ iterations and plotted 
  histograms of the location of the cluster boundary. In the middle and the right plot
  of Figure~\ref{fig-results-two-clusters} we see that these coincide with the 
  optimal cut predicted by theory.

  \section{Outlook}
  In this paper we have investigated the influence of the graph
  construction on the graph-based clustering measures normalized
  cut and Cheeger cut. We have seen that depending on the type of graph
  and the weights, the clustering quality measures converge to different 
  limit results.

  This means that ultimately, the question about the ``best $\ncut$'' or 
  ``best Cheeger cut'' clustering, given infinite amount of data, has different answers,
  depending on which underlying graph we use.  This observation opens
  Pandora's box on clustering criteria: the ``meaning'' of a clustering
  criterion does not only depend on the exact definition of the
  criterion itself, but also on how the graph on the finite sample is
  constructed.  
  This means that one graph clustering quality measure is not just
  ``one well-defined criterion'' on the underlying space, but it corresponds 
  to a whole bunch of criteria, which differ depending on the underlying graph.  
  More sloppy: A clustering quality measure applied to one neighborhood graph
  does something different in terms of partitions of the underlying space 
  than the same quality measure applied to a different neighborhood
  graph. 
  This shows that these criteria cannot be studied isolated from the graph 
  they are applied to.

  From a theoretical side, there are several directions in which our
  work can be improved. 
  In this paper we only consider partitions of Euclidean space 
  that are defined by hyperplanes. This restriction is made in order to
  keep the proofs reasonably simple. However, we are confident 
  that similar results could be proven for arbitrary smooth surfaces.

  Another extension would be to obtain uniform 
  convergence results. Here one
  has to take care that one uses a suitably restricted class of
  candidate surfaces $S$ (note that uniform convergence results over the
  set of all partitions of $\R^d$ are impossible,
  cf.~\citet{BubLux09}).
  This result would be especially useful, if there existed a practically
  applicable algorithm to compute the optimal surface out of the set of all
  candidate surfaces.

  For practice, it will be important to study how the different limit
  results influence clustering results. So far, we do not have much
  intuition about when the different limit expressions lead to different
  optimal solutions, and when these solutions will show up in practice.
  The examples we provided above already show that different graphs 
  indeed can lead to systematically different clusterings in
  practice. Gaining more understanding of this effect will be an
  important direction of research if one wants to understand the nature
  of different graph clustering quality measures.

  \section{Proofs} \label{sec:proofs}
  
  In many of the proofs that are to follow in this section a lot of technique 
  is involved in order to come to terms with problems that arise due to effects 
  at the boundary of our support $C$ and to the non-uniformity of the density $p$.
  However, if these technicalities are ignored, the basic ideas of the proofs
  are simple to explain and they are similar for the different types of 
  neighborhood graphs. In Section~\ref{sec:proofs_intuitively} we discuss these 
  ideas without the technical overhead and define some quantities that are necessary
  for the formulation of our results.

  In Section~\ref{sec:proofs_knn_graph} we present the results for the 
  $k$-nearest neighbor graph and in Section~\ref{sec:proofs_r_graph}
  we present those for the $r$-graph and the complete weighted graph. 
  Each of these sections consists of three parts:
  the first is devoted to the $\cut$, the second is devoted to the volume,
  and in the third we proof the main theorem for the considered graphs
  using the results for the $\cut$ and the volume. 

  The sections on the convergence of the $\cut$ and the volume always follow the 
  same scheme: First, a proposition concerning the convergence of the $\cut$ 
  or the volume for general monotonically decreasing weight functions is given. 
  Using this general proposition the results for the specific weight functions
  we consider in this paper follow as corollaries.

  Since the basic ideas of our proofs are the same for all the different graphs,
  it is not worth repeating the same steps for all the graphs.
  Therefore, we decided to give detailed proofs for the $k$-nearest neighbor graph,
  which is the most difficult case. The $r$-neighborhood graph and the complete weighted 
  graph can be treated together and we mainly discuss the differences to
  the proof for the $\kNN$ graph.

  The limits of the $\cut$ and the volume for general weight function
  are expressed in terms of certain integrals of the weight function over ``caps'' and
  ``balls'', which are explained later. 
  For a specific weight function these integrals have to be evaluated.
  This is done in the lemmas in Section~\ref{sec:gaussian_weight_function}.
  Furthermore, this section contains a technical lemma that helps us to 
  control boundary effects.

  \subsection{Basic ideas} \label{sec:proofs_intuitively}
  In this section we present the ideas of our convergence proofs non-formally.
  We focus here on $\ncut$, but all the ideas can easily be carried over to the 
  Cheeger cut.
  
  {\em First step: Decompose $\ncut_n$ into $\cut_n$ and $\vol_n$}\\
  Under our general assumptions there exist constants $c_1,c_2,c_3$, which may depend 
  on the limit values of the cut and the volume, such that for sufficiently large $n$
  \begin{align*}
    &\left| \frac{s_n^{\vol}}{s_n^{\cut}} \left( \frac{\cut_n}{\vol_n \left( H^+ \right)} 
    + \frac{\cut_n}{\vol_n \left( H^- \right)} \right)
    -  \frac{\cutlim}{\vollim \left( H^+ \right)} 
    + \frac{\cutlim}{\vollim \left( H^- \right)}\right|\\
    &\qquad \leq c_1 \underbrace{\left| \frac{\cut_n}{\cutrate_n}  - \cutlim \right|}_{\text{$\cut$ term}}
    + c_2 \underbrace{\left| \frac{\vol_n (H^+)}{\volrate_n} - \vollim (H^+) \right|}_{\text{volume-term}}
    + c_3 \underbrace{\left| \frac{\vol_n (H^-)}{\volrate_n} - \vollim (H^-) \right|}_{\text{volume-term}} .
  \end{align*}

  {\em Second step: Bias/variance decomposition of $\cut$ and volume terms}\\
  In order to show the convergence of the $\cut$-term we do a bias/variance decomposition  
  \begin{align*}
    \left| \frac{\cut_n}{\cutrate_n} - \cutlim \right|
    \leq \underbrace{\left| \frac{\cut_n}{\cutrate_n} - \E \left( \frac{\cut_n}{\cutrate_n} \right) \right|}_{\text{variance term}}
    + \underbrace{\left| \E \left( \frac{\cut_n}{\cutrate_n} \right) - \cutlim \right|}_{\text{bias term}} 
  \end{align*}
  and show the convergence to zero of these terms separately. 
  Clearly, the same decomposition can be done for the volume terms.
  In the following we call these terms the ``bias term of the $\cut$'' and the 
  ``variance term of the $\cut$'' and similarly for the volume.
   
  For both, the $\cut$ and the volume, there is one result in this section 
  dealing with the convergence properties of the bias term and the variance
  term on each particular graph type and weighting scheme.
  
  {\em Third step: Use concentration of measure inequalities for the variance term}\\
  Bounding the deviation of a random variable from its expectation is a well-studied
  problem in statistics and there are a couple of so-called concentration of measure
  inequalities that bound the probability of a large deviation from the mean.
  In this paper we use McDiarmid's inequality for the $\kNN$ graphs 
  and a concentration of measure result for $U$-statistics by Hoeffding
  for the $r$-neighborhood graph and the complete weighted graph.
  The reason for this is that each of the graph types has its particular advantages 
  and disadvantages when it comes to the prerequisites for the concentration inequalities:
  The advantage of the $\kNN$ graph is that we can bound the degree of a node
  linearly in the parameter $k$, whereas for the $r$-neighborhood graph we can 
  bound the degree only by the trivial bound $(n-1)$ and for the complete graph
  this bound is even attained. Therefore, using the same proof as for the $\kNN$-graph
  is suboptimal for the latter two graphs.
  On the other hand, in these graphs the connectivity between points is not random 
  given their position and it is always symmetric. This allows us to use a $U$-statistics
  argument, which cannot be applied to the $\kNN$-graph, since the connectivity there
  may be unsymmetric (at least for the directed one) and the connectivity between
  each two points depends on all the sample points. 
  
  Note that these results are of a probabilistic nature, that is we obtain results of 
  the form
  \begin{align*}
    \Pr \left( \left| \frac{\cut_n}{\cutrate_n} 
    - \E \left( \frac{\cut_n}{\cutrate_n} \right) \right| > \eps \right)
    \leq p_n ,
  \end{align*}
  for a sequence $(p_n)$ of non-negative real numbers. 
  If for all $\eps>0$ the sum $\sum_{i=1}^\infty p_i$ is finite, then we have 
  almost sure convergence of the variance term to zero by the Borel-Cantelli lemma.
  
  {\em Fourth step: Bias of the $\cut$ term}\\
  While all steps so far were pretty much standard, this part is the
  technically most challenging part of our convergence proof. We have
  to prove the convergence of $\E (\cut_n / \cutrate_n)$ to $\cutlim$
  (and similarly for the volume). Omitting all technical difficulties 
  like boundary effects and the variability of the density, the basic
  ideas can be described in a rather simple manner. 

  The first idea is to break the $\cut$ down into the contributions of each single
  edge. We define a random variable $W_{ij}$ that attains the weight of the edge
  between $x_i$ and $x_j$, if these points are connected in the graph and on different 
  sides of the hyperplane $S$, and zero otherwise.
  By the linearity of the expectation and the fact that the points are sampled i.i.d.
  \begin{displaymath}
    \E \left( \cut_{n} \right) = \sum_{i=1}^n \mathop{\sum_{j=1}^n}_{j\neq i} \E W_{ij} 
    = n (n-1) \E W_{12} . 
  \end{displaymath}
  
  Now we fix the positions of the points $x_1=x$ and $x_2=y$. In this case
  $W_{ij}$ can attain only two values: $f_n (\dist(x,y))$ if the points are 
  connected and on different sides of $S$, and zero otherwise.
  We first consider the $r$-neighborhood graph with parameter $r_n$, 
  since here the existence of an edge between two points is determined by
  their distance, and is not random as in the $\kNN$ graph.
  Two points are connected if their distance is not greater than $r_n$ and
  thus $W_{ij}=0$ if $\dist(x,y)>r_n$. Furthermore, $W_{ij}=0$ if $x$ and $y$
  are on the same side of $S$.
  That is, for a point $x \in H^+$ we have
  \begin{align*}
    \E (W_{12} \; | \; x_1 = x, x_2 = y)
    = \begin{cases}
      f_n (\dist(x,y)) & \text{if $y$ is in the cap $B(x,r_n) \cap H^-$}\\
      0 & \text{otherwise.}
    \end{cases}
  \end{align*}
  By integrating over $\R^d$ we obtain
  \begin{align*}
    \E (W_{12} \; | \; x_1 = x) 
    =  \int_{B(x,r_n) \cap H^-} f_n (\dist(x,y)) p(y) \ \mathrm{d} y 
  \end{align*}
  and denote the integral on the right hand side in the following by $g(x)$.  
  
  Integrating the conditional expectation over all possible positions
  of the point $x$ in $\R^d$ gives
  \begin{displaymath}
    \E \left( W_{12} \right) = \int_{\R^d} g(x) \  p(x) \ \mathrm{d} x
    = \int_{H^+} g(x) \  p(x) \ \mathrm{d} x + \int_{H^-} g(x) \  p(x) \ \mathrm{d} x .
  \end{displaymath}
  We only consider the integral over the halfspace $H^+$ here, since the other 
  integral can be treated analogously.
  The important idea in the evaluation of this integral is the following:
  Instead of integrating over $H^+$, we initially integrate over the hyperplane $S$ 
  and then, at each point $s \in S$, along the normal line through $s$, that is the line
  $s + t n_S$ for all $t\in \R_{\geq 0}$. 
  This leads to
  \begin{displaymath}
    \int_{H^+} g(x) \ p(x) \ \mathrm{d} x
    = \int_{S} \int_{0}^{\infty} g(s+t n_S)\  p(s+t n_S) \  \mathrm{d} t \  \mathrm{d} s .
  \end{displaymath}
  
  \begin{figure}[h!]
    \centering
    \includegraphics{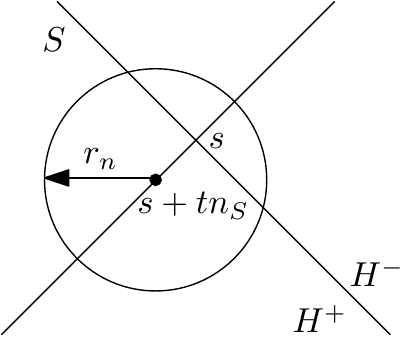}\\
    \caption{\label{fig:integration_r_graph} Integration along the normal line 
      through $s$. Obviously, for $t \geq r_n$ the
      intersection $B(s+tn_S,r_n) \cap H^-$ is empty and therefore $g (s+tn_S)=0$.
      For $0 \leq t<r_n$ the points in the cap are close to $s$ and therefore 
      the density in the cap is approximately $p(s)$.
    }    
  \end{figure}  
  This integration is illustrated in Figure~\ref{fig:integration_r_graph}.
  It has two advantages: First, if $x$ is far enough from $S$ (that
  is, $\dist(x,s)>r_n$ for all $s \in S$), then $g(x)=0$ and the
  corresponding terms in the integral vanish. Second, if $x$ is close to
  $s \in S$ and the radius $r_n$ is small,
  then the density on the ball $B(x, r_n)$ can
  be considered approximately uniform, 
  that is we assume $p(y)=p(s)$ for all $y \in B(x, r_n)$. Thus,
  \begin{align*}
    &\int_{0}^{\infty} g(s+t n_S) \ p(s+t n_S) \  \mathrm{d} t
    = \int_{0}^{r_n} g(s+t n_S) \ p(s+t n_S) \  \mathrm{d} t \\
    &\qquad = p(s) \int_{0}^{r_n} g(s+t n_S) \ \mathrm{d} t 
    = p^2 (s) \int_{0}^{r_n} \int_{B(x,r_n) \cap H^-} f_n (\dist(x,y)) 
    \ \mathrm{d} y \ \mathrm{d} t \\
    &\qquad = \eta_{d-1} \int_0^{r_n} u^d f_n (u) \ \mathrm{d} u \ p^2 (s) 
  \end{align*}
  where the last step follows with Lemma~\ref{lem:cap_integral}.
  
  Since this integral of the weight function $f_n$ over the ``caps'' plays 
  such an important role in the derivation of our results we introduce a special 
  notation for it:
  For a radius $r \in \R_{\geq 0}$ and $q=1,2$ we define
  \begin{align*}
    \capintegral^{(q)} (r) = \eta_{d-1} \int_{0}^{r} u^d f_n^q (u) \ \mathrm{d} u .
  \end{align*}
  Although these integrals also depend on $n$ we do not make this dependence 
  explicit. In fact, the parameter~$r$ is replaced by the radius $r_n$ in the 
  case of the $r$-neighborhood graph or by a different graph parameter depending on $n$
  for the other neighborhood graphs.
  Therefore the dependence of $\capintegral^{(q)} (r_n)$ on $n$ will be understood.
  Note that we allow the notation $\capintegral^{(q)} ( \infty)$, if the indefinite integral 
  exists.
  The integral $\capintegral^{(q)}$ for $q=2$ is needed for the following reason:
  For the $U$-statistics 
  bound on the variance term we do not only have to compute the expectation 
  of $W_{ij}$, but also their variance. But the variance can in turn be bounded
  by the expectation of $W_{ij}^2$, which is expressed in terms of $\capintegral^{(2)} (r_n)$.

  In the $r$-neighborhood graph points are only connected within a certain 
  radius~$r_n$, which means that to compute $\E (W_{12} \; | \; x_1 = x)$ 
  we only have to integrate over the ball $B(x,r_n)$, since all other points
  cannot be connected to $x_1=x$. This is clearly different for the complete
  graph, where every point is connected to every other point.
  The idea is to fix a radius $r_n$ in such a way as to make sure that
  the contribution of edges to points outside $B(x,r_n)$ can be neglected,
  because their weight is small. Since $W_{12}=f_n (\dist(x_1,x_2))$ if the points
  are on different sides of $S$ we have for $x \in H^+$
  \begin{align*}
    \E (W_{12} \; | \; x_1 = x) 
    &= \int_{B(x,r_n) \cap H^-} f_n (\dist(x,y)) \ p(y) \ \mathrm{d} y 
    + \int_{B(x,r_n)^c \cap H^-} f_n (\dist(x,y)) \ p(y) \ \mathrm{d} y \\
    &\leq g(x) + p_{\max} \int_{B(x,r_n)^c} f_n (\dist (x,y)) \ \mathrm{d} y .
  \end{align*}
  For the Gaussian weight function the integral converges to zero very quickly,
  if $r_n/\sigma_n \to \infty$ for $n \to \infty$.
  Thus we can treat the complete graph almost as the $r$-neighborhood graph.

  For the $k$-nearest neighbor graph the connectedness of points 
  depends on their $k$-nearest neighbor radii 
  that is, the distance of the point to its $k$-th nearest neighbor,
  which is itself a random variable.
  However, one can show that with very high probability the $k$-nearest
  neighbor radius of a point in a region with uniform density $p$ 
  is concentrated around $(k_n / ((n-1) \eta_d p)^{1/d}$.
  Since we assume that $k_n/n \to 0$ for $n \to \infty$ the expected $\kNN$ radius 
  converges to zero. Thus the density in balls with this radius
  is close to uniform and the estimate becomes more accurate.
  Upper and lower bounds on the $k$-nearest neighbor radius that hold with high probability
  are given in Lemma~\ref{lem:prob_knn_radii}.
  The idea is to perform the integration above for both, the lower bound on the 
  $\kNN$ radius and the upper bound on the $\kNN$ radius. 
  Then it is shown that these integrals converge to the same limit.

  {\em Fifth step: Bias of the volume terms}\\
  The bias of the volume term can be treated similarly to the cut
  term. We define $W_{ij}=f_n(\dist(x_i,x_j)$ 
  if $x_i$ and $x_j$ are connected in the
  graph and $W_{ij}=0$ otherwise. Note that we do not need the condition that
  the points have to be on different sides of the hyperplane~$S$ as for the $\cut$.
  Then, for a point $x \in C$ if we assume that the density is uniform within distance
  $r_n$ around $x$
  \begin{align*}
    \E ( W_{12} \; | \; x_1=x ) &= \int_{B(x,r_n)} f_n (\dist(x,y)) p(y) \ \mathrm{d} y
    = p(x) \int_{B(x,r_n)} f_n (\dist(x,y)) \ \mathrm{d} y \\
    &= d \eta_d \int_0^{r_n} u^{d-1} f_n(u) \ \mathrm{d} u \ p(x) ,
  \end{align*}
  where the last integral transform follows with Lemma~\ref{lem:ball_integral}.
  Integrating over $\R^d$ we obtain
  \begin{align*}
    \E ( W_{12} ) = \int_{\R^d} \E ( W_{12} \; | \; x_1=x ) p(x) \ \mathrm{d} x
    = d \eta_d \int_0^{r_n} u^{d-1} f_n(u) \ \mathrm{d} u 
    \int_{\R^d} p^2 (x) \ \mathrm{d} x .
  \end{align*}
  Since the integral over the balls is so important in the formulation of our
  general results we often call it the ``ball integral'' and introduce the notation
  \begin{align*}
    \ballintegral^{(q)} (r) = d \eta_d \int_0^{r_n} u^{d-1} f_n(u) \ \mathrm{d} u 
  \end{align*}
  for a radius $r>0$ and $q=1,2$. The remarks that were made on the ``cap integral'' 
  $\capintegral (r)$ above also apply to the ``ball integral'' $\ballintegral (r)$.

  {\em Sixth step: Plugging in the weight functions}\\
  Having derived results on the bias term of the $\cut$ and volume for general weight
  functions, we can now plug in the specific weight functions in which we are interested
  in this paper. This boils down to the evaluation of the ``cap'' and ``ball''
  integrals $\capintegral (r_n)$ and $\ballintegral (r_n)$ for these weight functions.
  For the unit weight function the integrals can be computed exactly, 
  whereas for the Gaussian weight function we study the asymptotic behavior of the
  ``cap'' and ``ball'' integral in the cases $r_n/\sigma_n \to 0$ and 
  $r_n/\sigma_n \to \infty$ for $n \to \infty$.

  \subsection{Proofs for the $k$-nearest neighbor graph}  
  \label{sec:proofs_knn_graph}
  
  As we have already mentioned we will give the proofs of our general propositions
  in detail here and then discuss in Section~\ref{sec:proofs_r_graph} how they
  have to be adapted to the $r$-neighborhood graph and the complete weighted graph.
  This means, that Lemmas~\ref{lem:cap_integral} and~\ref{lem:ball_integral}
  that are necessary for the proof of the general 
  propositions can be found in this section, although they are also needed for the
  $r$-graph and the $k$-nearest neighbor graph.
  
  This section consists of four subsections: In Section~\ref{sec:knn_radii}
  we define some quantities that help us to deal with the fact that the connectivity
  between two points is random even if we know their distance. These quantities 
  will play an important role in the succeeding sections.
  Section~\ref{sec:cut_term_knn} presents the results for the $\cut$ term, 
  whereas Section~\ref{sec:vol_term_knn} presents the results for the volume term.
  Finally, these results are used to proof
  Theorem~\ref{thm:limit_ncut_cheegercut_knn_graph}, the main theorem for the
  $k$-nearest neighbor graph in Section~\ref{sec:proof_main_theorem_knn}.

\begin{figure}[t]
    \begin{center}
      \includegraphics[width=0.8\textwidth]{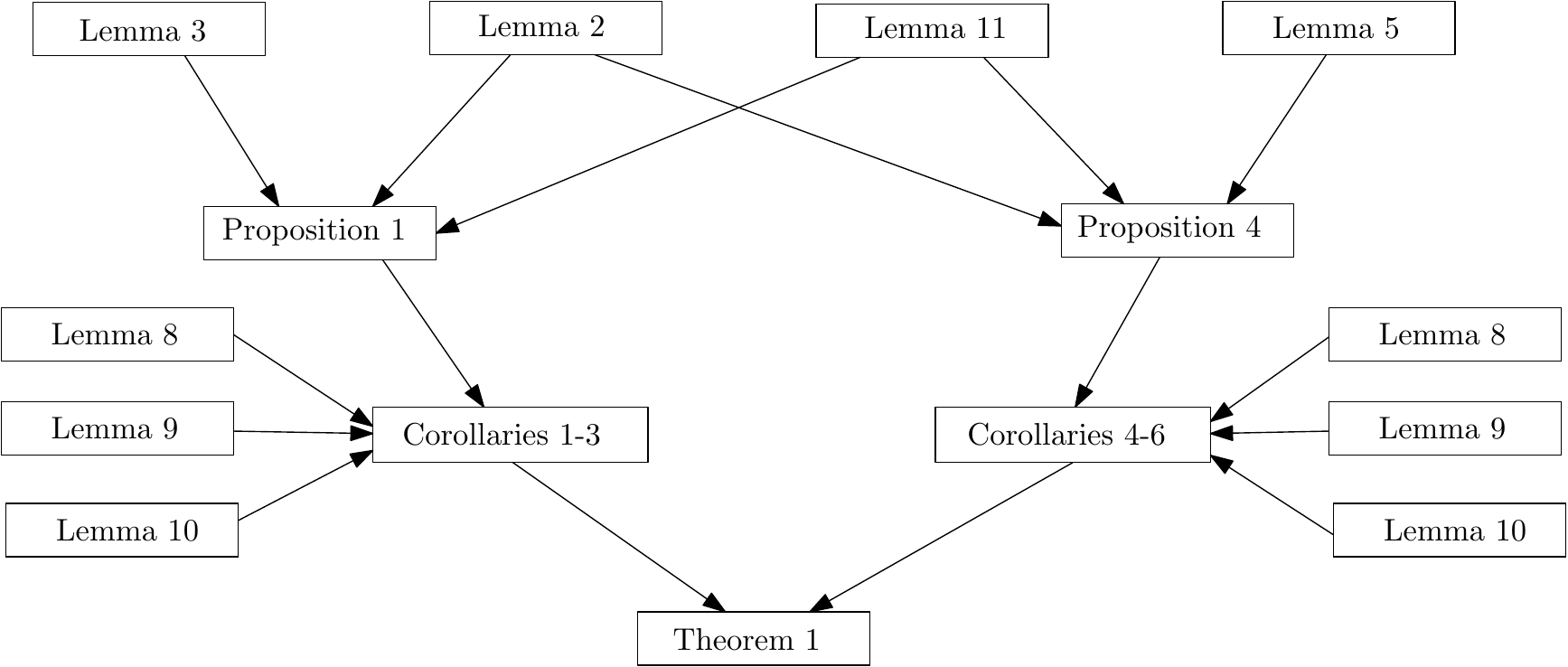}
    \end{center}
    \caption{The structure of the proofs in this section. 
      Proposition~\ref{pro:cut_knn_general} and~\ref{pro:volume_knn_general}
      state bounds for general weight functions on the bias and the variance term of the 
      $\cut$ and the volume, respectively. Lemma~\ref{lem:prob_knn_radii} shows the
      concentration of the $\kNN$ radii, Lemma~\ref{lem:boundary_strip_to_zero}
      is needed to bound the influence of points close to the boundary.
      Lemma~\ref{lem:cap_integral} and~\ref{lem:ball_integral} perform the integration
      of the weight function over ``caps'' and ``balls''. 
      In Lemmas~\ref{pro:integration_unit_weight_function}-\ref{pro:gaussian_weight_function} the general ``ball'' and ``cap'' integrals are evaluated for the specific weight functions we use.
      Using these results, Corollaries~\ref{cor:cut_knn_unweighted}-\ref{cor:cut_knn_sigma_to_r_to_zero} dealing with the $\cut$ and Corollaries~\ref{cor:vol_knn_unweighted}-\ref{cor:vol_knn_r_to_sigma_to_infinity} dealing with the volume are proved.
      Finally, in Theorem~\ref{thm:limit_ncut_cheegercut_knn_graph}
      the convergence of $\ncut$ and $\cheegercut$ are analyzed using the 
      result of these corollaries.
    }
    \label{fig:proof_structure}
  \end{figure}

  In the subsections on the $\cut$-term and the volume term we always present
  the proposition for general weight functions first. Then the lemmas follow 
  that are used in the proof of the proposition. Finally, we show corollaries 
  that apply these general results to the specific weight functions we consider in 
  this paper.
  An overview of the proof structure is given in Figure~\ref{fig:proof_structure}.
  
  \subsubsection{$k$-nearest neighbor radii} \label{sec:knn_radii}
  As we have explained in Section~\ref{sec:proofs_intuitively} the basic ideas of our
  convergence proofs are similar for all the graphs. However, there is one major 
  technical difficulty for the $k$-nearest neighbor graph: The existence of an edge
  between two points depends on all the other sample points and it is random, 
  even if we know the distance between the points. However, each sample 
  point $x_i$ is connected to its $k$ nearest neighbors,
  that means to all points with a distance not greater than that of the 
  $k$-th nearest neighbor. This distance is called the $k$-nearest neighbor radius of 
  point $x_i$. Unfortunately, given a sample point we do not know this radius
  without looking at all the other points. 
  The idea to overcome this difficulty is the following: Given the position of a 
  sample point we give lower and upper bounds on the $\kNN$ radius that depend on
  the density around the point and show that with high probability the true radius 
  is between these bounds. 
  Then we can replace the integration over balls of a fixed radius with the 
  integration over balls with the lower and upper bound on the $\kNN$ radius
  in the proof for the bias term
  and then show that these integrals converge towards each other.
  Furthermore, under our assumptions the radius of all the points can be bounded 
  from above, which helps to bound the influence of far-away points.
  
  In this section we define formally the bounds on the 
  $k$-nearest neighbor radii, since these will be used in the statement 
  of the general proposition. In Lemma~\ref{lem:prob_knn_radii} we state 
  the bounds on the probabilities that the true $\kNN$ radius is between 
  our bounds for the cases we need in the proofs.
  
  We first introduce the upper bound $r_n^{\max}$
  on the maximum $k$-nearest neighbor radius of a point not depending on its position.
  Second, we use that given a point $x$ (far enough) in the interior of $C$ 
  the conditional $\kNN$ radius of a sample point at $x$ is highly concentrated 
  around a radius $r_n (x)$.
  Formally, we define
  \begin{align*}
    r_n^{\max} = \sqrt[d]{\frac{4}{\gamma p_{\min} \eta_d} \frac{k_n}{n-1}} ,\qquad 
    \text{ and } \qquad r_n (x) = \sqrt[d]{\frac{k_n}{(n-1) p(x) \eta_d}} 
    \qquad \text{ for all $x \in C$.}
  \end{align*}
  As to the concentration we state sequences of lower and upper bounds, 
  $r_n^- (x)$ and $r_n^+ (x)$ that converge to $r_n (x)$ such that
  for all $x \in C$ that are not in a small boundary strip the probability
  that a point in $x$ is connected to a point in $y$ becomes small if the 
  distance between $x$ and $y$ exceeds $r_n^+ (x)$ and becomes large if the distance
  is smaller than $r_n^- (x)$.

  Clearly, the accuracy of the bounds depends on how much the density can vary
  around $x$. Setting $\xi_n = 2 p_{\max}' r_n^{\max} / p_{\min}$ the density
  in the ball of radius $2 r_n^{\max}$ around $x$ can vary between 
  $(1-\xi_n) p(x)$ and $(1+\xi_n) p(x)$.
  Furthermore, we have to ``blow up'' or shrink the radii a bit in order to be
  sure that the true $\kNN$ radius is between them. To this end we introduce 
  a sequence $(\delta_n)_{n \in \N}$ with $\delta_n \to 0$ and 
  $\delta_n k_n \to \infty$ for $n \to \infty$.
  Then we can define
  \begin{align*}
    r_n^- (x) = \sqrt[d]{(1-2\xi_n) (1-\delta_n)} r_n (x) 
    \quad \text{and} \quad 
    r_n^+ (x) = \sqrt[d]{(1+2\xi_n) (1+\delta_n)} r_n (x) .
  \end{align*}
  Note that $\xi_n$ converges to zero, since $r_n^{\max}$ converges to zero
  as $\sqrt[d]{k_n/n}$. The sequence $\delta_n$ is chosen such that
  it converges to zero reasonably fast, but that with high probability 
  $r_n^+ (x)$ and $r_n^- (x)$ are bounds on the $\kNN$ radius of a point at $x$.
    
  In order to quantify the probability of connections, which we seek to bound, we 
  define the function $c: \R^d \times \R^d \to [0,1]$ by
  \begin{align*}
    c (x,y) = 
    \begin{cases}
      \Pr \left( C_{12} \; \middle| \; x_1=x, x_2=y \right) 
      & \text{if $x \in C$ and $y \in C$} \\
      0 &\text{otherwise},
    \end{cases}
  \end{align*}
  where $C_{12}$ denotes the event that there is an edge between the sample
  points $x_1$ and $x_2$ in the (directed or undirected) 
  $k$-nearest neighbor graph.

  \subsubsection{The $\cut$ term in the $\kNN$ graph}
  \label{sec:cut_term_knn}

  \begin{proposition} \label{pro:cut_knn_general}
    Let $G_n$ be the directed, symmetric or mutual $k$-nearest neighbor graph
    with a monotonically decreasing weight function $f_n$.
    Set $\delta_n = \sqrt{(8 \delta_0 \log n)/k_n}$ for some 
    $\delta_0 \geq 2$ in the definition of $r_n^-(x)$.
    Then we have for the bias term
    \begin{align*}
      &\left| \E \left( \frac{\cut_n}{n (n-1)} \right) 
      - 2 \int_{S \cap C} p^2 (s) \capintegral^{(1)} \left( r_n (s) \right) \  \mathrm{d}s \right| 
      = O \left( \capintegral^{(1)} \left( r_n^{\max} \right) \sqrt[d]{\frac{k_n}{n}} \right) \\
      &\qquad + O \left( \min \left\{ n^{-\delta_0} f_n \left( \inf_{x \in C} r_n (x) \right),
      \ballintegral^{(1)} (\infty) - \ballintegral^{(1)} \left( \inf_{x \in C} r_n (x) \right) \right\} \right) \\
      &\qquad + O \left( \min \left\{\left( \sqrt[d]{\frac{k_n}{n}} + \sqrt{\frac{\log n}{k_n}} \right) f_n \left( \inf_{x \in C} r_n^- (x) \right) 
      \left( \frac{k_n}{n} \right)^{1+1/d}, 
      \capintegral^{(1)} (\infty) - \capintegral^{(1)} (\inf_{x \in C} r_n^- (x)) 
      \right\} \right)   .
    \end{align*}
    
    Furthermore, we have for the variance term for a suitable constant $\tilde{C}$
    \begin{align*}  
      &\Pr \left( \left| \cut_n - \E \left( \cut_n^{(i)} \right) \right| > \eps \right) 
      \leq 2 \exp \left( - \frac{\tilde{C} \eps^2}{n k_n^2 f_n^2 (0)} \right) .
    \end{align*}
  \end{proposition}

  \begin{proof}{}
    We define for $i,j \in \{1,\ldots,n\}$, $i \neq j$ the random variable 
    $W_{ij}$ as
    \begin{displaymath}
      W_{ij} = 
      \begin{cases}
	f_n ( \dist(x_i, x_j) 
	& \text{if } x_i \in H^+, x_j \in H^- \text{ and } (x_i,x_j) \text{ edge in } G_n \\
	0 & \text{otherwise} .
      \end{cases}	
    \end{displaymath}
    For both, a directed and an undirected graph we have
    \begin{displaymath}
      \cut_{n} = \sum_{i=1}^n \mathop{\sum_{j=1}^n}_{j\neq i} W_{ij} ,
    \end{displaymath}
    and by the linearity of expectation and the fact that
    the points are independent and identically distributed,
    we have
    \begin{displaymath}
      \E \left( \frac{\cut_{n}}{n (n-1)} \right) 
      = \frac{1}{n (n-1)} \sum_{i=1}^n \mathop{\sum_{j=1}^n}_{j\neq i} \E ( W_{ij} ) 
      = \frac{1}{n (n-1)} n (n-1) \E ( W_{12} ) 
      = \E ( W_{12} ) .
    \end{displaymath}
    In the convergence proof for the variance term of the $\cut$ 
    for the $r$-neighborhood graph in Proposition~\ref{pro:lim_cut_general_weight_r_graph}
    we need a bound on $\E ( W_{12}^2)$.
    Since this can be derived similarly to $\E ( W_{12})$ we state the following
    for $\E ( W_{12}^q)$ for $q=1,2$.
   
    We define $C_{12}$ to be the event that the sample points $x_1$ and $x_2$ 
    are connected in the graph.
    Conditioning on the location of the points $x_1 \in C$ and $x_2 \in C$ we obtain
    $W_{12}=0$ if $x_1$ and $x_2$ on the same side of the hyperplane $S$, otherwise
    \begin{align*}
      W_{12} = 
      \begin{cases}
	f_n ( \dist (x_1, x_2) ) & \text{if } C_{12}=1 \\
	0 & \text{otherwise} .
      \end{cases}
    \end{align*}
    Therefore, if $x_1 \in C$ and $x_2 \in C$ are on different sides of $S$
    \begin{align*}
      \E \left( W_{12}^q \; \middle| \; x_1=x, x_2=y \right)
      &= f_n^q ( \dist (x, y) ) \Pr \left( C_{12} \; \middle| \; x_1=x, x_2=y \right) .
    \end{align*}
    With $c(x,y)$ as above we have
    \begin{align*}
      \E ( W_{12}^q ) &= \int_{C} \int_{C} \E ( W_{12}^q \; | \; x_1=x, x_2 = y ) 
      p (y ) \ \mathrm{d}y \ p ( x ) \ \mathrm{d}x \\
      &= \int_{H^+ \cap C} \int_{H^- \cap C} f_n^q ( \dist (x, y) ) 
      \Pr \left( C_{12} \; \middle| \; x_1=x, x_2=y \right)
      p (y ) \ \mathrm{d}y \ p ( x ) \ \mathrm{d}x \\
      & \quad + \int_{H^- \cap C} \int_{H^+ \cap C} f_n^q ( \dist (x, y) ) 
      \Pr \left( C_{12} \; \middle| \; x_1=x, x_2=y \right) p (y ) \ \mathrm{d}y \ p ( x ) \ \mathrm{d} x \\
      &= \int_{H^+ } \int_{H^-} f_n^q ( \dist (x, y) ) 
      c(x,y) p (y ) \ \mathrm{d}y \ p ( x ) \ \mathrm{d}x \\
      & \quad + \int_{H^-} \int_{H^+} f_n^q ( \dist (x, y) ) 
      c (x,y) p (y ) \ \mathrm{d}y \ p ( x ) \ \mathrm{d} x .
    \end{align*}

    Setting
    \begin{align*}
      g (x) = 
      \begin{cases}
	\int_{H^-} f_n^q ( \dist (x, y) ) c(x,y)
	p (y ) \ \mathrm{d}y & \text{if } x \in H^+ \\
	\int_{H^+} f_n^q ( \dist (x, y) ) c(x,y)
	p (y ) \ \mathrm{d}y & \text{if } x \in H^-
      \end{cases}
    \end{align*}
    we obtain
    \begin{align*}
      \E ( W_{12}^q ) &= \int_{\R^d} g(x) p(x) \ \mathrm{d} x 
      = \int_{H^+} g(x) p(x) \ \mathrm{d} x + \int_{H^-} g(x) p(x) \ \mathrm{d} x  .
    \end{align*}

    We only deal with the first integral here, the second can be computed analogously.
    By a simple transformation of the coordinate system we can write this
    integral as an integral along the hyperplane $S$, and for each points 
    $s$ in $S$ we integrate over the normal line through $s$.
    In the following we find lower and upper bounds on the integral
    \begin{align*}
      \int_{S} \int_{0}^{\infty} g (s + tn_S) p \left( s + t n_S \right) 
      \mathrm{d}t \mathrm{d}s 
      = \int_{S} h_n (s) \mathrm{d}s ,
    \end{align*}
    where we have set
    \begin{align*}
      h_n (s) = \int_{0}^\infty g \left( s + t n_S \right) p \left( s + t n_S \right) 
      \ \mathrm{d}t .
    \end{align*}

    We set $\mathcal{I}_n = \{x \in C \; | \; \dist (x, \partial C) \geq 2 r_n^{\max} \}$ 
    and use the following decomposition of the integral
    \begin{align}
      \label{eq:decomposition1}
      &\left| \int_{S} h_n (s) \  \mathrm{d}s 
      - \int_{S} p^2 (s) \capintegral^{(q)} \left( r_n (s) \right) \  \mathrm{d}s \right|
      \leq \left| \int_{S} h_n (s) \  \mathrm{d}s 
      - \int_{S \cap \mathcal{I}_n} h_n (s) \  \mathrm{d}s \right| \\
      \label{eq:decomposition2}
      &\qquad + \left| \int_{S \cap \mathcal{I}_n} h_n (s) \  \mathrm{d}s 
      - \int_{S \cap \mathcal{I}_n} p^2 (s) \capintegral^{(q)} \left( r_n (s) \right) \  \mathrm{d}s \right| \\
      \label{eq:decomposition3}
      &\qquad + \left| \int_{S \cap \mathcal{I}_n} p^2 (s) \capintegral^{(q)} \left( r_n (s) \right) \  \mathrm{d}s 
      - \int_{S \cap C} p^2 (s) \capintegral^{(q)} \left( r_n (s) \right) \  \mathrm{d}s \right| .
    \end{align}

    We first give a bound on the right hand side of Equation~\eqref{eq:decomposition1}. 
    Setting    
    $\mathcal{R}_n = \{ x \in \R^d \; | \; \dist (x, \partial C) < 2 r_n^{\max} \}$ and 
    $\mathcal{A}_n = \R^d \setminus ( \mathcal{I}_n \cup \mathcal{R}_n )$,
    we have (considering that the integrand is positive and $S \cap \mathcal{I}_n \subseteq S$)
    \begin{align*}
      \left| \int_{S} h_n (s) \ \mathrm{d} s 
      - \int_{S \cap \mathcal{I}_n} h_n (s) \ \mathrm{d} s \right|
      = \int_{S \cap \mathcal{R}_n} h_n (s) \ \mathrm{d} s 
      + \int_{S \cap \mathcal{A}_n} h_n (s) \ \mathrm{d} s ,
    \end{align*}
    that is, we have to derive upper bounds on the two integrals on the right hand side.
    
    First let $s \in S \cap \mathcal{A}_n$, that is $s \notin C$ and 
    $\dist(s,C) \geq 2 r_n^{\max}$.
    Consequently $p(s + tn_S)=0$ for $t < 2 r_n^{\max}$.
    On the other hand, if $t \geq 2 r_n^{\max}$ we have $\dist (s+tn_S,y) \geq 2 r_n^{\max}$
    for all $y \in H^-$.
    Setting $c_n = 2 \exp (-k_n / 8 )$ we have with Lemma~\ref{lem:prob_knn_radii}
    $c (s+tn_S,y) \leq c_n$ for all $y \in H^-$.
    Hence
    \begin{align*}
      g (s+tn_S) &\leq \int_{B(s+tn_S,r_n^{\max}) \cap H^-} f_n^q ( \dist (s+tn_S, y) ) 
      c (s+tn_S,y) p (y ) \ \mathrm{d}y \\
      &\quad \quad + \int_{B(s+tn_S,r_n^{\max})^c \cap H^-} f_n^q ( \dist (s+tn_S, y) ) 
      c (s+tn_S,y) p (y ) \ \mathrm{d}y \\
      &\leq f_n^q \left( r_n^{\max} \right) \int_{H^-} c (s+tn_S,y) p (y ) \ \mathrm{d}y
      \leq c_n f_n^q \left( r_n^{\max} \right) ,
    \end{align*}
    since $B(s+tn_S,r_n^{\max}) \cap H^- = \emptyset$ for $t>r_n^{\max}$ and 
    $f_n$ is monotonically decreasing.
    Therefore, for all $s \in S \cap \mathcal{A}_n$
    \begin{align*}
      h_n (s) &= \int_{0}^{\infty} g \left( s + t n_S \right) p \left( s + t n_S \right) 
      \ \mathrm{d}t 
      \leq  \int_{2 r_n^{\max}}^{\infty} g (s+tn_S) p \left( s + t n_S \right) \ \mathrm{d}t \\
      &\leq c_n f_n^q \left( r_n^{\max} \right) 
      \int_{0}^{\infty} p \left( s + t n_S \right) \ \mathrm{d}t ,
    \end{align*}
    and thus
    \begin{align*}
      \int_{S \cap \mathcal{A}_n} h_n (s) \ \mathrm{d} s
      &\leq \int_{S \cap \mathcal{A}_n}  c_n f_n^q \left( r_n^{\max} \right) 
      \int_{0}^{\infty} p \left( s + t n_S \right) \ \mathrm{d}t  \ \mathrm{d} s \\
      &\leq c_n f_n^q \left( r_n^{\max} \right)  \int_{S}  
      \int_{0}^{\infty} p \left( s + t n_S \right) \ \mathrm{d}t  \ \mathrm{d} s 
      \leq c_n f_n^q \left( r_n^{\max} \right) .
    \end{align*}

    Now let $s \in S \cap \mathcal{R}_n$. Then 
    \begin{align*}
      g (s+tn_S) &=\int_{H^-} f_n^q ( \dist (s+tn_S, y) ) c(s+tn_S,y) p (y ) \ \mathrm{d}y \\
      &\leq \int_{B(s+tn_S,r_n^{\max}) \cap H^-} f_n^q ( \dist (s+tn_S, y) ) 
      c (s+tn_S,y) p (y ) \ \mathrm{d}y \\
      &\quad \quad + \int_{B(s+tn_S,r_n^{\max})^c \cap H^-} f_n^q ( \dist (s+tn_S, y) ) 
      c (s+tn_S,y) p (y ) \ \mathrm{d}y \\
      &\leq p_{\max} \int_{B(s+tn_S,r_n^{\max}) \cap H^-} f_n^q ( \dist (s+tn_S, y) ) 
      \ \mathrm{d}y  + c_n f_n^q \left( r_n^{\max} \right) .
    \end{align*}
    Considering that $B(s+tn_S,r_n^{\max}) \cap H^- = \emptyset$ for $t>r_n^{\max}$
    and therefore the first integral vanishes in this case, we have
    for all $s \in S \cap \mathcal{R}_n$
    \begin{align*}
      h_n (s) &= \int_{0}^\infty g \left( s + t n_S \right) p \left( s + t n_S \right) 
      \ \mathrm{d}t \\
      &\leq \int_{0}^{r_n^{\max}} p_{\max} \int_{B(s+tn_S,r_n^{\max}) \cap H^-} 
      f_n^q ( \dist (s+tn_S, y) ) \ \mathrm{d}y \ p \left( s + t n_S \right) \ \mathrm{d}t \\
      &\quad \quad + c_n f_n^q \left( r_n^{\max} \right) \int_{0}^\infty p \left( s + t n_S \right) 
      \ \mathrm{d}t \\
      &\leq p_{\max}^2 \int_{0}^{r_n^{\max}} \int_{B(s+tn_S,r_n^{\max}) \cap H^-} 
      f_n^q ( \dist (s+tn_S, y) ) \ \mathrm{d}y \ \mathrm{d}t \\
      &\quad \quad + c_n f_n^q \left( r_n^{\max} \right) 
      \int_{0}^\infty p \left( s + t n_S \right) \ \mathrm{d}t \\
      &\leq p_{\max}^2 \capintegral^{(q)} \left( r_n^{\max} \right)
      +c_n f_n^q \left( r_n^{\max} \right) \int_{0}^\infty p \left( s + t n_S \right) 
      \ \mathrm{d}t ,
    \end{align*}
    and thus
    \begin{align*}
      \int_{S \cap \mathcal{R}_n} h_n (s) \ \mathrm{d} s
      &\leq \int_{S \cap \mathcal{R}_n} p_{\max}^2 \capintegral^{(q)} \left( r_n^{\max} \right)
      + c_n f_n^q \left( r_n^{\max} \right) \int_{0}^\infty p \left( s + t n_S \right) 
      \ \mathrm{d}t \ \mathrm{d} s \\
      &\leq p_{\max}^2 \capintegral^{(q)} \left( r_n^{\max} \right) 
      \lebesguevol_{d-1} \left( S \cap \mathcal{R}_n \right)
      + c_n f_n^q \left( r_n^{\max} \right) .
    \end{align*}

    For some weight functions, for example the Gaussian, it is preferable to use
    that for all $x \in \R^d$ and all radii $r$ 
    \begin{align*}
      &\int_{B(x,r)^c \cap H^-} f_n^q ( \dist (x, y) ) 
      c (x,y) p (y ) \ \mathrm{d}y 
      \leq p_{\max} \int_{B(x,r)^c} f_n^q ( \dist (x, y) ) \ \mathrm{d}y \\
      &\qquad= p_{\max} \left( \int_{\R^d} f_n^q ( \dist (x, y) ) \ \mathrm{d}y 
      - \int_{B(x,r)} f_n^q ( \dist (x, y) ) \ \mathrm{d}y \right)
      = p_{\max} \left( \ballintegral^{(q)} (\infty) - \ballintegral^{(q)} (r) \right) .
    \end{align*}

    We have according to Lemma~\ref{lem:boundary_strip_to_zero} 
    $\lebesguevol_{d-1} \left( S \cap \mathcal{R}_n \right) = O (r_n^{\max})$.
    Consequently, using $r_n^{\max} = O (\sqrt[d]{k_n/n})$ and plugging in $c_n$
    \begin{align*}
      &\left| \int_{S} h_n (s) \ \mathrm{d} s 
      - \int_{S \cap \mathcal{I}_n} h_n (s) \ \mathrm{d} s \right| \\
      &\qquad = O \left( \capintegral^{(q)} \left( r_n^{\max} \right) \sqrt[d]{\frac{k_n}{n}}
      + \min \left\{ \exp \left( - k_n/8 \right) f_n^q \left( \inf_{x \in C} r_n (x) \right), \left( \ballintegral^{(q)} (\infty) - \ballintegral^{(q)} (r_n^{\max}) \right) \right\} \right) .
    \end{align*}

    Now we consider the term in Equation~\eqref{eq:decomposition2}. 
    In the following, note that with $\xi_n = 2 p_{\max}' r_n^{\max} / p_{\min}$ 
    we have for all $x \in C$ with $B(x,2r_n^{\max}) \subseteq C$ and 
    $y \in B(x,2r_n^{\max})$
    \begin{align*}
      (1-\xi_n) p(x) \leq p(y) \leq (1+\xi_n) p(x) .
    \end{align*}
    We assume that $n$ is sufficiently large such that $\xi_n < 1/2$.

    For any $s \in S \cap \mathcal{I}_n$ and any $t \geq 0$ we have
    \begin{align*}
     g (s+tn_S) &= \int_{H^-} f_n^q ( \dist (s+tn_S, y) ) c(s+tn_S,y) p (y ) \ \mathrm{d}y \\
      & \geq \int_{B(s+tn_S,r_n^- (s)) \cap H^-} f_n^q ( \dist (s+tn_S, y) ) 
      c (s+tn_S,y) p (y ) \ \mathrm{d}y .
    \end{align*}
    If $t>r_n^- (s)$ we use the trivial bound $g(s+tn_S) \geq 0$.
    Otherwise we have with Lemma~\ref{lem:prob_knn_radii} for all $y \in B(s+tn_S,r_n^- (s)) \cap H^-$ 
    that $c(s+tn_S,y) \geq 1-a_n$ with $a_n=6  \exp \left( - \delta_n^2 k_n/3 \right)$.
    Using, furthermore, the bound $p(y) \geq (1-\xi_n) p(s)$ we obtain
    \begin{align*}
      g (s+tn_S) &\geq \int_{B(s+tn_S,r_n^- (s)) \cap H^-} f_n^q ( \dist (s+tn_S, y) ) 
      (1-a_n) (1-\xi_n) p(s) \ \mathrm{d}y \\
      &= (1-a_n) (1-\xi_n) p(s) \int_{B(s+tn_S,r_n^- (s)) \cap H^-} f_n^q ( \dist (s+tn_S, y) ) \ \mathrm{d}y . 
    \end{align*}
     That is, we obtain for $s \in \mathcal{I}_n$
    \begin{align*}
      & h_n (s) = \int_{0}^\infty g \left( s + t n_S \right) p \left( s + t n_S \right) 
      \ \mathrm{d}t 
      \geq \int_{0}^{r_n^- (s)} g \left( s + t n_S \right) p \left( s + t n_S \right) 
      \ \mathrm{d}t  \\
      &\quad \geq (1-\xi_n) p(s) \int_{0}^{r_n^- (s)} g \left( s + t n_S \right) 
      \ \mathrm{d}t \\
      &\quad \geq (1-a_n) (1-\xi_n)^2 p^2(s) \int_{0}^{r_n^- (s)} \int_{B(s+tn_S,r_n^- (s)) \cap H^-} f_n^q ( \dist (s+tn_S, y) ) \ \mathrm{d}y \  \ \mathrm{d}t \\ 
      &\quad \geq (1-a_n) (1-\xi_n)^2 p^2(s) \capintegral^{(q)} \left( r_n^- (s) \right) .
    \end{align*}
    where in the last inequality we have applied Lemma~\ref{lem:cap_integral}.

    Therefore
    \begin{align*}
      &\int_{S \cap \mathcal{I}_n} h_n (s) \ \mathrm{d} s
      \geq (1-a_n) (1-\xi_n)^2 \int_{S \cap \mathcal{I}_n} p^2(s) 
      \capintegral^{(q)} \left( r_n^- (s) \right) \ \mathrm{d} s \\
      &\quad \geq (1-a_n) (1-\xi_n)^2 \int_{S \cap \mathcal{I}_n} p^2(s) 
      \capintegral^{(q)} \left( r_n (s) \right) \ \mathrm{d} s \\
      &\quad \quad - \int_{S \cap \mathcal{I}_n} p^2(s) 
      \left( \capintegral^{(q)} \left( r_n (s) \right) - \capintegral^{(q)} \left( r_n^- (s) \right) \right) \ \mathrm{d} s \\
      &\quad \geq \int_{S \cap \mathcal{I}_n} p^2(s) 
      \capintegral^{(q)} \left( r_n (s) \right) \ \mathrm{d} s 
      - (a_n + \xi_n) \int_{S \cap \mathcal{I}_n} p^2(s) 
      \capintegral^{(q)} \left( r_n (s) \right) \ \mathrm{d} s \\
      &\quad \quad - p_{\max}^2 \int_{S \cap \mathcal{I}_n} 
      \left( \capintegral^{(q)} \left( r_n (s) \right) - \capintegral^{(q)} \left( r_n^- (s) \right) \right) \ \mathrm{d} s ,
    \end{align*}
    and thus
    \begin{align}
      \nonumber &\int_{S \cap \mathcal{I}_n} h_n (s) \ \mathrm{d} s
      - \int_{S \cap \mathcal{I}_n} p^2(s) 
      \capintegral^{(q)} \left( r_n (s) \right) \ \mathrm{d} s \\
      \nonumber &\quad \geq - (a_n + \xi_n) \int_{S \cap \mathcal{I}_n} p^2(s) 
      \capintegral^{(q)} \left( r_n (s) \right) \ \mathrm{d} s \\
      &\quad \quad - p_{\max}^2 \lebesguevol_{d-1} (S \cap C) \sup_{s \in S \cap \mathcal{I}_n} 
      \left( \capintegral^{(q)} \left( r_n^+ (s) \right) - \capintegral^{(q)} \left( r_n (s) \right) \right) .
      \label{eq:second_term_lower_bound}
    \end{align}

    Now, we want to find an upper bound on $g (s+tn_S)$ for $s \in S \cap \mathcal{I}_n$, 
    that is $B(s, 2 r_n^{\max}) \subseteq C$. We use the following decomposition
    \begin{align*}
      g (s+tn_S) &= \int_{H^-} f_n^q ( \dist (s+tn_S, y) ) c(s+tn_S,y) p (y ) \ \mathrm{d}y \\
      & \leq \int_{B(s+tn_S,r_n^+ (s)) \cap H^-} f_n^q ( \dist (s+tn_S, y) ) 
      c (s+tn_S,y) p (y ) \ \mathrm{d}y \\
      &\quad \quad +  \int_{B(s+tn_S,r_n^+ (s))^c \cap H^-} f_n^q ( \dist (s+tn_S, y) ) 
      c (s+tn_S,y) p (y ) \ \mathrm{d}y .
    \end{align*}
    We use in the first term the trivial bound $c(s+tn_S,y) \leq 1$ and 
    in the second term the monotonicity of $f_n$ and the bound $b_n= 6 \exp ( - \delta_n^2 k_n / 4)$ 
    on the probability of connectedness when the distance is greater than $r_n^+ (s)$
    from Lemma~\ref{lem:prob_knn_radii} to obtain
    \begin{align*}
      g (s+tn_S) &\leq \int_{B(s+tn_S,r_n^+ (s)) \cap H^-} f_n^q ( \dist (s+tn_S, y) ) 
      p (y ) \ \mathrm{d}y \\
      &\quad \quad + b_n f_n^q \left( r_n^+ (s) \right) \int_{B(s+tn_S,r_n^+ (s))^c \cap H^-} 
      p (y ) \ \mathrm{d}y \\
&\leq \int_{B(s+tn_S,r_n^+ (s)) \cap H^-} f_n^q ( \dist (s+tn_S, y) ) 
      p (y ) \ \mathrm{d}y + b_n f_n^q \left( r_n^+ (s) \right) .
    \end{align*}
    Using a bound on the density in the balls $B(s+tn_S,r_n^+ (s))$ we obtain
    \begin{align*}
      g (s+tn_S) &\leq
      (1+\xi_n) p(s) \int_{B(s+tn_S,r_n^+ (s)) \cap H^-} f_n^q ( \dist (s+tn_S, y) ) 
      \ \mathrm{d}y + b_n f_n \left( r_n^+ (s) \right) ,
    \end{align*}
    and observe that $g(s+tn_S) \leq b_n f_n^q (r_n^+ (s) )$ if $t > r_n^+ (s)$
    since in this case $B(s+tn_S,r_n^+ (s)) \cap H^- = \emptyset$.
 
    That is,
    \begin{align*}
      h_n (s) &= \int_{0}^\infty g \left( s + t n_S \right) p \left( s + t n_S \right) 
      \ \mathrm{d}t \\
      &\leq \int_{0}^{r_n^+(s)} (1+\xi_n) p(s) \int_{B(s+tn_S,r_n^+ (s)) \cap H^-} f_n^q ( \dist (s+tn_S, y) ) 
      \ \mathrm{d}y  \ p \left( s + t n_S \right) \ \mathrm{d}t \\
      &\quad \quad + \int_{0}^{\infty} b_n f_n^q \left( r_n^+ (s) \right) p \left( s + t n_S \right) 
      \ \mathrm{d}t \\
      &\leq (1+\xi_n)^2 p^2 (s) \int_{0}^{r_n^+(s)} \int_{B(s+tn_S,r_n^+ (s)) \cap H^-} f_n^q ( \dist (s+tn_S, y) ) 
      \ \mathrm{d}y \ \mathrm{d}t \\
      &\quad \quad + b_n f_n^q \left( r_n^+ (s) \right) \int_{0}^{\infty}  p \left( s + t n_S \right) 
      \ \mathrm{d}t \\
      &= (1+\xi_n)^2 p^2 (s) \capintegral^{(q)} \left( r_n^+ (s) \right) 
      + b_n f_n^q \left( r_n^+ (s) \right) \int_{0}^{\infty}  p \left( s + t n_S \right) \ \mathrm{d}t 
    \end{align*}

    Therefore, considering that $\xi_n < 1/2$
    \begin{align*}
      \int_{S \cap \mathcal{I}_n} h_n (s) \ \mathrm{d} s
      &\leq (1+\xi_n)^2 \int_{S \cap \mathcal{I}_n} p^2 (s) \capintegral^{(q)} \left( r_n^+ (s) \right) \ \mathrm{d}s \\
      & \quad + b_n \int_{S \cap \mathcal{I}_n} f_n^q \left( r_n^+ (s) \right) \int_{0}^{\infty}  p \left( s + t n_S \right) \ \mathrm{d}t \ \mathrm{d}s \\
      &\leq (1 + 3 \xi_n) \int_{S \cap \mathcal{I}_n} p^2 (s) \capintegral^{(q)} \left( r_n (s) \right) \ \mathrm{d}s \\
      &\quad + 3 \int_{S \cap \mathcal{I}_n} p^2 (s) \left( \capintegral^{(q)} \left( r_n^+ (s) \right) - \capintegral^{(q)} \left( r_n (s) \right) \right) \ \mathrm{d}s \\
      &\quad + b_n f_n^q \left( \inf_{s \in S \cap C} r_n^+ (s) \right)
    \end{align*}
    Consequently,
    \begin{align}
      \nonumber &\int_{S \cap \mathcal{I}_n} h_n (s) \ \mathrm{d} s
      - \int_{S \cap \mathcal{I}_n} p^2 (s) \capintegral^{(q)} \left( r_n (s) \right) \ \mathrm{d}s \\
      \nonumber &\quad \leq 3 p_{\max}^2 \sup_{s \in S \cap \mathcal{I}_n} 
      \left( \capintegral^{(q)} \left( r_n^+ (s) \right) - \capintegral^{(q)} \left( r_n (s) \right) \right)
      \lebesguevol_{d-1} (S \cap C) \\
      &\quad \quad + 3 \xi_n \int_{S \cap \mathcal{I}_n} p^2 (s) \capintegral^{(q)} \left( r_n (s) \right) \ \mathrm{d}s
      + b_n f_n^q \left( \inf_{s \in S \cap C} r_n^+ (s) \right)
      \label{eq:second_term_upper_bound}
    \end{align}

    Similarly to the remark above we can replace 
    $b_n f_n^q \left( \inf_{s \in S \cap C} r_n^+ (s) \right)$
    by 
    \begin{align*}
      p_{\max} \left( \ballintegral^{(q)} (\infty) - \ballintegral^{(q)} ( \inf_{s \in S \cap C}r_n(s)) \right) ,
    \end{align*}
    which gives a better bound for some weight functions, especially the Gaussian.

    Combining Equation~\eqref{eq:second_term_lower_bound} and 
    Equation~\eqref{eq:second_term_upper_bound},
    using the monotonicity of $\capintegral^{(q)}$ and $f$ we obtain
    \begin{align*}
      &\left| \int_{S \cap \mathcal{I}_n} h_n (s) \ \mathrm{d} s
      - \int_{S \cap \mathcal{I}_n} p^2 (s) \capintegral^{(q)} \left( r_n (s) \right) \ \mathrm{d}s \right| \\
      &\qquad = O \left( \sup_{s \in S \cap \mathcal{I}_n} 
      \left( \capintegral^{(q)} \left( r_n^+ (s) \right) - \capintegral^{(q)} \left( r_n^- (s) \right) \right)
      \right) \\
      &\qquad \qquad + O \left( ( a_n + \xi_n ) \capintegral^{(q)} \left( r_n^{\max} \right)  
      + \min \left\{ b_n f_n^q \left( \inf_{x \in C} r_n (s) \right), 
      \ballintegral^{(q)} (\infty) - \ballintegral^{(q)} ( \inf_{x \in C} r_n(x)) \right\} \right) .
    \end{align*}
    
    We still have to bound the first term. For some weight functions, especially the
    Gaussian, we have
    \begin{align*}
      \sup_{s \in S \cap \mathcal{I}_n} \left( \capintegral^{(q)} \left( r_n^+ (s) \right) 
      - \capintegral^{(q)} \left( r_n^- (s) \right) \right)
      \leq \capintegral^{(q)} \left( \infty \right) - \capintegral^{(q)} \left( \inf_{x \in C} r_n^- (x) \right) .
    \end{align*}

    For the other weight functions we use
    \begin{align*}
      & \capintegral^{(q)} \left( r_n^+ (s) \right) - \capintegral^{(q)} \left( r_n^- (s) \right)
      = \int_{0}^{r_n^+ (s)} u^d f_n^q (u) \ \mathrm{d} u 
      - \int_{0}^{r_n^- (s)} u^d f_n^q (u) \ \mathrm{d} u \\
      &\quad \leq f_n^q \left( r_n^- (s) \right) \int_{r_n^- (s)}^{r_n^+ (s)} u^d \ \mathrm{d} u = \frac{1}{d+1} f_n^q \left( r_n^- (s) \right) \left(  \left( r_n^+ (s) \right)^{d+1}
      - \left( r_n^- (s) \right)^{d+1} \right) \\
      &\quad = \frac{1}{d+1} f_n^q \left( r_n^- (s) \right) r_n^{d+1} (s)
      \left(  \left( \frac{r_n^+ (s)}{r_n (s)} \right)^{d+1} - \left( \frac{r_n^- (s)}{r_n (s)} \right)^{d+1} \right) .
    \end{align*}
    Since, with $\xi_n < 1/2$ and $\delta_n<1$,
    \begin{align*}
      \left( \frac{r_n^+ (s)}{r_n (s)} \right)^{d+1}
      &= \left( \frac{(1+2\xi_n) (1+2 \delta_n) k_n}{(n-1) p(s) \eta_d} \frac{(n-1) p(s) \eta_d}{k_n} \right)^{1+1/d} \\
      &= \left( (1+2\xi_n) (1+2 \delta_n) \right)^{1+1/d}
      \leq 1 + 54 \xi_n + 8 \delta_n
    \end{align*}
    and a similar bound holds for the other quotient
    we have
    \begin{align*}
      \capintegral^{(q)} \left( r_n^+ (s) \right) - \capintegral^{(q)} \left( r_n^- (s) \right)
      &= O \left( ( \xi_n + \delta_n ) f_n^q \left( \inf_{x \in C} r_n^- (x) \right) 
      \left( r_n^{\max} \right)^{d+1} \right).
    \end{align*}

    With our choice of $\delta_n$ we have, considering that $\delta_0 \geq 2$,
    \begin{align*}
      a_n &= 6  \exp \left( - \delta_n^2 k_n/3 \right)
      = 6 \exp \left( - (8 \delta_0 \log n) / 3 \right)
      \leq 6 \exp \left( - 5 \log n \right)
      = 6 / n^5 ,
    \end{align*}
    that is, for $n$ sufficiently large such that $6/n^5 \leq \xi_n$,
    considering that $\xi_n = O(\sqrt[d]{k_n/n}$ and plugging in $b_n$ we have
    \begin{align*}
      &\left| \int_{S \cap \mathcal{I}_n} h_n (s) \ \mathrm{d} s
      - \int_{S \cap \mathcal{I}_n} p^2 (s) \capintegral^{(q)} \left( r_n (s) \right) \ \mathrm{d}s \right| \\
      &\qquad = O \left( \min \left\{ \left( \sqrt[d]{\frac{k_n}{n}} + \delta_n \right) f_n^q \left( \inf_{x \in C} r_n^- (x) \right) 
      \left( r_n^{\max} \right)^{d+1}, \capintegral^{(q)} \left( \infty \right) - \capintegral^{(q)} \left( \inf_{x \in C} r_n^- (x) \right) \right\} \right) \\ 
      &\qquad \quad + O \left( \sqrt[d]{\frac{k_n}{n}} \capintegral^{(q)} \left( r_n^{\max} \right)  
	+ \min \left\{ \exp \left( - \delta_n^2 \frac{k_n}{4} \right) f_n^q \left( \inf_{x \in C} r_n (s) \right), 
	\ballintegral^{(q)} (\infty) - \ballintegral^{(q)} ( \inf_{x \in C} r_n(x)) \right\} \right)     
    \end{align*}

    Finally, we bound the term in Equation~\eqref{eq:decomposition3}.
    Setting $\mathcal{R}_n' = C \setminus \mathcal{I}_n$ we have
    \begin{align*}
      &\left| \int_{S \cap \mathcal{I}_n} p^2 (s) \capintegral^{(q)} \left( r_n (s) \right) \  \mathrm{d}s 
      - \int_{S \cap C} p^2 (s) \capintegral^{(q)} \left( r_n (s) \right) \  \mathrm{d}s \right| 
      = \int_{S \cap \mathcal{R}_n'} p^2 (s) \capintegral^{(q)} \left( r_n (s) \right) \  \mathrm{d}s  \\
      &\qquad \leq p_{\max}^2  \capintegral^{(q)} \left( \max_{x \in C} r_n (x) \right) 
      \lebesguevol_{d-1} \left( S \cap \mathcal{R}_n' \right) 
      \leq p_{\max}^2  \capintegral^{(q)} \left( \max_{x \in C} r_n (x) \right) 
      \lebesguevol_{d-1} \left( S \cap \mathcal{R}_n \right) .
    \end{align*}
    Using Lemma~\ref{lem:boundary_strip_to_zero} we have 
    $\lebesguevol_{d-1} \left( S \cap \mathcal{R}_n \right) = O (r_n^{\max})$, 
    and thus
    \begin{align*}
      &\left| \int_{S \cap \mathcal{I}_n} p^2 (s) \capintegral^{(q)} \left( r_n (s) \right) \  \mathrm{d}s 
      - \int_{S \cap C} p^2 (s) \capintegral^{(q)} \left( r_n (s) \right) \  \mathrm{d}s \right| 
      = O \left( \capintegral^{(q)} \left( \max_{x \in C} r_n (x) \right) \sqrt[d]{\frac{k_n}{n}} \right).
    \end{align*}

    Deriving the same bounds for the other halfspace and collecting the three bounds 
    we obtain the result, considering that $k_n/8 \geq \delta_n^2 k_n /8$,
    $\delta_n^2 k_n /4 \geq \delta_n^2 k_n /8$
    and $r_n^{\max} \geq \max_{x \in C} r_n (x)$ due to the monotonicity of 
    $\capintegral^{(1)}$.

    Finally, we discuss the choice of $\delta_n$.
    With this choice of $\delta_n$ we have  
    $\exp \left( - \delta_n^2 k_n/8 \right) = n^{-\delta_0}$.
    Note that this is the fastest convergence rate of $\delta_n$ 
    for which the exponential term converges polynomially in $1/n$, 
    which we will need in the proof of the following corollaries.
    In all the other terms above $\delta_n$ has to be chosen as small as
    possible, so this is the best convergence rate for $\delta_n$.
    Note further that for this choice of $\delta_n$ we require $k_n / \log n \to \infty$,
    since $\delta_n$ has to converge to zero.

    Now we proof the bound for the variance term.
    According to Corollary~3.2.3 from~\citet{Miller/Teng:1997} the maximum degree of 
    the symmetric $k_n$-nearest neighbor graph is bounded by $(\tau_d +1 ) k_n$,
    where $\tau_d$ denotes the kissing number in dimension $d$,
    that is, the maximum number of unit hypershpheres that touch another
    unit hypersphere without any intersections.

    Thus, removing a point from the graph and inserting it in a different place
    the number of (undirected) edges in the $\cut$ can change by at most $2 (\tau_d + 1)$. 
    Since we count undirected edges twice we obtain for all types of $k$-nearest
    neighbor graphs
    \begin{align*}
      \left| \cut_n - \cut_n^{(i)} \right| \leq 4 (\tau_d +1) k_n f_n (0) ,
    \end{align*}
    where $\cut_n^{(i)}$ denotes the value of the $\cut$ in a graph where exactly 
    one point has been moved to a different place.
    Thus by McDiarmid's inequality for a suitable constant $\tilde{C}>0$
    \begin{align*}
      \Pr \left( \left| \cut_n - \E \left( \cut_n^{(i)} \right) \right| > \eps \right) 
      \leq 2 \exp \left( - \frac{2 \eps^2}{n \left( 4 (\tau_d+1) k_n f_n (0) \right)^2 } \right)
      = 2 \exp \left( - \frac{\tilde{C} \eps^2}{n k_n^2 f_n^2 (0)} \right) .
    \end{align*}
  \end{proof}

  The following lemma states bounds on $c (x,y)$, that is the probability of 
  edges between points at $x$ and $y$, in the cases that we need in the convergence proofs
  for the $\cut$ and the volume.
  \begin{lemma}[$\kNN$ radii] \label{lem:prob_knn_radii}
    Let $G_n$ be the directed, mutual or symmetric $k_n$-nearest neighbor graph.
    Let $k_n/n$ be sufficiently small such that $r_n^{\max} \leq r_{\gamma}$.
    Then, if $x,y \in \R^d$ and $\dist (x,y) \geq r_n^{\max}$ we have
    $c (x,y) \leq 2 \exp \left( - k_n/8 \right)$.

    Set $\xi_n = 2 p_{\max}' r_n^{\max} / p_{\min}$ and 
    define $\mathcal{I}_n = \{ s \in C \; | \; B(s, 2 r_n^{\max}) \subseteq C \}$.
    Let $n$ be sufficiently large such that $\xi_n < 1/2$ and
    let $\delta_n \in (0,1)$ with $\delta_n \to 0$ for $n \to \infty$ and
    $k_n \delta_n > 1$ for sufficiently large $n$. 
      
    Let $x = s + t n_S$ with $s \in \mathcal{I}_n \cap S$. 
    If $t \in \R_{\geq 0}$ and $y \in H^-$  or $t \in \R_{\leq 0}$ and $y \in H^+$,
    and, furthermore, $\dist(x,y) \geq r_n^+ (s)$ then  
      $c (x,y) \leq 6 \exp \left( - \delta_n^2 k_n/4 \right)$.
    The same holds for $x \in \mathcal{I}_n$ and $y \in C$ with 
    $\dist(x,y) \geq r_n^+ (x)$.

    Let $x = s + t n_S$ with $t \in [0,r_n^- (s)]$ and 
    $y \in H^-$ or $t \in [-r_n^- (s),0]$ and $y \in H^+$.
    If $\dist(x,y) \leq r_n^- (s)$ then
    $c(x,y) \geq 1 - 6  \exp \left( - \delta_n^2 k_n/3 \right)$.
    The same holds for $x \in \mathcal{I}_n$ and $y \in C$ with 
    $\dist(x,y) \leq r_n^- (x)$.
  \end{lemma}
   
  \begin{proof}{}
    We first show bounds on the probability of connectedness
    for the directed $k$-nearest neighbor graph. 
    These are used in the second part of this proof in order to show
    bounds for the undirected graph as well.    
    Let $D_{ij}$ denote the event that there exists an edge between $x_i$ and $x_j$ 
    in the directed $k$-nearest neighbor graph.

    First we show the statement concerning the maximal $k$-nearest neighbor radius.
    For any $x \in C$ we have
    \begin{align*}
      \mu (B(x, r_n^{\max}) 
      &=\mu \left( B \left( x, \sqrt[d]{\frac{4}{\gamma p_{\min} \eta_d} \frac{k_n}{n-1}} \right) \right) 
      \geq p_{\min} \lebesguevol_d \left( B \left( x, \sqrt[d]{\frac{4}{\gamma p_{\min} \eta_d} \frac{k_n}{n-1}} \right) \cap C \right) \\
      &\geq p_{\min} \gamma \lebesguevol_d \left( B \left( x, \sqrt[d]{\frac{4}{\gamma p_{\min} \eta_d} \frac{k_n}{n-1}} \right) \right) 
      = p_{\min} \gamma \frac{4}{\gamma p_{\min} \eta_d} \frac{k_n}{n-1} \eta_d 
      = \frac{4 k_n}{n-1} .
    \end{align*}
    
    Now suppose we fix $x_1$ and $x_2$ with $\dist (x_1,x_2) \geq r_n^{\max}$.
    If $U$ denotes the random variable that counts the number of points 
    $x_3, \ldots ,x_n$ in $B(x_1,r_n^{\max})$
    we have $U \sim \Bin (n-2,\mu (B(x_1, r_n^{\max}) ))$. 
    Setting $V \sim \Bin (n-2, 4 k_n / (n-1))$, we certainly have
    $0 < k_n/(n-2) < 4 k_n/(n-1)$ for $n \geq 3$ 
    and thus we obtain with a tail bound for the binomial distribution from 
    \citet{Srivastav/Stangier:1996}, which was first proved 
    in~\citet{Angluin/Valiant:1979}, 
    \begin{align*}
      \Pr \left( D_{12} \right) 
      &= \Pr \left( U < k_n \right) \leq \Pr \left( V < k_n \right) 
      \leq \exp \left( - \frac{1}{2} \frac{\left( (n-2) \frac{4 k_n}{n-1} - k_n \right)^2}{(n-2)\frac{4 k_n}{n-1}} \right) %
      \leq \exp \left( - \frac{k_n}{8} \right) . 
    \end{align*}

    In the following we show the statements concerning the upper bound 
    $r_n^+ (s)$ on the $k$-nearest neighbor radii of points
    in regions of relatively homogeneous density.
    The proof for the lower bound $r_n^- (s)$ is similar and is therefore omitted.
    Note, however, that the technical condition $\delta_n k_n > 1$ is needed for this case.
    
    First we show how we can bound the density in the balls $B(s, 2 r_n^{\max})$:
    For any $z \in B(s, 2 r_n^{\max})$ we have by Taylor's theorem 
    \begin{align*}
      p(s) - 2 p_{\max}' r_n^{\max} \leq p(y) \leq p(s) + 2 p_{\max}' r_n^{\max} ,
    \end{align*}
    and thus, with $\xi_n = 2 p_{\max}' r_n^{\max} / p_{\min}$,
    \begin{align*}
      ( 1 - \xi_n ) p(s) \leq p(y) \leq ( 1 + \xi_n ) p(s) .
    \end{align*}
    These bounds are used below to bound the probability mass of balls 
    within $B(s, 2 r_n^{\max})$.

    Now, we bound the probability mass in $B(x,\dist(x,y))$ and $B(y,\dist(x,y))$
    from below, when $\dist (x,y) \geq r_n^+ (s)$.
    We first observe that
    \begin{align*}
      r_n^+ (s) = \sqrt[d]{\frac{(1+2\xi_n) (1+\delta_n) k_n}{(n-1) p(s) \eta_d}} 
      \leq \sqrt[d]{\frac{4 k_n}{(n-1) \gamma p_{\min} \eta_d}} = r_n^{\max} .
    \end{align*}

    Suppose $t = \dist(x,s) \leq r_n^+ (s)$. Then 
    \begin{align*}
      \mu \left( B(x, \dist(x,y)) \right)
      &\geq \mu \left( B(x, r_n^+ (s)) \right) 
    \end{align*}
    with $B(x,r_n^+ (s)) \subseteq B(s,2 r_n^{\max})$.
    If $t = \dist(x,s) > r_n^+ (s)$ we know that $\dist(x,y)>\dist(x,s)$, since 
    $x$ and $y$ are on different sides of the hyperplane $S$.
    We set $x'=s + r_n^+ (s) n_S$, that is the point on the line connecting
    $s$ and $x$ with distance $r_n^+ (s)$ from $s$. Then, 
    by construction, $B(x', r_n^+ (s) ) \subseteq B(s,2 r_n^{\max})$
    and $B(x', r_n^+ (s)) \subseteq B(x, \dist(x,s))$. Thus
    \begin{align*}
      \mu \left( B(x, \dist(x,y)) \right)
      &\geq \mu \left( B(x, \dist(x,s)) \right)
      \geq \mu \left( B(x', r_n^+ (s) ) \right) .
    \end{align*}

    Now we consider balls around the other point $y$. 
    First, suppose $\dist(y,s) = r_n^+ (s)$. 
    Then
    \begin{align*}
      \mu \left( B(y, \dist(x,y)) \right)
      &\geq \mu \left( B(y, r_n^+ (s)) \right) 
    \end{align*}
    with $B(y, r_n^+ (s)) \subseteq B (s, 2 r_n^{\max})$.

    If $\dist(y,s) > r_n^+ (s)$  we set $y'=s + (y-s)/\norm{y-s}$, 
    that is the point on the line connecting
    $s$ and $y$ with distance $r_n^+ (s)$ from $s$. 
    Then, by construction, $B(y', r_n^+ (s) ) \subseteq B(s,2 r_n^{\max})$
    and $B(y', r_n^+ (s)) \subseteq B(y, \dist(y,s))$. 
    Since $x$ and $y$ are on different sides of $S$ we have $\dist(y,s)) \leq \dist(y,x)$.
    Therefore
    \begin{align*}
      \mu \left( B(y, \dist(y,x)) \right)
      & \geq \mu \left( B(y, \dist(y,s)) \right)
      \geq \mu \left( B(y', r_n^+ (s)) \right) .      
    \end{align*}

    We show how to bound $\mu (B(x, r_n^+ (s)))$. The same bound can be shown for 
    the probability mass in $B(x', r_n^+ (s) )$, $B(y, r_n^+ (s))$ and $B(y', r_n^+ (s))$,
    since all of these balls lie in $B (s, 2 r_n^{\max})$.
    We have, since $\xi_n < 1/2$,
    \begin{align*}
      \mu \left(  B(x, r_n^+ (s)) \right) 
      &\geq (1-\xi_n) p(s) \eta_d \left( r_n^+ (s)) \right)^d
      = (1-\xi_n) p(s) \eta_d \frac{(1+2\xi_n) (1+\delta_n) k_n}{(n-1) p(s) \eta_d} \\
      &= (1-\xi_n) (1+2\xi_n) (1+\delta_n) \frac{k_n}{n-1}
      \geq (1+\delta_n) \frac{k_n}{n-1} .
    \end{align*}

    Let $U_x^+ \sim \Bin \left( n-2, \mu \left(  B(x,r_n^+ (s)) \right) \right)$
    and $V_x^+ \sim \Bin \left( n-2, (1+\delta_n) k_n / (n-1) \right)$.
    Then, we have for $(n-2) \delta_n > 1$ 
    \begin{align*}
      0 \leq \frac{k_n}{n-2}  =  \left( 1+\frac{1}{n-2} \right) \frac{k_n}{n-1} < (1+\delta_n) \frac{k_n}{n-1}
    \end{align*}
    and thus, by the tail bound from~\citet{Angluin/Valiant:1979}, 
    \begin{align*}
      \Pr (D_{12} ) &= \Pr \left( U_x^+ < k \right) 
      \leq \Pr \left( V_x^+ < k \right)
      \leq \exp \left( - \frac{1}{2} \frac{\left( (n-2) (1+\delta_n) \frac{k_n}{n-1} - k_n \right)^2}{(n-2) (1+\delta_n) \frac{k_n}{n-1}} \right) .
    \end{align*}
    We have 
    \begin{align*}
      &\left( (n-2) (1+\delta_n) \frac{k_n}{n-1} - k_n \right)^2
      = \left( \left( 1 - \frac{1}{n-1} \right) (1+\delta_n) k_n - k_n \right)^2 \\
      &\quad \quad = \left( \delta_n k_n - \frac{1+\delta_n}{n-1} k_n \right)^2 
      \geq \delta_n^2 k_n^2 - 2 \delta_n (1+\delta_n) \frac{k_n}{n-1} k_n
      \geq \delta_n^2 k_n^2 - 4 \delta_n k_n 
    \end{align*}
    and 
    \begin{align*}
      (n-2) (1+\delta_n) \frac{k_n}{n-1} 
      &= \left( 1- \frac{1}{n-1} \right) (1+\delta_n) k_n
      \leq 2 k_n ,
    \end{align*}
    and thus, using $\delta_n < 1$,
    \begin{align*}
      \Pr (D_{12} ) 
      &\leq \exp \left( - \frac{\delta_n^2 k_n^2 - 4 \delta_n k_n}{4 k_n} \right)
      \leq \exp \left( - \frac{\delta_n^2 k_n}{4} + \delta_n \right) 
      \leq 3 \exp \left( - \frac{\delta_n^2 k_n}{4} \right) .
    \end{align*}
    This analysis can be carried over to the case $t>r_n^+ (s))$
    and the same bound holds.
    
    The same bound holds also for $\Pr (D_{21} )$, since the same bounds for the 
    probability mass in the balls $B(y,r_n^+ (s))$ and $B(y', r_n^+ (s))$ hold.

    In the final step of the proof we use the results derived so far to
    show the results for the undirected $k$-nearest neighbor graphs.
    For the mutual $\kNN$ graph we have by definition
    $\Pr \left( C_{12} \right) =  \Pr \left( C_{21} \right)
    = \Pr \left( D_{12} \cap D_{21} \right)$.
    Thus, clearly, $\Pr \left( C_{12} \right) \leq \Pr \left( D_{12} \right)$
    and 
    \begin{align*}
      \Pr \left( C_{12} \right)
      &= \Pr \left( D_{12} \cap D_{21} \right)
      = 1 - \Pr \left( D_{12}^c \cup D_{21}^c \right)
      \geq 1 - \Pr \left( D_{12}^c \right) - \Pr \left( D_{21}^c \right) \\
      &= 1 - \left( 1 - \Pr \left( D_{12} \right) \right) -  
      \left( 1 - \Pr \left( D_{21} \right) \right)
      = \Pr \left( D_{12} \right) + \Pr \left( D_{21} \right) - 1 .
    \end{align*} 
    This implies
    \begin{multline*}
      \Pr \left( D_{12} \; \middle| \; x_1=x, x_2=y \right) 
      + \Pr \left( D_{21} \; \middle| \; x_1=x, x_2=y \right) - 1 \\
      \leq \Pr \left( C_{12} \; \middle| \; x_1=x, x_2=y \right) 
      \leq \Pr \left( D_{12} \; \middle| \; x_1=x, x_2=y \right) .
    \end{multline*}

    For the symmetric $\kNN$ graph we have 
    $\Pr \left( C_{12} \right) =  \Pr \left( C_{21} \right)
    = \Pr \left( D_{12} \cup D_{21} \right)$,
    which implies $\Pr \left( C_{12} \right) \geq \Pr \left( D_{12} \right)$
    and by a union bound $\Pr \left( C_{12} \right) 
    \leq \Pr \left( D_{12} \right) + \Pr \left( D_{21} \right)$.
    Therefore 
    \begin{multline*}
      \Pr \left( D_{12} \; \middle| \; x_1=x, x_2=y \right)
      \leq \Pr \left( C_{12} \; \middle| \; x_1=x, x_2=y \right) \\
      \leq \Pr \left( D_{12} \; \middle| \; x_1=x, x_2=y \right) 
      + \Pr \left( D_{21} \; \middle| \; x_1=x, x_2=y \right) .
    \end{multline*}

    Thus, using the worse out of the two possible bounds we obtain 
    for both undirected $\kNN$ graph types
    \begin{multline*}
      \Pr \left( D_{12} \; \middle| \; x_1=x, x_2=y \right) 
      + \Pr \left( D_{21} \; \middle| \; x_1=x, x_2=y \right) - 1
      \leq \Pr \left( C_{12} \; \middle| \; x_1=x, x_2=y \right) \\
      \leq \Pr \left( D_{12} \; \middle| \; x_1=x, x_2=y \right) 
      + \Pr \left( D_{21} \; \middle| \; x_1=x, x_2=y \right) .
    \end{multline*}

    Plugging in the results for $\Pr ( D_{12})$ and $\Pr ( D_{21} )$
    in the cases studied above, we obtain the result.

  \end{proof}

  \begin{lemma}{(Integral over caps)} \label{lem:cap_integral}
    Let the general assumptions hold and let $f:\R_{\geq 0} \to \R_{\geq 0}$ 
    be a monotonically decreasing function and $s \in S$.
    Then we have for any $R \in \R_{>0}$
    \begin{align*}
      \int_{0}^{R} \int_{B(s+t n_S, R) \cap H^-} f ( \dist (s+tn_S, y) ) \ \mathrm{d}y \ \mathrm{d}t
      = \eta_{d-1} \int_{u=0}^R u^{d} f(u) \ \mathrm{d} u
    \end{align*}
    and
    \begin{align*}
      \int_{-R}^{0} \int_{B(s+tn_S, R) \cap H^-} f ( \dist (s+tn_S, y) ) \ \mathrm{d}y \ \mathrm{d}t
      = \eta_{d-1} \int_{u=0}^R u^{d} f(u) \ \mathrm{d} u
    \end{align*}
  \end{lemma}
  
  \begin{proof}{}
    By a translation and rotation of our coordinate system in $\R^d$
    such that $s+tn_S$ is the origin and $-n_S$ the first coordinate axis
    we obtain for $t \geq 0$
    \begin{align*}
      &\int_{B(s+t n_S, R) \cap H^-} f ( \dist (s+tn_S, y) ) \ \mathrm{d}y
      = \int_{B(0, R) \cap \{z_1 \geq t\}} f ( \dist (0, z) ) \ \mathrm{d} z \\
      &\quad \quad = \int_{z_1=t}^R \int_{\{z_2^2+ \ldots + z_d^2 \leq R^2-z_1^2\}}
      f ( \dist (0, z) ) \ \mathrm{d} z_d \ldots \ \mathrm{d} z_2 
      \ \mathrm{d} z_1 \\
      &\quad \quad = \int_{z_1=t}^R \int_{\{z_2^2+ \ldots + z_d^2 \leq R^2-z_1^2\}}
      f \left( \sqrt{z_1^2 + \ldots + z_d^2} \right) 
      \ \mathrm{d} z_d \ldots \ \mathrm{d} z_2 \ \mathrm{d} z_1 \\
      &\quad \quad = \int_{z_1=t}^R A(z_1) \ \mathrm{d} z_1 ,
    \end{align*}
    where we have set
    \begin{align*}
      A (r) = \int_{\{z_2^2+ \ldots +z_{d}^2 \leq R^2-r^2\}}
      f \left( \sqrt{r^2 + z_2^2 + \ldots + z_{d}^2} \right)
      \mathrm{d} z_{d} \ldots \mathrm{d} z_{2} .
    \end{align*}
    Thus,
    \begin{align*}
      &\int_{t=0}^{R} \int_{B(s+t n_S, R) \cap H^-} f ( \dist (s+tn_S, y) ) 
      \ \mathrm{d}y \ \mathrm{d}t
      = \int_{t=0}^{R} \int_{r=t}^R A ( r ) \ \mathrm{d}r \ \mathrm{d}t \\
      &\quad \quad = \int_{r=0}^R \int_{t=0}^r A(r) \  \mathrm{d} t \ \mathrm{d} r 
      = \int_{r=0}^R  A(r) \int_{t=0}^r \ \mathrm{d} t \ \mathrm{d} r 
      = \int_{r=0}^R  r A(r) \ \mathrm{d} r 
    \end{align*}

    Similarly, by the same translation and a rotation such that $n_S$ 
    is the first coordinate axis we obtain for $t<0$
    \begin{align*}
      \int_{B(s+t n_S, R) \cap H^+} f ( \dist (s+tn_S, y) ) \ \mathrm{d}y
      &= \int_{B(0, R) \cap \{z_1 \geq -t\}} f ( \dist (0, z) ) \ \mathrm{d} z \\
      &= \int_{z_1=-t}^R A(z_1) \ \mathrm{d} z_1 ,
    \end{align*}
    that is, 
    \begin{align*}
      &\int_{-R}^{0} \int_{B(s+tn_S, R) \cap H^-} f ( \dist (s+tn_S, y) ) 
      \ \mathrm{d}y \ \mathrm{d}t
      = \int_{t=-R}^{0} \int_{r=-t}^R A (r) \ \mathrm{d}r  \ \mathrm{d}t\\
      &\quad \quad = \int_{r=0}^R \int_{t=-r}^0 A(r) 
      \ \mathrm{d}t \ \mathrm{d}r
      = \int_{r=0}^R  A(r) \int_{t=-r}^0 \ \mathrm{d}t \ \mathrm{d}r
      = \int_{r=0}^R  r A(r) \ \mathrm{d}r .
    \end{align*}
    
    Therefore, both the integrals we want to compute are equal to
    $\int_{r=0}^R  r A(r) \ \mathrm{d}r$ which we will treat in
    the following. First we are going to compute the $(d-1)$-dimensional 
    integral $A(r)$. Setting $\tilde{f}_r ( s ) = f ( \sqrt{r^2 + s^2} )$
    we can write $A(r)$ as the following integral in $\R^{d-1}$:
    \begin{align*}
      A (r) 
      &= \int_{\{x_1^2+ \ldots +x_{d-1}^2 \leq R^2-r^2\}}
      f \left( \sqrt{r^2 + x_1^2 + \ldots + x_{d-1}^2} \right)
      \mathrm{d} x_{d-1} \ldots \mathrm{d} x_{1} \\
      &= \int_{\norm{x} \leq \sqrt{R^2-r^2}} \tilde{f}_r (\norm{x}) 
      \ \mathrm{d} x
      = \int_{0}^{\sqrt{R^2-r^2}} (d-1) \eta_{d-1} s^{d-2}  \tilde{f}_r ( s ) 
      \ \mathrm{d} s \\
      &= (d-1) \eta_{d-1} \int_{0}^{\sqrt{R^2-r^2}} s^{d-2} 
      f \left( \sqrt{r^2 + s^2} \right) \ \mathrm{d} s .
    \end{align*}

    Plugging in this expression for $A(r)$ we obtain
    \begin{align*}
      \int_{r=0}^R r A(r) \ \mathrm{d}r 
      = (d-1) \eta_{d-1} \int_{r=0}^R  \int_{s=0}^{\sqrt{R^2-r^2}} r s^{d-2}  
      f \left( \sqrt{r^2 + s^2} \right) \ \mathrm{d} s \ \mathrm{d} r .
    \end{align*}
    
    Substituting with polar coordinates $(r,s)=(u \cos \theta, u \sin \theta)$ 
    with $u \in [0,R]$ and $\theta \in [0,\pi/2]$, we have 
    \begin{align*}
      &\int_{r=0}^R  \int_{s=0}^{\sqrt{R^2-r^2}} r s^{d-2}  
      f \left( \sqrt{r^2 + s^2} \right) \ \mathrm{d} s \ \mathrm{d} r \\
      &\quad \quad = \int_{u=0}^R  \int_{\theta=0}^{\pi/2}
      u \cos \theta u^{d-2} \sin^{d-2} \theta f(u) u 
      \ \mathrm{d} \theta \ \mathrm{d} u \\
      &\quad \quad = \int_{u=0}^R u^{d} f(u)
      \int_{\theta=0}^{\pi/2} \cos \theta \sin^{d-2} \theta  
      \ \mathrm{d} \theta \ \mathrm{d} u \\
      &\quad \quad = \int_{u=0}^R u^{d} f(u)
      \left[ \frac{1}{d-1} \sin^{d-1} \theta \right]_{\theta=0}^{\pi/2}
      \ \mathrm{d} u 
      = \frac{1}{d-1} \int_{u=0}^R u^{d} f(u) \ \mathrm{d} u 
    \end{align*}
    
    Combining the last two equations we obtain
    \begin{align*}
      &\int_{r=0}^R  r A(r) \ \mathrm{d} r 
      = \eta_{d-1} \int_{u=0}^R u^{d} f(u) \ \mathrm{d} u .
    \end{align*}
    
    Note that the integral exists due to the monotonicity of $f$ 
    and the compactness of the interval $[0,R]$.
  \end{proof}

  \begin{corollary}{(Unweighted $\kNN$-graph)} 
    \label{cor:cut_knn_unweighted}
    Let $G_n$ be the unweighted $k$-nearest neighbor graph and 
    let $f_n$ be the unit weight function. %
    Then
    \begin{align*}
      &\left| \frac{1}{n k_n} \sqrt[d]{\frac{n}{k_n}} \cut_n
      - \frac{2 \eta_{d-1}}{(d+1) \eta_d^{1+1/d}} \int_{S} p^{1-1/d} (s) \ \mathrm{d} s \right| 
      = O \left( \sqrt[d]{\frac{k_n}{n}} + \sqrt{\frac{\log n}{k_n}} \right) 
    \end{align*}
    and, for a suitable constant $\tilde{C}>0$
    \begin{align*}
      &\Pr \left( \left| \frac{1}{n k_n} \sqrt[d]{\frac{n}{k_n}} \cut_n 
      - \E \left( \frac{1}{n k_n} \sqrt[d]{\frac{n}{k_n}} \cut_n\right) \right| > \eps \right)
      \leq 2 \exp \left( - \tilde{C} \eps^2 n^{1-2/d} k_n^{2/d} \right) .
    \end{align*}
  \end{corollary}

  \begin{proof}{}
    With Lemma~\ref{pro:integration_unit_weight_function}
    we have for any $s \in S \cap C$, plugging in the definition of $r_n (s)$, 
    \begin{align*}
      \capintegral^{(1)} \left( r_n (s) \right)  
      &= \frac{\eta_{d-1}}{d+1} \left( \frac{k_n}{(n-1) p(s) \eta_d} \right)^{1+1/d} 
      = \frac{\eta_{d-1}}{(d+1) \eta_d^{1+1/d}} \left( \frac{k_n}{n-1} \right)^{1+1/d}
      p^{-1-1/d} (s) .
    \end{align*}    
    
    Therefore,
    \begin{align*}
      &2 \int_{S \cap C} p^2 (s) \capintegral^{(1)} \left( r_n (s) \right) \  \mathrm{d}s
      = 2 \int_{S \cap C} p^2 (s) \frac{\eta_{d-1}}{(d+1) \eta_d^{1+1/d}} 
      \left( \frac{k_n}{n-1} \right)^{1+1/d} p^{-1-1/d} (s) \ \mathrm{d} s \\
      &\qquad = \left( \frac{k_n}{n-1} \right)^{1+1/d} \frac{2 \eta_{d-1}}{(d+1) \eta_d^{1+1/d}}
      \int_{S} p^{1-1/d} (s) \ \mathrm{d} s .
    \end{align*}
    Multiplying this term with the factor $(k_n/(n-1))^{-1-1/d}$ we obtain a constant
    limit. We now multiply the inequality for the bias term in 
    Proposition~\ref{pro:cut_knn_general} with this factor and deal with the error terms.
    
    For the first on we derive an upper bound on $\capintegral^{(1)} ( r_n^{\max} )$ 
    similarly to above and obtain
    \begin{align*}
      &\left( \frac{k_n}{n-1} \right)^{-1-1/d} \capintegral^{(1)} \left( r_n^{\max} \right) 
      \sqrt[d]{\frac{k_n}{n}} 
      = O \left( \sqrt[d]{\frac{k_n}{n}} \right) .      
    \end{align*}
    
    For the second error term we have with $\delta_0=3$ and $f_n \equiv 1$
    \begin{align*}
      &\left( \frac{k_n}{n-1} \right)^{-1-1/d} 
      n^{-\delta_0} f_n \left( \inf_{x \in C} r_n (x) \right)
      \leq n^2 n^{-3} = O \left( n^{-1} \right) .
    \end{align*}

    For the last error term we have
    \begin{align*}
      &\left( \frac{k_n}{n-1} \right)^{-1-1/d} \left( \sqrt[d]{\frac{k_n}{n}} + \sqrt{\frac{\log n}{k_n}} \right) 
      f_n \left( \inf_{x \in C} r_n^- (x) \right) \left( \frac{k_n}{n} \right)^{1+1/d} 
      = O \left( \sqrt[d]{\frac{k_n}{n}} + \sqrt{\frac{\log n}{k_n}} \right) .
    \end{align*}
    
    Thus, considering that $n^{-1} \leq \sqrt[d]{k_n/n}$, we obtain
    \begin{align*}
      &\left| \frac{1}{n k_n} \sqrt[d]{\frac{n-1}{k_n}} \cut_n
      - \frac{2 \eta_{d-1}}{(d+1) \eta_d^{1+1/d}} \int_{S} p^{1-1/d} (s) \ \mathrm{d} s \right| \\
      &\qquad = \left( \frac{n-1}{k_n} \right)^{1+1/d} \left| \frac{\cut_n}{n (n-1)} 
      - 2 \int_{S} p^2 (s) \capintegral^{(1)} \left( r_n (s) \right) \  \mathrm{d}s \right|
      = O \left( \sqrt[d]{\frac{k_n}{n}} + \sqrt{\frac{\log n}{k_n}} \right) .
    \end{align*}

    For the {\em variance term} we have with Proposition~\ref{pro:cut_knn_general} 
    and $f_n (0)=1$
    \begin{align*}
      &\Pr \left( \left| \frac{1}{n k_n} \sqrt[d]{\frac{n-1}{k_n}} \cut_n 
      - \E \left( \frac{1}{n k_n} \sqrt[d]{\frac{n-1}{k_n}} \cut_n\right) \right| > \eps \right) 
      = \Pr \left( \left| \cut_n - \E \left( \cut_n \right) \right| > 
      n k_n \sqrt[d]{\frac{k_n}{n-1}} \eps \right) \\
      &\qquad \leq 2 \exp \left( - \tilde{C} \frac{\eps^2 n^2 k_n^2 (k_n/(n-1))^{2/d}}{n k_n^2 f_n^2 (0) } \right) 
      \leq 2 \exp \left( - \tilde{C} \eps^2 n^{1-2/d} k_n^{2/d} \right) .
    \end{align*}  
    
    Since $1/n=O(\sqrt[d]{k_n/n})$ we can change $\sqrt[d]{(n-1)/k_n}$ in the scaling 
    factor to $\sqrt[d]{n/k_n}$ without changing the convergence rate.

  \end{proof}

  \begin{corollary}[Gaussian weights and $1/\sigma_n (k_n/n)^{1/d} \to 0$]
    \label{cor:cut_knn_r_to_sigma_to_zero}
    Let $G_n$ be the $k$-nearest neighbor graph with Gaussian weight function
    and let $1/\sigma_n (k_n/n)^{1/d} \to 0$.
    Then
    \begin{align*}
      \left| \E \left( \frac{\sigma_n^d}{n k_n} \sqrt[d]{\frac{n}{k_n}} \cut_n \right) 
      - \frac{2 \eta_{d-1} \eta_d^{-1-1/d}}{(d+1) (2 \pi)^{d/2}} 
      \int_S p^{1-1/d} (s) \ \mathrm{d} s   \right|
      = O \left( \left( \frac{1}{\sigma_n} \sqrt[d]{\frac{k_n}{n}} \right)^{2} + \sqrt[d]{\frac{k_n}{n}} 
      + \sqrt{\frac{\log n}{k_n}} \right) 
    \end{align*}
    and, for a suitable constant $\tilde{C}>0$
    \begin{align*}
      &\Pr \left( \left| \frac{1}{n k_n} \sqrt[d]{\frac{n}{k_n}} \cut_n 
      - \E \left( \frac{1}{n k_n} \sqrt[d]{\frac{n}{k_n}} \cut_n\right) \right| > \eps \right)
      \leq 2 \exp \left( - \tilde{C} \eps^2 n^{1-2/d} k_n^{2/d} \right) .
    \end{align*}
  \end{corollary}

  \begin{proof}{}
    According to Lemma~\ref{pro:gaussian_weights_r_to_sigma_to_zero}
    we have for all $s \in S \cap C$
    \begin{align*}
      \left|\frac{\sigma_n^{qd}}{r_n^{d+1} (s)} \capintegral^{(q)} (r_n (s)) 
      - \frac{\eta_{d-1}}{(d+1) (2 \pi)^{qd/2}} \right|
      \leq 2 \left( \frac{r_n (s)}{\sigma_n} \right)^2 .
    \end{align*}
    Plugging in $r_n (s) = \sqrt[d]{k_n/((n-1) \eta_d p(s))}$ we obtain
    \begin{align*}
      \left| \sigma_n^{qd} \left(\frac{n-1}{k_n} \right)^{1+1/d}
      \left( \eta_d p(s) \right)^{1+1/d} \capintegral^{(q)} (r_n (s)) 
      - \frac{\eta_{d-1}}{(d+1) (2 \pi)^{qd/2}} \right|
      \leq 2 \left( \frac{1}{\sigma_n} \sqrt[d]{\frac{k_n}{(n-1) \eta_d p(s)}} 
      \right)^2 
    \end{align*}
    and therefore
    \begin{multline*}
      \left| \sigma_n^{qd} \left(\frac{n-1}{k_n} \right)^{1+1/d} \capintegral^{(q)} (r_n (s)) 
      - \frac{\eta_{d-1} \eta_d^{-1-1/d}}{(d+1) (2 \pi)^{qd/2}} p(s)^{-1-1/d} \right| \\
      \leq 2 (\eta_d p(s) )^{-1-1/d} \left( \frac{1}{\sigma_n} \sqrt[d]{\frac{k_n}{(n-1) \eta_d p(s)}} 
      \right)^2 
      \leq \tilde{C}_1 \left( \frac{k_n}{\sigma_n^d n } \right)^{2/d} 
    \end{multline*}
    for a suitable constant $\tilde{C}_1>0$.
    Therefore
    \begin{align*}
      &\left| \sigma_n^{d}  \left( \frac{n-1}{k_n} \right)^{1+1/d} \ 2 \int_{S \cap C} p^2 (s) \capintegral^{(1)} (r_n (s)) \ \mathrm{d} s
      - \frac{2 \eta_{d-1} \eta_d^{-1-1/d}}{(2 \pi)^{d/2} (d+1)} 
      \int_S p^{1-1/d} (s) \ \mathrm{d} s \right|      \\
      &\quad = \left| \sigma_n^{d}  \left( \frac{n-1}{k_n} \right)^{1+1/d} \ 2 \int_{S \cap C} p^2 (s) \capintegral^{(1)} (r_n (s)) \ \mathrm{d} s
      - 2 \int_S p^2 (s) \frac{\eta_{d-1} \eta_d^{-1-1/d}}{(2 \pi)^{d/2} (d+1)}  p^{-1-1/d} (s)
      \ \mathrm{d} s \right|      \\
      &\quad \leq  2 \int_{S \cap C}  p^2 (s) \left| \sigma_n^{d}  \left( \frac{n-1}{k_n} \right)^{1+1/d} 
      \capintegral^{(1)} (r_n (s)) - \frac{\eta_{d-1} \eta_d^{-1-1/d}}{(2 \pi)^{d/2} (d+1)}  p^{-1-1/d} (s)
      \right| \ \mathrm{d} s \\
      &\quad \leq  2 \int_{S \cap C}  p^2 (s) \tilde{C}_1 \left( \frac{k_n}{n \sigma_n^d} \right)^{2/d} \ \mathrm{d} s
      = 2 \tilde{C}_1 \left( \frac{k_n}{n \sigma_n^d} \right)^{2/d} p_{\max}^2 
      \lebesguevol_{d-1} \left( S \cap C \right) .
    \end{align*}
    
    Now, we consider the error terms of Proposition~\ref{pro:cut_knn_general}.
    For the first one we have, 
    using that $\capintegral^{(1)} ( r_n^{\max} ) = O ( (r_n^{\max})^{d+1}/\sigma_n^d )$
    and, furthermore, $r_n^{\max} = O(\sqrt[d]{k_n/(n-1)})$
    \begin{align*}
      &\sigma_n^{d}  \left( \frac{n-1}{k_n} \right)^{1+1/d}
      \capintegral^{(1)} ( r_n^{\max} ) \sqrt[d]{\frac{k_n}{n}} 
      = O \left( \sigma_n^{d}  \left( \frac{n-1}{k_n} \right)^{1+1/d}
      \sigma_n^{-d} \left( \frac{k_n}{n-1} \right)^{1+1/d} 
      \sqrt[d]{\frac{k_n}{n}} \right)
      = O \left( \sqrt[d]{\frac{k_n}{n}} \right) .
    \end{align*}

    For the second error term we have with $\delta_0=4$
    \begin{align*}
      &\sigma_n^{d}  \left( \frac{n-1}{k_n} \right)^{1+1/d}
      n^{-\delta_0} f_n \left( \inf_{x \in C} r_n (x) \right)
      \leq \sigma_n^{d}  n^2 n^{-4} \frac{1}{(2 \pi)^{d/2} \sigma_n^d}  
      = O \left( n^{-2} \right) .
    \end{align*}
   
    For the third error term we have with $f_n (0) = O (\sigma_n^{-d})$
    and the monotonicity of $f_n$
    \begin{align*}
      &\sigma_n^{d}  \left( \frac{n-1}{k_n} \right)^{1+1/d}
      \left( \sqrt[d]{\frac{k_n}{n}} + \sqrt{\frac{\log n}{k_n}} \right) f_n \left( \inf_{x \in C} r_n^- (x) \right)
      \left( \frac{k_n}{n} \right)^{1+1/d} 
      = O \left( \sqrt[d]{\frac{k_n}{n}} + \sqrt{\frac{\log n}{k_n}} \right).
    \end{align*}

    For the {\em variance term} we have with Proposition~\ref{pro:cut_knn_general} 
    and $f_n (0)=(2 \pi)^{-d/2} \sigma_n^{-d}$ for a suitable constant $\tilde{C}'>0$
    \begin{align*}
      &\Pr \left( \left| \frac{\sigma_n^d}{n k_n} \sqrt[d]{\frac{n-1}{k_n}} \cut_n 
      - \E \left( \frac{\sigma_n^d}{n k_n} \sqrt[d]{\frac{n-1}{k_n}} \cut_n\right) \right| > \eps \right) = \Pr \left( \left| \cut_n - \E \left( \cut_n \right) \right| > 
      \frac{n k_n}{\sigma_n^d} \sqrt[d]{\frac{k_n}{n-1}} \eps \right) \\
      &\qquad \leq 2 \exp \left( - \tilde{C}' \frac{\eps^2 n^2 k_n^2 \sigma_n^{-2d} (k_n/(n-1))^{2/d}}{n k_n^2 f_n^2 (0) } \right)
      \leq 2 \exp \left( - \tilde{C} \eps^2 n^{1-2/d} k_n^{2/d} \right) ,
    \end{align*}
    where we have set $\tilde{C} = (2 \pi)^{d} \tilde{C}'$.

    Since $1/n=O(\sqrt[d]{k_n/n})$ we can change $\sqrt[d]{(n-1)/k_n}$ in 
    the scaling factor to $1/(nk_n) \sqrt[d]{n/k_n}$ without changing the 
    convergence rate.
  \end{proof}

  \begin{corollary}[Gaussian weights and $\sigma_n (k_n/n)^{-1/d} \to 0$ ]    
    \label{cor:cut_knn_sigma_to_r_to_zero}
    We consider the $\kNN$ graph with Gaussian weight function. 
    Let $\sigma_n (k_n/n)^{-1/d} \to 0$ and $n \sigma_n^{d+1} \to \infty$
    for $n \to \infty$.
    Then there exists a constant $\tilde{C}>0$ such that
    \begin{align*}
      \left| \E \left( \frac{1}{n^2 \sigma_n} \cut_n \right)
      - \frac{2}{\sqrt{2 \pi}} \int_S p^2 (s) \ \mathrm{d} s \right|
      = O \left( \sqrt[d]{\frac{k_n}{n}} 
      + \frac{1}{\sigma_n} \exp \left( - \tilde{C} \left( \frac{1}{\sigma_n} 
      \sqrt[d]{\frac{k_n}{n}} \right)^2 \right) \right) .
    \end{align*}
    Furthermore, suppose $\sqrt[d]{k_n/n} \geq \sigma_n^\alpha$ for an $\alpha \in (0,1)$
    and $n$ sufficiently large.
    Then there exist non-negative random variables $D_n^{(1)},D_n^{(2)}$
    such that
    \begin{align*} 
      \left| \frac{\cut_{n}}{n^2 \sigma_n} 
      - \E \left( \frac{\cut_{n}}{n^2 \sigma_n} \right) \right|
      = O ( \sigma_n ) + D_n^{(1)} + D_n^{(2)} ,
    \end{align*}
    with $\Pr ( D_n^{(1)} > \eps) \leq 2 \exp ( \tilde{C}_2 n \sigma_n^{d+1} \eps^2 )$
    for a constant $\tilde{C}_2>0$, and $\Pr ( D_n^{(2)}>\sigma_n ) \leq 1/n^3$.
  \end{corollary}

  \begin{proof}{}
    With Lemma~\ref{pro:gaussian_weight_function} we have for 
    for $\sqrt[d]{k_n/n} / \sigma_n$ sufficiently large
    \begin{align*}
      \left| \frac{2}{\sigma_n} \int_{S \cap C} p^2 (s) \capintegral^{(1)} (r_n (s)) \ \mathrm{d}s
      - \frac{2}{\sqrt{2 \pi}} \int_S p^2 (s) \ \mathrm{d}s \right| 
      &\leq 2 \int_{S \cap C} p^2 (s) 
      \left| \frac{1}{\sigma_n} \capintegral^{(1)} (r_n (s)) - \frac{1}{\sqrt{2 \pi}} \right| \ \mathrm{d} s \\
      &= O \left( \exp \left( - \frac{1}{4 (p_{\max} \eta_d)^{2/d}} \left( \frac{1}{\sigma_n} \sqrt[d]{\frac{k_n}{n}} \right)^2 \right) \right) ,
    \end{align*}
    where we use that $p$ and $\lebesguevol_{d-1} (S \cap C)$ are bounded.

    Now we bound the error terms from Proposition~\ref{pro:cut_knn_general}
    of the other difference
    \begin{align*}
      &\left| \E \left( \frac{1}{n (n-1) \sigma_n} \cut_n \right)
      - \frac{2}{\sigma_n} \int_{S \cap C} p^2 (s) \capintegral^{(1)} (r_n (s)) \ \mathrm{d}s \right| .
    \end{align*}
    
    For the first one we observe that with 
    Lemma~\ref{pro:gaussian_weight_function} 
    we have $\capintegral^{(1)} ( r_n^{\max} )= O(\sigma_n)$ and therefore 
    $\sigma_n^{-1} \capintegral^{(1)} ( r_n^{\max} ) \sqrt[d]{k_n/n}=O (\sqrt[d]{k_n/n} )$.
   
    For the second one we have with Lemma~\ref{pro:gaussian_weight_function} 
    \begin{align*}
      \frac{1}{\sigma_n} \left( \ballintegral^{(1)} (\infty) - \ballintegral^{(1)} (\inf_{x \in C} r_n (x)) \right) 
      = O \left( \frac{1}{\sigma_n} \exp \left( - \frac{1}{4 (p_{\max} \eta_d)^{2/d}} \left( \frac{1}{\sigma_n} \sqrt[d]{\frac{k_n}{n}} \right)^2 \right) \right) .   
    \end{align*}

    For the third error term we observe that if $n$ is sufficiently large such that 
    $\delta_n \leq 1/2$ and $\xi_n \leq 1/4$ then for all $x \in C$,
    \begin{align*}
      r_n^- (x) = \sqrt[d]{\frac{(1-2\xi_n)(1-\delta_n) k_n}{(n-1) p(x) \eta_d}}
      \geq \sqrt[d]{\frac{k_n}{4 p_{\max} \eta_d n}} .
    \end{align*}
    Then we have with Lemma~\ref{pro:gaussian_weight_function} 
    \begin{align*}
      \frac{1}{\sigma_n} \left( \capintegral^{(1)} (\infty) - \capintegral^{(1)} (\inf_{x \in C} r_n^- (x)) 
      \right) 
      = O \left( \exp \left( - \frac{1}{4 (4 p_{\max} \eta_d)^{2/d}} \left (\frac{1}{\sigma_n} \sqrt[d]{\frac{k_n}{n}} \right)^2 \right)  \right) .
    \end{align*}

    Now we proof the bound for the variance term. Unfortunately, the bound
    in Proposition~\ref{pro:cut_knn_general} based on McDiarmid's inequality
    does not give good results. Therefore we proof a bound on the variance term
    directly.
    We set $\overline{\cut_n}$ to be the $\cut_n$ in the complete graph with Gaussian 
    weights on the sample and we set $\cut_n^\text{miss}$ to be sum of the weights 
    of the edges that are in the $\cut$ but not in the $\kNN$ graph.
    Then $\cut_n = \overline{\cut_n}-\cut_n^\text{miss}$ and we have
    \begin{align*}
      &\left| \frac{\cut_n}{n (n-1) \sigma_n} 
      - \E \left( \frac{\cut_n}{n (n-1) \sigma_n} \right) \right| \\
      &\qquad = \left|\frac{\overline{\cut_n}}{n (n-1) \sigma_n} 
      - \E \left( \frac{\overline{\cut_n}}{n (n-1) \sigma_n} \right)
      - \left( \frac{\cut_n^\text{miss}}{n (n-1) \sigma_n} 
      - \E \left( \frac{\cut_n^\text{miss}}{n (n-1) \sigma_n} \right) \right) \right| \\
      &\qquad \leq \left|\frac{\overline{\cut_n}}{n (n-1) \sigma_n} 
      - \E \left( \frac{\overline{\cut_n}}{n (n-1) \sigma_n} \right) \right|
      + \frac{\cut_n^\text{miss}}{n (n-1) \sigma_n} 
      + \E \left( \frac{\cut_n^\text{miss}}{n (n-1) \sigma_n} \right).
    \end{align*}
    The first deviation term is dealt with in Corollary~\ref{cor:cut_complete_graph}.
    
    We denote with $\mathcal{D}$ the event that the $k$-nearest neighbor radius 
    of all the points is greater than 
    $r_n^{\min} = \sqrt[d]{k_n/(2 p_{\max} \eta_d (n-1))}$.
    One can show similarly to the proof of Lemma~\ref{lem:prob_knn_radii} 
    that $\Pr (\mathcal{D}^c) \leq \exp (\log n - k_n/8)$
    and thus $\Pr (\mathcal{D}^c) \leq 1/n^3$ for sufficiently large $n$,
    since $k_n/\log n \to \infty$.
    If $\mathcal{D}$ holds, all the edges in $\cut_n^\text{miss}$
    must have weight lower than $f_n (r_n^{\min})$, whereas 
    if $\mathcal{D}^c$ holds the maximum edge weight is $f_n (0)$.
    There are $n (n-1)$ possible edges and thus
    \begin{align*}
      \E \left( \frac{\cut_n^\text{miss}}{n (n-1) \sigma_n} \right)
      &\leq \frac{1}{n (n-1) \sigma_n} n (n-1) f_n (0) \Pr (\mathcal{D}^c )
      + \frac{1}{n (n-1) \sigma_n} n (n-1) f_n (r_n^{\min}) \Pr (\mathcal{D}) \\
      &= O \left( \frac{1}{\sigma_n^{d+1}} \frac{1}{n^3}
      + \frac{1}{\sigma_n^{d+1}} \exp \left(-\frac{(r_n^{\min})^2}{2 \sigma_n^2}  \right)  \right)
      = O \left( \frac{1}{n^2} + \frac{1}{\sigma_n^{d+1}} 
      \exp \left(-\frac{(r_n^{\min})^2}{2 \sigma_n^2}  \right)  \right) ,
    \end{align*}
    since $n \sigma_n^{d+1} \to \infty$ for $n \to \infty$.
    
    Under the condition $\sqrt[d]{k_n/n} \geq \sigma_n^\alpha$ with $\alpha \in (0,1)$
    we have for sufficiently large $n$ and a suitable constant $\tilde{C}_1$
    \begin{align*}
      \frac{1}{\sigma_n^{d+1}} \exp \left(-\frac{(r_n^{\min})^2}{2 \sigma_n^2}  \right)
      \leq \frac{1}{\sigma_n^{d+1}} \exp \left(- \tilde{C}_1 \sigma_n^{2 (\alpha -1)}  \right)
      \leq \sigma_n ,
    \end{align*}
    where we use that the exponential term converges to zero faster than any power of
    $\sigma_n$.

    For the other term we clearly have for $n$ sufficiently large
    \begin{align*}
      \Pr \left( \frac{\cut_n^\text{miss}}{n (n-1) \sigma_n} > \sigma_n \right)
      \leq \Pr \left( \frac{\cut_n^\text{miss}}{n (n-1) \sigma_n} > \frac{1}{\sigma_n^{d+1}} \exp \left(-\frac{(r_n^{\min})^2}{2 \sigma_n^2}  \right) \right)
      \leq \Pr (\mathcal{D}^c) \leq \frac{1}{n^3} .
    \end{align*}
    
    Clearly, we can replace $n (n-1)$ in the scaling factor by $n^2$ without changing the 
    convergence rate.
  \end{proof}

  \subsubsection{The volume term of the $\kNN$ graph}
  \label{sec:vol_term_knn}

  \begin{proposition} \label{pro:volume_knn_general}
    Let $G_n$ be the $k$-nearest neighbor graph with a monotonically decreasing
    weight function $f_n$ and let $H=H^+$ or $H=H^-$.
    Then  
    \begin{align*}
      &\left| \E \left( \frac{\vol_n (H)}{n(n-1)} \right) 
      - \int_{H \cap C} 
      \ballintegral^{(1)} \left( r_n (x) \right) p^2 (x)  \ \mathrm{d} x  \right| \\
      &\qquad =O \left( \sqrt[d]{\frac{k_n}{n}} \ballintegral^{(1)} \left( r_n^{\max} \right) \right)+ O \left( \min \left\{ f_n^q \left( \inf_{x \in C} r_n (x) \right) 
      n^{-\delta_0} , \ballintegral^{(1)} (\infty) - \ballintegral^{(1)} (\inf_{x \in C} r_n (x))   \right\}   \right) \\ 
      &\qquad \quad + O \left( \min \left\{ f_n^q \left( \inf_{x \in C} r_n^- (x) \right) \left(\sqrt[d]{\frac{k_n}{n}} + \sqrt{\frac{\log n}{k_n}} \right) 
      \frac{k_n}{n}, \ballintegral^{(1)} \left( \infty \right) - \ballintegral^{(1)} \left( \inf_{x \in C} r_n^- (x) \right)  \right\} \right) .
    \end{align*}
    where we set $\delta_n = \sqrt{(4 \delta_0 \log n)/k_n}$ for a $\delta_0 \geq 2$
    in the definition of $r_n^- (x)$.

    For the variance term we have for a suitable constant $\tilde{C}>0$
    \begin{align*}
      \Pr \left( \left| \vol_n (H) - \E \left( \vol_n (H) \right) \right| > \eps \right) 
      \leq 2 \exp \left( - \tilde{C} \frac{\eps^2}{n k_n^2 f_n^2 (0)} \right) .
    \end{align*}
  \end{proposition}
  
  \begin{proof}{}
    Similarly to the proof of for the $\cut$ we define for 
    $i,j \in \{1,\ldots,n\}$, $i \neq j$ the random variable $W_{ij}$ as
    \begin{displaymath}
      W_{ij} = 
      \begin{cases}
	f_n ( \dist(x_i, x_j) 
	& \text{if } x_i \in H \text{ and } (x_i,x_j) \text{ edge in } G_n \\
	0 & \text{otherwise} 
      \end{cases}	
    \end{displaymath}
    and then have $\E \left( \vol_{n} (H) \right) = n (n-1) \E ( W_{12} )$.  
    With a function $c(x,y)$ that indicates the probability of connectedness
    we obtain
    \begin{align*}
     \E ( W_{12}^q ) &= \int_{H \cap C} \int_{C} f_n^q ( \dist (x, y) ) 
      c(x,y) p (y ) \ \mathrm{d}y \ p ( x ) \ \mathrm{d}x .
    \end{align*}
    Setting $\mathcal{R}_n = \{y \in H \cap C \; | \; \dist (y, \partial (H \cap C))\leq 2 r_n^{\max} \}$ and $\mathcal{I}_n = (H \cap C) \setminus  \mathcal{R}_n$
    we can decompose the outer integral into integrals over $\mathcal{R}_n$ and 
    $\mathcal{I}_n$.
    
    First suppose $x \in \mathcal{R}_n$ and let $c_n$ denote a bound on the 
    probability that points in distance at least $r_n^{\max}$ are connected.
    Then, using $c_n \leq 2 \exp \left( - k_n/8 \right)$ and Lemma~\ref{lem:ball_integral},
    \begin{align*}
      \int_{C} f_n^q ( \dist (x, y) ) c(x,y) p (y ) \ \mathrm{d} y 
      &\leq p_{\max} \int_{B(x,r_n^{\max}) \cap C} f_n^q ( \dist (x, y) ) \ \mathrm{d} y
      + f_n^q \left( r_n^{\max} \right) c_n 
      \int_{C}  p (y ) \ \mathrm{d} y\\
      &\leq p_{\max} d \eta_d \int_0^{r_n^{\max}} u^{d-1} f_n^q (u) \ \mathrm{d} u
      + 2 f_n^q \left( r_n^{\max} \right) \exp \left( - k_n/8 \right) \\
      &= p_{\max} \ballintegral^{(q)} \left( r_n^{\max} \right) 
      + 2 f_n^q \left( r_n^{\max} \right) \exp \left( - k_n/8 \right) .
    \end{align*}

    As was explained in the proof for the $\cut$ we can replace the term
    $2 f_n^q \left( r_n^{\max} \right) \exp \left( - k_n/8 \right)$
    by the term
    \begin{align*}
      p_{\max} \left( \ballintegral^{(q)} (\infty) - \ballintegral^{(q)} ( r_n^{\max}) \right) ,
    \end{align*}
    which is better suited, for example for the Gaussian.
    
    Therefore, using that according to Lemma~\ref{lem:boundary_strip_to_zero} 
    the volume of $\mathcal{R}_n$ is in $O(r_n^{\max})$,  
    \begin{align*}
      &\int_{\mathcal{R}_n} \int_{C} f_n^q ( \dist (x, y) ) c(x,y) p (y ) \ \mathrm{d} y 
      \ \mathrm{d} x
      = O \left( \sqrt[d]{\frac{k_n}{n}} \ballintegral^{(q)} \left( r_n^{\max} \right) \right)\\
      &\qquad + O \left( \min \left\{ \sqrt[d]{\frac{k_n}{n}} \left( \ballintegral^{(q)} (\infty) - \ballintegral^{(q)} ( r_n^{\max}) \right), 
      \sqrt[d]{\frac{k_n}{n}} f_n^q \left( r_n^{\max} \right) \exp \left( - k_n/8 \right)
      \right\} \right) .
    \end{align*}

    For $x \in \mathcal{I}_n$ we introduce as in the proof for the $\cut$ radii
    $r_n^- (x) \leq  r_n^{\max}$ and $r_n^+ (x) \leq  r_n^{\max}$
    that depend on $\delta_n$ and $\xi_n$ defined there. 
    These radii approximate the true $\kNN$ radius.
    For a lower bound we obtain
    \begin{align*}
      \int_{C} f_n^q ( \dist (x, y) ) c(x,y) p (y ) \ \mathrm{d}y
      \geq & \ballintegral^{(q)} \left( r_n (x) \right) p(x)
      - p_{\max} \left( \ballintegral^{(q)} \left( r_n (x) \right)
      - \ballintegral^{(q)} \left( r_n^- (x) \right) \right) \\
      &- \left( \xi_n + 6 \exp \left( - \delta_n^2 k_n / 3 \right) \right) p_{\max}
      \ballintegral^{(q)} \left( r_n^{\max} \right) .
    \end{align*}
    For some weight functions, especially the Gaussian, we can use
    \begin{align*}
      \ballintegral^{(q)} \left( r_n (x) \right) - \ballintegral^{(q)} \left( r_n^- (x) \right)
      \leq \ballintegral^{(q)} \left( \infty \right) - \ballintegral^{(q)} \left( \inf_{x \in C} r_n^- (x) \right) ,
    \end{align*}
    whereas for other ones it is better to use
    \begin{align*}
      \ballintegral^{(q)} \left( r_n (x) \right) - \ballintegral^{(q)} \left( r_n^- (x) \right)  
      &= d \eta_d  \int_{r_n^- (x)}^{r_n(x)} u^{d-1} f_n^q (u) \ \mathrm{d} u \\
      &\leq \eta_d f_n^q \left( \inf_{x \in C} r_n^- (x) \right) 
      \left( \xi_n + \delta_n \right) \left( r_n^{\max} \right)^d .
    \end{align*}
    
    Similarly we obtain an upper bound, with an additional term
    $f_n^q \left( \inf_{x \in C} r_n (x) \right) \exp \left( - \delta_n^2 k_n / 4 \right)$
    or $p_{\max} ( \ballintegral^{(q)} (\infty) - \ballintegral^{(q)} (\inf_{x \in C} r_n (x)) )$
    bounding the influence of points that are further away than $r_n^+ (x)$.
    Combining the bounds we obtain
    \begin{align*}
      &\left| \int_{\mathcal{I}_n} \int_{C} f_n^q ( \dist (x, y) ) c(x,y) p (y ) \ \mathrm{d}y
      - \int_{\mathcal{I}_n} \ballintegral^{(q)} \left( r_n (x) \right) p^2 (x)
      \ \mathrm{d} x \right| \\
      &\qquad = O \left( (\xi_n + \exp \left( - \delta_n^2 k_n / 3 \right) )  
      \ballintegral^{(q)} \left( r_n^{\max} \right) \right) \\ 
      &\qquad \quad + O \left( \min \left\{ f_n^q \left( \inf_{x \in C} r_n^- (x) \right) \left( \xi_n + \delta_n \right) 
      \left( r_n^{\max} \right)^d, \ballintegral^{(q)} \left( \infty \right) - \ballintegral^{(q)} \left( \inf_{x \in C} r_n^- (x) \right)  \right\} \right)\\
      &\qquad \quad + O \left( \min \left\{ f_n^q \left( \inf_{x \in C} r_n (x) \right) 
      \exp \left( - \delta_n^2 k_n / 4\right) , \ballintegral^{(q)} (\infty) - \ballintegral^{(q)} (\inf_{x \in C} r_n (x))   \right\}   \right) .
    \end{align*}
    Setting $\delta_n = \sqrt{(4 \delta_0 \log n)/ k_n}$ we obtain 
    $\exp \left( - \delta_n^2 k_n / 3 \right) \leq n^{-\delta_0}$
    and the same for $\exp \left( - \delta_n^d k_n / 4 \right)$.
    Clearly, for $\delta_0 \geq 2$ we have $n^{-\delta_0}\leq \xi_n$
    and $n^{-\delta_0} \leq ( \xi_n r_n^{\max} )^d$ .
    Thus, with $\xi_n = O(r_n^{\max}) = O(\sqrt[d]{k_n/n})$,
    \begin{align*}
      &\left| \int_{\mathcal{I}_n} \int_{C} f_n^q ( \dist (x, y) ) c(x,y) p (y ) \ \mathrm{d}y
      - \int_{\mathcal{I}_n} \ballintegral^{(q)} \left( r_n (x) \right) p^2 (x)
      \ \mathrm{d} x \right| \\
      &\qquad = O \left( \sqrt[d]{\frac{k_n}{n}} \ballintegral^{(q)} \left( r_n^{\max} \right) \right) \\ 
      &\qquad \quad + O \left( \min \left\{ f_n^q \left( \inf_{x \in C} r_n^- (x) \right) \left(\sqrt[d]{\frac{k_n}{n}} + \sqrt{\frac{\log n}{k_n}} \right) 
      \frac{k_n}{n}, \ballintegral^{(q)} \left( \infty \right) - \ballintegral^{(q)} \left( \inf_{x \in C} r_n^- (x) \right)  \right\} \right)\\
      &\qquad \quad + O \left( \min \left\{ f_n^q \left( \inf_{x \in C} r_n (x) \right) 
      n^{-\delta_0} , \ballintegral^{(q)} (\infty) - \ballintegral^{(q)} (\inf_{x \in C} r_n (x))   \right\}   \right) .
    \end{align*}

    Finally, by finding an upper bound on the integrand and the volume of 
    $(H \cap C)\setminus \mathcal{I}_n$ we obtain
    \begin{align*}
      &\left| \int_{\mathcal{I}_n} 
      \ballintegral^{(q)} \left( r_n (x) \right) p(x) \ \mathrm{d} x 
      - \int_{H \cap C} 
      \ballintegral^{(q)} \left( r_n (x) \right) p^2 (x)  \ \mathrm{d} x  \right| 
      = O \left( \sqrt[d]{\frac{k_n}{n}} \ballintegral^{(q)} \left( r_n^{\max} \right) \right).
    \end{align*}

    Combining all the bounds above we obtain the result for the bias term.
    The bound for the variance term can be obtained with McDiarmid's inequality
    similarly to the proof for the $\cut$ in Proposition~\ref{pro:cut_knn_general}.
  \end{proof}
  
  The following lemma is necessary for the proof of the general theorem for both, 
  the $r$-graph and the $\kNN$-graph.
  It is an elementary lemma and therefore stated without proof.
  \begin{lemma}[Integration over balls] \label{lem:ball_integral}
    Let $f_n:\R_{\geq 0} \to \R_{\geq 0}$ be a monotonically decreasing function and $x \in \R^d$.
    Then we have for any $R \in \R_{>0}$
    \begin{align*}
      \int_{B(x,R)} f( \dist(x,y) ) \ \mathrm{d} y 
      = d \eta_d \int_0^R u^{d-1} f(u) \ \mathrm{d} u .
    \end{align*}
  \end{lemma}

  \begin{corollary}[Unweighted $\kNN$-graph] \label{cor:vol_knn_unweighted}
    Let $G_n$ be the unweighted $\kNN$ graph with weight function $f_n \equiv 1$ 
    and let $H=H^+$ or $H=H^-$. 
    Then we have for the bias term
    \begin{align*}
      \left| \frac{\vol_n (H)}{n k_n} - \int_{H} p(x) \ \mathrm{d} x \right|
      &= O \left( \sqrt[d]{\frac{k_n}{n}} + \sqrt{\frac{\log n}{k_n}} \right) .
    \end{align*}
    and for the variance term for a suitable constant $\tilde{C}$
    \begin{align*}
      &\Pr \left( \left| \frac{\vol_n (H) }{n k_n} - \E \left( \frac{\vol_n (H)}{n k_n} \right) \right| > \eps \right) 
      \leq 2 \exp \left( - \tilde{C} n \eps^2 \right) .
    \end{align*}
  \end{corollary}

  \begin{proof}{}
    With Lemma~\ref{pro:integration_unit_weight_function}
    we have, plugging in the definition of $r_n (x)$, 
    \begin{align*}
      \int_{H \cap C} \ballintegral^{(1)} (r_n (x)) p^2 (x) \ \mathrm{d} x
      = \int_{H \cap C} \eta_d \frac{k_n}{(n-1) \eta_d p(x)} p^2 (x) \ \mathrm{d} x
      = \frac{k_n}{n-1} \int_{H} p(x) \ \mathrm{d} x .
    \end{align*}
    
    Therefore by multiplying the expression in Proposition~\ref{pro:volume_knn_general}
    with $(n-1)/k_n$ we obtain for any $\delta_0 \geq 2$
    \begin{align*}
      \left| \frac{\vol_n (H)}{n k_n} - \int_{H} p(x) \ \mathrm{d} x \right|
      & \leq O \left( \frac{n-1}{k_n} \sqrt[d]{\frac{k_n}{n}} \ballintegral^{(1)} \left( r_n^{\max} \right) \right) \\
      &\quad + O \left( \frac{n-1}{k_n} f_n \left( \inf_{x\in C} r_n^{-} (x) \right) n^{-\delta_0} \right) \\
      &\quad + O \left( \frac{n-1}{k_n} \frac{k_n}{n} \left( \sqrt[d]{\frac{k_n}{n}} + \sqrt{\frac{\log n}{k_n}} \right) 
      f_n \left( \inf_{x \in C} r_n^- (x) \right)  \right) .
    \end{align*}
    Using $\ballintegral^{(1)} (r_n^{\max}) \sim (n-1)/k_n$ and $f_n \equiv 1$
    we obtain
    \begin{align*}
      \left| \frac{\vol_n (H)}{n k_n} - \int_{H} p(x) \ \mathrm{d} x \right|
      &= O \left( \sqrt[d]{\frac{k_n}{n}} + \sqrt{\frac{\log n}{k_n}} \right) .
     \end{align*}

    For the variance term we use the bound in Proposition~\ref{pro:volume_knn_general} 
    and plug in $f_n (0)=1$.
  \end{proof}

  \begin{corollary}[Gaussian weights and $(k_n/n)^{1/d}/\sigma_n \to 0$]
    \label{cor:vol_knn_r_to_sigma_to_zero}
    Consider the $\kNN$ graph with Gaussian weights and $(k_n/n)^{1/d}/\sigma_n \to 0$.
    Let $H=H^+$ or $H=H^-$. Then we have for the bias term
    \begin{align*}
      \left| \frac{\sigma_n^{d}}{n k_n} \vol_n (H) 
      - \frac{1}{(2 \pi)^{d/2}} \int_H p(x) \ \mathrm{d} x \right|
      &= O \left( \left( \frac{1}{\sigma_n} \sqrt[d]{\frac{k_n}{n}} \right)^2
	+ \sqrt[d]{\frac{k_n}{n}} + \sqrt{\frac{\log n}{k_n}} \right)
    \end{align*}
    and for the variance term, for a suitable constant $\tilde{C}>0$,
    \begin{align*}
      &\Pr \left( \left| \frac{\sigma_n^{d}}{n k_n} \vol_n (H) - \E \left( \frac{\sigma_n^{d}}{n k_n} \vol_n (H) \right) \right| > \eps \right) 
      \leq 2 \exp \left( - \tilde{C} n \eps^2 \right). 
    \end{align*}
    
  \end{corollary}

  \begin{proof}{}
    According to Lemma~\ref{pro:gaussian_weights_r_to_sigma_to_zero} we have
    for all $x \in C$
    \begin{align*}
      \left|\frac{\sigma_n^{q d}}{r_n^{d} (x)} \ballintegral^{(q)} (r_n (x)) 
      - \frac{\eta_{d}}{(2 \pi)^{qd/2}} \right|
      \leq 3 \left( \frac{r_n (x)}{\sigma_n} \right)^2 .
    \end{align*}
    Plugging in $r_n (x) = \sqrt[d]{k_n/((n-1) \eta_d p(x))}$ and dividing 
    by $\eta_d p(x)$ we obtain for points in the support of $p$
    \begin{align*}
      \left| \sigma_n^{q d} \left(\frac{n-1}{k_n} \right) \ballintegral^{(q)} (r_n (x)) 
      - \frac{1}{(2 \pi)^{qd/2} p(x)} \right|
      = O \left( \left( \frac{k_n}{\sigma_n^d n } \right)^{2/d} \right) . 
    \end{align*}
    Therefore, using the boundedness of $p$ 
    \begin{align*}
      &\left| \sigma_n^{d}  \left( \frac{n-1}{k_n} \right) \ \int_{H \cap C} p^2 (x) \ballintegral^{(1)} (r_n (x)) \ \mathrm{d} x
      - \frac{1}{(2 \pi)^{d/2}} \int_H p (x) \ \mathrm{d} x \right|
      = O \left( \left( \frac{k_n}{n \sigma_n^d} \right)^{2/d} \right) .
    \end{align*}

    Now, we consider the error terms from Proposition~\ref{pro:volume_knn_general} 
    of the other difference 
    \begin{align*}
      \left| \frac{\sigma_n^{d}}{n k_n} \vol_n (H) -      
      \sigma_n^{d}  \left( \frac{n-1}{k_n} \right) \  \int_{H \cap C} p^2 (x) \ballintegral^{(1)} (r_n (x)) \ \mathrm{d} x \right| .
    \end{align*}
    As we have seen above $\sigma_n^{d} (n-1)/k_n \ballintegral^{(1)} ( r_n^{\max})$
can be bounded by a constant. Thus we have for the first term
    \begin{align*}
      \sigma_n^{d} \left( \frac{n-1}{k_n} \right) \sqrt[d]{\frac{k_n}{n}} \ballintegral^{(1)} \left( r_n^{\max} \right)
      = O \left( \sqrt[d]{\frac{k_n}{n}} \right) .
    \end{align*}
    
    For the second term we have for $n$ sufficiently large and setting $\delta_0 = 3$
    \begin{align*}
      \sigma_n^{d} \left( \frac{n-1}{k_n} \right) 
      f_n \left( \inf_{x \in C} r_n^- (x) \right) n^{-\delta_0}
      &\leq \sigma_n^{d} \left( \frac{n-1}{k_n} \right) 
      f_n \left( 0 \right) n^{-\delta_0} 
      \leq \left( \frac{n-1}{k_n} \right) n^{-\delta_0} \leq n^{-2} .
    \end{align*}
    
    For the third term we have 
    \begin{align*}
      \sigma_n^{d} \left( \frac{n-1}{k_n} \right)
      \frac{k_n}{n} \left( \sqrt[d]{\frac{k_n}{n}} + \sqrt{\frac{\log n}{k_n}} \right)
      f_n \left( \inf_{x \in C} r_n^- (x) \right)
      &\leq \left( \sqrt[d]{\frac{k_n}{n}} + \sqrt{\frac{\log n}{k_n}} \right)
      \sigma_n^{d} f_n \left( 0 \right) \\
      &=  \left( \sqrt[d]{\frac{k_n}{n}} + \sqrt{\frac{\log n}{k_n}} \right)
      \frac{1}{(2 \pi)^{d/2}} .
    \end{align*}
    
    For {\em the variance term} we have for a suitable constant $\tilde{C}'>0$
    \begin{align*}
      &\Pr \left( \left| \frac{\sigma_n^{d}}{n k_n} \vol_n (H) 
      - \E \left( \frac{\sigma_n^{d}}{n k_n} \vol_n (H) \right) \right| > \eps \right) 
      = \Pr \left( \left| \vol_n (H) 
      - \E \left( \vol_n (H) \right) \right| > n k_n \sigma_n^{-d} \eps \right) \\
      &\qquad \leq 2 \exp \left( - \tilde{C}' \frac{n^2 k_n^2 \sigma_n^{-2d} \eps^2}{n k_n^2 f_n^2 (0)} \right) 
      \leq 2 \exp \left( - \tilde{C}' \frac{n \sigma_n^{-2d} \eps^2}{\frac{1}{(2 \pi)^{d}} \sigma_n^{-2d}} \right) 
      = 2 \exp \left( - \tilde{C} n \eps^2 \right) ,
    \end{align*}
    where we have set $\tilde{C} = (2 \pi)^d \tilde{C}'$.
  \end{proof}

  \begin{corollary}[Gaussian weights and $(k_n/n)^{1/d}/\sigma_n \to \infty$]
    \label{cor:vol_knn_r_to_sigma_to_infinity}
    Let $G_n$ be the $\kNN$ graph with Gaussian weights. %
    Then for the bias term for a constant $\tilde{C}_1>0$ 
    \begin{align*}
      &\left| \E \left( \frac{\vol_n (H)}{n^2}  \right)
      - \int_H p^2 (x) \ \mathrm{d} x \right|
      = O \left( \sqrt[d]{\frac{k_n}{n}} 
      + \exp \left( - \tilde{C}_1  \left( \frac{1}{\sigma_n} \sqrt[d]{\frac{k_n}{n}} \right)^{2}  \right)  \right) .
    \end{align*}
    Let, furthermore, $\sqrt[d]{k_n/n} \geq \sigma_n^\alpha$ for an $\alpha \in (0,1)$
    and $n$ sufficiently large. 
    Then there exist non-negative random variables $D_n^{(1)},D_n^{(2)}$
    such that
    \begin{align*} 
      \left| \frac{\vol_{n} (H)}{n^2} 
      - \E \left( \frac{\vol_{n} (H)}{n^2} \right) \right|
      = O ( \sigma_n ) + D_n^{(1)} + D_n^{(2)} ,
    \end{align*}
    with $\Pr ( D_n^{(1)} > \eps) \leq 2 \exp ( \tilde{C}_2 n \sigma_n^{d+1} \eps^2 )$
    for a constant $\tilde{C}_2>0$, and $\Pr ( D_n^{(2)}>\sigma_n ) \leq 1/n^3$.
  \end{corollary}

  \begin{proof}{}
    With Lemma~\ref{pro:gaussian_weight_function} we have for 
    $n$ sufficiently large such that $r_n (x) / \sigma_n$ sufficiently large
    uniformly over all $x \in C$
    \begin{align*}
      &\left| \int_{H \cap C}  \ballintegral^{(1)} (r_n (x)) p^2 (x) \ \mathrm{d} x
      - \int_H p^2 (x) \ \mathrm{d} x \right|
      \leq \int_{H \cap C} \left| \ballintegral^{(1)} (r_n (x)) 
      - 1 \right| p^2 (x) \ \mathrm{d} x \\
      &\qquad \qquad \qquad = O \left( \exp \left( - \frac{1}{4 (p_{\max} \eta_d)^{2/d}} \frac{1}{\sigma_n^2} \left( \frac{k_n}{n} \right)^{2/d}  \right)  \right) .
    \end{align*}

    Now we bound the error terms from Proposition~\ref{pro:volume_knn_general} 
    of the other difference 
    \begin{align*}
      &\left| \E \left( \frac{1}{n (n-1)} \vol_n (H) \right)
      - \int_{H \cap C} p^2 (x) \capintegral^{(1)} (r_n (x)) \ \mathrm{d}x \right| .
    \end{align*}

    For the first error term we use that according to 
    Lemma~\ref{pro:gaussian_weight_function} 
    $\ballintegral^{(1)} ( r_n^{\max} )$ is bounded by one
    for $n$ sufficiently large. 
    Therefore $\sqrt[d]{k_n/n} \ballintegral^{(1)} ( r_n^{\max} )
    = O (\sqrt[d]{k_n/n} )$.  
    
    For the second and third error term we observe that if $n$ is sufficiently large 
    such that $\delta_n \leq 1/2$ and $\xi_n \leq 1/4$ then
    \begin{align*}
      \inf_{x \in C} r_n (x) \geq \inf_{x \in C} r_n^- (x) 
      = \inf_{x \in C}\sqrt[d]{\frac{(1-2\xi_n)(1-\delta_n) k_n}{(n-1) p(x) \eta_d}}
      \geq \sqrt[d]{\frac{k_n}{4 p_{\max} \eta_d n}} ,
    \end{align*}
    and therefore, for both, the second and the third error term,
    \begin{align*}
      \ballintegral^{(1)} (\infty) - \ballintegral^{(1)} (\inf_{x \in C} r_n (x) ) 
      = O \left( \exp \left( - \frac{1}{4 (4 p_{\max} \eta_d)^{2/d}} \left( \frac{1}{\sigma_n}
      \sqrt[d]{\frac{k_n}{n}} \right)^2 \right) \right).
    \end{align*}
    
    The proof of the bound for the variance term is identical to the corresponding
    part in the proof of Corollary~\ref{cor:cut_knn_sigma_to_r_to_zero}.
    Therefore, we do not repeat it here.
    
    Clearly, we can replace $n (n-1)$ in the scaling factor by $n^2$ without changing the 
    convergence rate.
  \end{proof}

  \subsubsection{The main theorem for the $\kNN$ graph}
  \label{sec:proof_main_theorem_knn}
  
  \begin{proof}{ of Theorem~\ref{thm:limit_ncut_cheegercut_knn_graph}}
    As discussed in Section~\ref{sec:proofs_intuitively} we can study the
    convergence of the bias and variance terms of the $\cut$ and the volume 
    separately.

    For the {\em unweighted graph} we have with Corollary~\ref{cor:cut_knn_unweighted}
    that under the condition $k_n / \log n \to \infty$ the bias term for the $\cut$ is
    in $O ( \sqrt[d]{k_n/n} + \sqrt{\log n/k_n} )$.
    For some $\eps>0$ the probability that the variance term exceeds $\eps$ 
    is bounded by $2 \exp ( - \tilde{C} \eps^2 n^{1-2/d} k_n^{2/d} )$
    for a suitable constant $\tilde{C}$.
    Clearly, the bias term converges to zero under the condition $k_n / \log n \to \infty$.
    For the almost sure convergence of the variance term we need the stricter 
    condition in dimension $d=1$.
    The convergence of the volume-term follows with 
    Corollary~\ref{cor:vol_knn_unweighted},
    since the requirements for this convergence are weaker.
    In the case $d \geq 2$ we obtain the optimal rates by equating the two bounds
    of the bias term and checking that the variance term converges as well at this rate.
    In the case $d=1$ the optimal rate is determined by the variance term.

    For the {\em $\kNN$-graph with Gaussian weights and $r_n/\sigma_n \to \infty$} 
    we need the stronger condition $r_n \geq \sigma_n^\alpha$  for an $\alpha \in (0,1)$
    in order to show convergence
    of both, the bias term and the variance term. Under this condition
    we have according to Corollaries~\ref{cor:cut_knn_sigma_to_r_to_zero} and 
    \ref{cor:vol_knn_r_to_sigma_to_infinity} that the bias term of both, 
    the $\cut$ and the volume, is in $O ( r_n )$, since the exponential term
    converges as $\sigma_n$.
  
    Furthermore, the almost sure convergence of the variance term 
    can be shown with the Borel-Cantelli lemma if $n \sigma_n^{d+1}/\log n \to \infty$
    for $n \to \infty$.

    For the {\em $\kNN$-graph with Gaussian weights and $r_n/\sigma_n \to 0$} 
    according to Corollary~\ref{cor:cut_knn_r_to_sigma_to_zero} the bias
    term of the $\cut$ is in $O ( r_n + (r_n/\sigma_n)^2 + \sqrt{\log n / k_n})$.
    The probability that the variance term of the $\cut$ exceeds an $\eps>0$ 
    is bounded by $2 \exp (- \tilde{C} n^{1-2/d} k_n^{2/d} )$ for a suitable constant
    $\tilde{C}$, which is the same 
    expression as in the unweighted case. Therefore, we have almost sure convergence
    of the $\cut$-term to zero under the same conditions as for the unweighted 
    $\kNN$ graph.
    
    From Corollary~\ref{cor:vol_knn_r_to_sigma_to_zero} we can see that the 
    convergence conditions for the volume are less strict than that of the $\cut$.  
  \end{proof}

  \subsection{The $r$-graph and the complete weighted graph}
  \label{sec:proofs_r_graph}

  This section consists of three parts: In the first one the convergence of the
  bias and variance term of the $\cut$ is studied, whereas in the second part that 
  convergence is studied for the volume. Combining these results we can proof the
  main theorems on the convergence of $\ncut$ and $\cheegercut$ for the $r$-graph 
  and the complete weighted graph.

  Section~\ref{sec:cut_r_graph_complete_graph} and 
  Section~\ref{sec:volume_r_graph_complete_graph} are built up similarly:
  First, a proposition for a general weight function is given. The results
  are stated in terms of the ``cap'' and ``ball'' integrals and some properties
  of the weight function. 
  Then four corollaries follow, where the general result is applied to the
  complete weighted graph with Gaussian weight function and to the $r$-graph with the
  specific weight functions we consider in this paper.
  
  Some words on the proofs: The results on the bias terms for general weight
  functions can be shown analogously to the corresponding results for the $\kNN$
  graph. Since the connectivity in these graphs given the position of two points
  is not random they are even simpler. Furthermore, all the error terms in the result
  for the $\kNN$ graph that are due to the uncertainty in the connectivity radius  
  can be dropped for the $r$-graph and the complete weighted graph. 
  Therefore, in the proof of the bias term of the $\cut$ we only discuss the 
  adaptations that are made to the proof of the $\kNN$ graph.

  As explained in Section~\ref{sec:proofs_intuitively} the situation is different 
  for the variance term, where the convergence proof for the $\kNN$-graph would 
  lead to suboptimal results when carried over to the other two graphs.  
  For this reason we give a different proof for the convergence of the variance term 
  in the proof of the general result for the $\cut$. 
  It can be easily carried over to the volume and thus we omit it there.

  As to the corollaries we only proof two of them: that for the complete weighted
  graph and that for the $r$-graph with Gaussian weights and $r_n/\sigma_n \to 0$
  for $n \to \infty$. The proof of the corollary for the unweighted graph 
  is very simple, that of the corollary for the $r$-graph with Gaussian weights and
  $\sigma_n/r_n \to 0$ is identical to the proof for the complete weighted graph
  where we can ignore one term.

  The proofs in Section~\ref{sec:volume_r_graph_complete_graph} are completely 
  omitted: The general result on the bias term can be proved analogously to that
  for the $\kNN$ graph, if the adaptations that are discussed in the proof for
  the bias term of the $\cut$ are made. The general result on the variance term
  of the volume is proved analogously to that on the variance term of the $\cut$.
  The proofs of the corollaries also work analogously to the corresponding proofs
  for the $\cut$.

  The proofs of the main theorems in Section~\ref{sec:proofs_main_theorems_r_graph}
  collect the bounds of the corollaries and identify the conditions that have to hold
  for the convergence of $\ncut$ and $\cheegercut$.

  \subsubsection{The $\cut$ term in the $r$-graph and the complete weighted graph}
  \label{sec:cut_r_graph_complete_graph}
  
  \begin{proposition}[The $\cut$ in the $r$-neighborhood and the complete weighted graph]
    \label{pro:lim_cut_general_weight_r_graph}
    Let $(r_n)_{n \in \N}$ be a sequence that fulfills the conditions on parameter
    sequences of the $r$-neighborhood graph. 
    Let $G_n$ denote the $r$-neighborhood graph with parameter $r_n$ 
    or the complete weighted graph on $x_1, \ldots ,x_n$ 
    with a monotonically decreasing weight function 
    $f_n:\R_{\geq 0} \to \R_{\geq 0}$. We set
    \begin{align*}
      \indicator_c =
      \begin{cases}
	$1$ & \text{if $G_n$ is the complete weighted graph} \\
	$0$ & \text{if $G_n$ is the $r_n$-neighborhood graph.}
      \end{cases}
    \end{align*}
        
    Then for the bias term
    \begin{align*}
      \left| \E \left( \frac{\cut_n}{n (n-1) \capintegral^{(1)} (r_n)} \right) 
      - 2 \int_{S} p^2 (s)  \ \mathrm{d}s \right|
      = O \left( r_n + \frac{\ballintegral^{(1)} (\infty) - \ballintegral^{(1)} (r_n)}{\capintegral^{(1)} (r_n)}  \indicator_{c} \right) .
    \end{align*}
    Furthermore, there are constants $\tilde{C}_1,\tilde{C}_2$ such that 
    for the variance term
    \begin{multline*}
      \Pr \left( \left| \frac{\cut_{n}}{n (n-1) \capintegral^{(1)} (r_n)} 
      - \E \left( \frac{\cut_{n}}{n (n-1) \capintegral^{(1)} (r_n)} \right) \right| \geq \eps \right) \\
      \leq 2 \exp \left(- \frac{ n \left( \capintegral^{(1)} (r_n) \right)^{2} \eps^2}{\tilde{C}_1 \capintegral^{(2)} (r_n) + \tilde{C}_2 (\ballintegral^{(2)} (\infty) - \ballintegral^{(2)} (r_n) ) \indicator_{c} + 2 \eps  \capintegral^{(1)} (r_n) \ f_n (0)} \right) .
    \end{multline*}
  \end{proposition}

  \begin{proof}{}    
    As was said in the introduction we do not give the detailed proof of this proposition
    here, since it is similar to the proof of the corresponding proposition 
    for the $\kNN$-graph but simpler:
    the radius $r_n$ is the same everywhere, that is we can set
    $r_n^{\max}=r_n^+(s)+=r_n^-(s)=r_n$ for all $s \in S$.
    Furthermore, the connectivity is not random, that is we can set $a_n=b_n=c_n=0$
    for the $r$-neighborhood graph, whereas we set $a_n=0$,$b_n=1$ and $c_n=1$
    for the complete weighted graph.
    We obtain
    \begin{align*}
      \left| \E \left( W_{12}^q \right) 
      - 2 \capintegral^{(q)} (r_n) \ \int_{S} p^2 (s)  \ \mathrm{d} s \right| 
      = O \left( \capintegral^{(q)} (r_n) r_n + \left( \ballintegral^{(q)} (\infty) - \ballintegral^{(q)} (r_n)\right) \indicator_{c} \right) ,
    \end{align*}
    and thus the result for the bias term immediately.
    
    In order to bound the variance term we use a $U$-statistics 
    argument. We have
    \begin{align*}
      \frac{\cut_{n}}{n (n-1) \capintegral^{(1)} \left( r_n \right)}
      = \frac{1}{n (n-1)} \sum_{i=1}^n \mathop{\sum_{j=1}^n}_{j \neq i} 
      \frac{1}{\capintegral^{(1)} \left( r_n \right)} W_{ij} .
    \end{align*}

    For the upper bound on the properly rescaled variable $W_{ij}$ clearly 
    \begin{align*}
      \frac{1}{\capintegral^{(1)} \left( r_n \right)} W_{ij} 
      \leq \frac{1}{\capintegral^{(1)} \left( r_n \right)} f_n (0)
    \end{align*}
    and for the variance 
    \begin{align*}
      \Var \left( \frac{1}{\capintegral^{(1)} \left( r_n \right)} W_{ij} \right) 
      &= \E \left(  \left( \frac{1}{\capintegral^{(1)} \left( r_n \right)} W_{ij} \right)^2 \right) 
      - \left( \E  \left( \frac{1}{\capintegral^{(1)} \left( r_n \right)} W_{ij} \right) \right)^2 
      \leq \left( \frac{1}{\capintegral^{(1)} \left( r_n \right)} \right)^2 
      \E \left(  W_{ij}^2 \right) .
    \end{align*}

    With a Bernstein-type concentration inequality for $U$-statistics 
    from~\citet{Hoeffding63} we obtain
    \begin{align*}
      &\Pr \left( \left| \frac{\cut_{n}}{n (n-1) \capintegral^{(1)} \left( r_n \right)} 
      - \E \left( \frac{\cut_{n}}{n (n-1) \capintegral^{(1)} \left( r_n \right)} \right) \right| \geq \eps \right) \\
      &\qquad \leq 2 \exp \left(- \frac{ \lfloor n/2 \rfloor \eps^2}{2 \left( \frac{1}{\capintegral^{(1)} \left( r_n \right)} \right)^2 \E \left(  W_{ij}^2 \right) + \frac{2}{3} \frac{1}{\capintegral^{(1)} \left( r_n \right)} \eps f_n (0)} \right) 
      \leq 2 \exp \left(- \frac{n \eps^2 \left( \capintegral^{(1)} \left( r_n \right) \right)^2}{6 \E \left(  W_{ij}^2 \right) + 2 \eps \capintegral^{(1)} \left( r_n \right) f_n (0)} \right) 
    \end{align*}
    where we have used $\lfloor n/2 \rfloor \geq n/3$ for $n \geq 2$
   
    Clearly, for $r_n \to 0$ we can find constants (depending on $p$ and $S$)
    $\tilde{C}_1$ and $\tilde{C}_2$ such that for $n$ sufficiently large
    $6 \E (W_{ij}^2) \leq \tilde{C}_1 \capintegral^{(2)} (r_n) + \tilde{C}_2 (\ballintegral^{(2)} (\infty) - \ballintegral^{(2)} (r_n) ) \indicator_{c}$.
  \end{proof}

  The following corollary can be proved by plugging in the results of 
  Lemma~\ref{pro:integration_unit_weight_function} into the bounds
  of Proposition~\ref{pro:lim_cut_general_weight_r_graph}.
  We do not give the details here.
  \begin{corollary}[Unweighted $r$-graph] \label{cor:cut_unweighted}
    For the $r$-neighborhood graph and the weight function $f_n = 1$ we obtain
    \begin{align*}
      \left| \E \left( \frac{\cut_n}{n^2 r_n^{d+1}} \right) 
      - \frac{2 \eta_{d-1}}{d+1} \int_{S} p^2 (s)  \ \mathrm{d}s \right|  = O ( r_n ) .
    \end{align*}
    and, for a suitable constant $\tilde{C}>0$,
    \begin{align*}
      &\Pr \left( \left| \frac{\cut_{n}}{n^2 r_n^{d+1}} 
      - \E \left( \frac{\cut_{n}}{n^2 r_n^{d+1}} \right) \right| \geq \eps \right)
      \leq 2 \exp \left(- \tilde{C} n  r_n^{d+1} \eps^2 \right) .
    \end{align*}
  \end{corollary}

  \begin{corollary}[Complete weighted graph] 
    \label{cor:cut_complete_graph}
    Consider the complete weighted graph $G_n$ with Gaussian weight function.
    Then we have for the bias term for any $\alpha \in (0,1)$
    \begin{align*}
      \left| \E \left( \frac{\cut_{n}}{n^2 \sigma_n} \right) 
      - \frac{2}{\sqrt{2 \pi}} \int_S p^2 (s) \ \mathrm{d} s \right|     
      = O \left( \sigma_n^\alpha \right) .
    \end{align*}
    For the variance term we can find a constant $\tilde{C}>0$ 
    such that for $n$ sufficiently large
    \begin{align*}
      \Pr \left( \left| \frac{\cut_{n}}{n^2 \sigma_n} 
      - \E \left( \frac{\cut_{n}}{n^2 \sigma_n} \right) \right| \geq \eps \right)
      &\leq 2 \exp \left(- \tilde{C} n \sigma_n^{d+1} \eps^2 \right) .
    \end{align*}  
  \end{corollary}
  
  \begin{proof}{}
    Let $r_n$ be a sequence with $r_n \to 0$ and $r_n/\sigma_n \to \infty$ 
    for $n \to \infty$.
    We use the bound from Proposition~\ref{pro:lim_cut_general_weight_r_graph} 
    and the fact that $\capintegral^{(1)} (r_n)/\sigma_n$ can be bounded by a constant 
    due to Lemma~\ref{pro:gaussian_weight_function} to obtain
    \begin{align*}
      \left| \E \left( \frac{\cut_{n}}{n (n-1) \sigma_n} \right) -
      2 \frac{\capintegral^{(1)} (r_n)}{\sigma_n} \int_S p^2 (s) \ \mathrm{d} s \right|
      &=  O \left( r_n + \frac{\ballintegral^{(1)} (\infty) - \ballintegral^{(1)} (r_n)}{\sigma_n} \right)\\
      &= O \left( r_n  + \frac{1}{\sigma_n} \exp \left(- \frac{r_n^2}{4 \sigma_n^2} \right) \right) .
    \end{align*}
   
    On the other hand, using  Lemma~\ref{pro:gaussian_weight_function},
    the boundedness of $p$ and $\lebesguevol_{d-1} (S \cap C)$,
    we have for $r_n / \sigma_n$ sufficiently large
    \begin{align*}
      &\left|2 \frac{\capintegral^{(1)} (r_n)}{\sigma_n} \int_S p^2 (s) \ \mathrm{d} s 
      - \frac{2}{\sqrt{2 \pi}} \int_S p^2 (s) \ \mathrm{d} s \right|
      \leq \left| \frac{\capintegral^{(1)} (r_n)}{\sigma_n} - \frac{1}{\sqrt{2 \pi}} \right| 
      2 \int_S p^2 (s) \ \mathrm{d} s 
      = O \left( \exp \left( - \frac{r_n^2}{4 \sigma_n^2} \right) \right) .
    \end{align*}

    Combining these two bounds und using $\log \sigma_n \leq 0$ for 
    $n$ sufficiently large we obtain
    \begin{align*}
      &\left| \E \left( \frac{\cut_{n}}{n (n-1) \sigma_n} \right) 
      - \frac{2}{\sqrt{2 \pi}} \int_S p^2 (s) \ \mathrm{d} s \right|
      = O \left( r_n + \exp \left( - \frac{r_n^2}{4 \sigma_n^2} \right) \right) . 
    \end{align*}
    
    Setting $r_n=\sigma_n^\alpha$ we have to show that the exponential term 
    converges as fast. We have
    \begin{align*}
      \sigma_n^{-\alpha} \exp \left( - \frac{r_n^2}{4 \sigma_n^2} \right)
      = \sigma_n^{-\alpha} \exp \left( - \frac{1}{4} \sigma_n^{2\alpha-2} \right)
      = \left(\sigma_n^{2\alpha-2} \right)^{\frac{-\alpha}{2\alpha-2}}
      \exp \left( - \frac{1}{4} \sigma_n^{2\alpha-2} \right)
      \to 0
    \end{align*}
    for $n \to \infty$, since $x^r \exp(-x) \to 0$ for $x \to \infty$ and all $r \in \R$.

    For the variance term we have with 
    Proposition~\ref{pro:lim_cut_general_weight_r_graph}
    and for constants $\tilde{C}_1,\tilde{C}_2$
    \begin{align*}
      &\Pr \left( \left| \frac{\cut_{n}}{n (n-1) \sigma_n} 
      - \E \left( \frac{\cut_{n}}{n (n-1) \sigma_n} \right) \right| \geq \eps \right) \\
      &\qquad = \Pr \left( \left| \frac{\cut_{n}}{n (n-1) \capintegral^{(1)} (r_n)} 
      - \E \left( \frac{\cut_{n}}{n (n-1) \capintegral^{(1)} (r_n)} \right) \right| \geq \frac{\sigma_n}{\capintegral^{(1)} (r_n)} \eps \right) \\
      &\qquad \leq 2 \exp \left(- \frac{ n \sigma_n^{2} \eps^2}{\tilde{C}_1 \capintegral^{(2)} (r_n)
	+ \tilde{C}_2 (\ballintegral^{(2)} (\infty) - \ballintegral^{(2)} (r_n) )  + 2 \eps  \capintegral^{(1)} (r_n) \ f_n (0)} \right) .
    \end{align*}
    
    With Lemma~\ref{pro:gaussian_weight_function} we have for $r_n / \sigma_n$
    sufficiently large $\capintegral^{(2)} (r_n) = O (\sigma_n^{1-d} )$, 
    and
    \begin{align*}
      \ballintegral^{(2)} (\infty) - \ballintegral^{(2)} (r_n)
      = O \left( \sigma_n^{-d} \exp \left(-\frac{r_n^2}{4 \sigma_n^2} \right) \right) 
      = O \left( \sigma_n^{1-d} \right) ,
    \end{align*}
    if we choose $r_n =\sigma_n^\alpha$ for $\alpha \in (0,1)$ similarly to above.

    For the last term in the denominator we have
    $\capintegral^{(1)} (r_n) \ f_n (0) = O \left( \sigma_n \sigma_n^{-d} \right)
    = O \left( \sigma_n^{1-d} \right)$.
    Therefore, we can find a constant $\tilde{C}_3>0$ such that
    \begin{align*}
      \Pr \left( \left| \frac{\cut_{n}}{n (n-1) \sigma_n} 
      - \E \left( \frac{\cut_{n}}{n (n-1) \sigma_n} \right) \right| \geq \eps \right)
      &\leq 2 \exp \left(- \tilde{C}_3 \frac{n \sigma_n^{2} \eps^2}{\sigma_n^{1-d}} \right)
      = 2 \exp \left(- \tilde{C}_3 n \sigma_n^{d+1} \eps^2 \right) .
    \end{align*}  

    Since we assume that $n \sigma_n \to \infty$ for $n \to \infty$ 
    we can replace $n (n-1)$ in the scaling factor by $n^2$.
  \end{proof}

  We do not state the proof of the following corollary, since it is similar to 
  the proof of the last one. The difference is, that we do not have to consider
  the $\indicator_c$-terms, which are zero in the case of the $r$-graph.
  \begin{corollary}[$r$-graph with Gaussian weights and $\sigma_n/r_n \to 0$] 
    \label{cor:cut_r_graph_sigma_to_r_to_zero}
    Let $G_n$ be the $r$-graph with Gaussian weight function
    and let $\sigma_n / r_n \to 0$ for $n \to \infty$. 
    Then we have for the bias term 
    \begin{align*}
      \left| \E \left( \frac{\cut_{n}}{n^2 \sigma_n} \right) 
      - \frac{2}{\sqrt{2 \pi}} \int_S p^2 (s) \ \mathrm{d} s \right|     
      = O \left( r_n + \exp \left( - \frac{r_n^2}{4 \sigma_n^2} \right) \right) .
    \end{align*}
    For the variance term we can find a constant $\tilde{C}_2>0$ 
    such that
    \begin{align*}
      \Pr \left( \left| \frac{\cut_{n}}{n^2 \sigma_n} 
      - \E \left( \frac{\cut_{n}}{n^2 \sigma_n} \right) \right| \geq \eps \right)
      &\leq 2 \exp \left(- \tilde{C}_2 \ n \ \sigma_n^{d+1} \eps^2 \right) .
    \end{align*}  
  \end{corollary}

  \begin{corollary}[$r$-graph with Gaussian weights and $r_n / \sigma_n \to 0$]
    \label{cor:cut_r_graph_gaussian_weights_r_to_sigma_to_zero}
    Consider the $r$-neighborhood graph with Gaussian weight function and 
    let $r_n / \sigma_n \to 0$ for $n \to \infty$.
    Then we can find a constant $\tilde{C}>0$ such that
    \begin{align*}
      &\left| \E \left( \frac{\sigma_n^d}{r_n^{d+1}} \frac{\cut_{n}}{n^2} \right) 
      - \frac{2 \eta_{d-1}}{(d+1) (2 \pi)^{d/2}} \int_S p^2 (s) \ \mathrm{d} s \right|
      = O \left( r_n + \frac{r_n^2}{\sigma_n^2} \right) . 
    \end{align*}
    and 
    \begin{align*}
      \Pr \left( \left| \frac{\sigma_n^d}{r_n^{d+1}} \frac{\cut_{n}}{n^2} 
      - \E \left( \frac{\sigma_n^d}{r_n^{d+1}} \frac{\cut_{n}}{n^2} \right) \right| \geq \eps \right) 
      &\leq 2 \exp \left(- \tilde{C}  n \eps^2 r_n^{d+1} \right) .
    \end{align*}
  \end{corollary}

  \begin{proof}{}
    Multiplying the bound in Proposition~\ref{pro:lim_cut_general_weight_r_graph} 
    with $\sigma_n^d \capintegral^{(1)} (r_n)/r_n^{d+1}$, which can be bounded by a constant 
    according to Lemma~\ref{pro:gaussian_weights_r_to_sigma_to_zero},
    and using $\indicator_{c}=0$ we obtain
    \begin{align*}
      &\left| \E \left( \frac{\sigma_n^d \capintegral^{(1)} (r_n)}{r_n^{d+1}} \frac{\cut_{n}}{n (n-1)} \right) -
      2 \frac{\sigma_n^d \capintegral^{(1)} (r_n)}{r_n^{d+1}} \int_S p^2 (s) \ \mathrm{d} s \right|
      = O \left( r_n \right) .
    \end{align*}

    On the other hand, by the boundedness of $p$ and $\lebesguevol_{d-1} (S \cap C)$,
    and with Lemma~\ref{pro:gaussian_weights_r_to_sigma_to_zero}
    \begin{align*}
      &\left|2 \frac{\sigma_n^d \capintegral^{(1)} (r_n)}{r_n^{d+1}} \int_S p^2 (s) \ \mathrm{d} s 
      -  \frac{2 \eta_{d-1}}{(d+1) (2 \pi)^{d/2}} \int_S p^2 (s) \ \mathrm{d} s \right| 
      = O \left( \frac{r_n^2}{\sigma_n^2} \right) .
    \end{align*}
    Combining these two bounds we obtain the result for the bias term.

    For the variance term we have with Proposition~\ref{pro:lim_cut_general_weight_r_graph}
    and for a constant $\tilde{C_1}$
    \begin{align*}
      &\Pr \left( \left| \frac{\sigma_n^d}{r_n^{d+1}} \frac{\cut_{n}}{n (n-1)} 
      - \E \left( \frac{\sigma_n^d}{r_n^{d+1}} \frac{\cut_{n}}{n (n-1)} \right) \right| \geq \eps \right) \\
      &\qquad = \Pr \left( \left| \frac{\cut_{n}}{n (n-1) \capintegral^{(1)} (r_n)} 
      - \E \left( \frac{\cut_{n}}{n (n-1) \capintegral^{(1)} (r_n)} \right) \right| \geq \frac{r_n^{d+1}}{\sigma_n^d \capintegral^{(1)} (r_n)} \eps \right) \\
      &\qquad \leq 2 \exp \left(- \frac{ n \left( r_n^{d+1}/\sigma_n^{d} \right)^2 \eps^2}
      {\tilde{C}_1 \capintegral^{(2)} (r_n) \ + 2 \eps  \capintegral^{(1)} (r_n) \ f_n (0)} \right) .
    \end{align*}
    
    With Lemma~\ref{pro:gaussian_weights_r_to_sigma_to_zero}
    we obtain $\capintegral^{(2)} (r_n)=O ( r_n^{d+1}/\sigma_n^{2d})$ for sufficiently
    large $n$. 
    With the same proposition and plugging in $f_n (0)$ we obtain
    $\capintegral^{(1)} (r_n) f_n (0) = O ( r_n^{d+1}/\sigma_n^{2d} )$.
    Plugging in these results above we obtain the bound for the variance term.

    Since we always assume that $n r_n \to \infty$ for $n \to \infty$ we can 
    replace $n (n-1)$ in the scaling factor by $n^2$.
  \end{proof}

  \subsubsection{The volume term in the $r$-graph and the complete weighted graph}
  \label{sec:volume_r_graph_complete_graph}

  The following results are stated without proof: 
  Proposition~\ref{pro:volume_r_graph_general} can be proved analogously 
  to Proposition~\ref{pro:volume_knn_general} if the remarks 
  on the difference between the $\kNN$-graph and $r$-neighborhood graph
  in the proof of Proposition~\ref{pro:lim_cut_general_weight_r_graph}
  are considered.
  The corollaries can be shown similarly to the corresponding 
  corollaries in the previous section.
  
  \begin{proposition} \label{pro:volume_r_graph_general}
    Let $G_n$ be the $r_n$-neighborhood graph or the complete weighted graph with 
    a weight function $f_n$ and set $\indicator_{c}$ as in 
    Proposition~\ref{pro:lim_cut_general_weight_r_graph}.
    Then
    \begin{align*}
      \left| \E \left( \frac{\vol_n (H)}{n (n-1) \ballintegral^{(1)} (r_n)} \right) - 
      \int_{H}  p^2 (x) \ \mathrm{d} x \right| 
      \leq O \left( r_n + \frac{\ballintegral^{(1)} (\infty) - \ballintegral^{(1)} (r_n)}{\ballintegral^{(1)} (r_n)}  \indicator_{c} \right) .
    \end{align*}
    For the variance term we have 
    \begin{multline*}
      \Pr \left( \left| \frac{\vol_n (H)}{n (n-1) \ballintegral^{(1)} (r_n)} 
      - \E \left( \frac{\vol_n (H)}{n (n-1)  \ballintegral^{(1)} (r_n)} \right) \right| \geq \eps \right) \\
      \leq 2 \exp \left( - \frac{ n \eps^2 \left( \ballintegral^{(1)} (r_n) \right)^2}
           {\tilde{C_1}  \ballintegral^{(2)} (r_n) + \tilde{C}_2 \indicator_{c} (\ballintegral^{(2)} (\infty) - \ballintegral^{(2)} (r_n) ) + 
             2 \eps f_n (0) \ballintegral^{(1)} (r_n)} \right) .
    \end{multline*}
  \end{proposition}

  \begin{corollary}[Unweighted graph] \label{cor:volume_unweighted}
    For $f_n \equiv 1$ and the $r_n$-neighborhood graph we have 
    \begin{align*}
      \left| \E \left( \frac{\vol_n (H)}{n^2 r_n^d} \right) - 
      \eta_d \int_{H \cap C}  p^2 (x) \ \mathrm{d} x \right| 
      \leq O (r_n) 
    \end{align*}
    and, for a constant $\tilde{C}>0$,
    \begin{align*}
      \Pr \left( \left| \frac{\vol_n (H)}{n^2 r_n^d} 
      - \E \left( \frac{\vol_n (H)}{n^2 r_n^d} \right) \right| \geq \eps \right) 
      \leq 2 \exp \left( - \tilde{C} n \eps^2 r_n^d \right) .
    \end{align*}
  \end{corollary}

  \begin{corollary}[Complete weighted graph with Gaussian weights]
    \label{cor:volume_complete_graph}
    Consider the complete weighted graph with the Gaussian weight function
    and a parameter sequence $\sigma_n \to 0$. 
    Then we have for any $\alpha \in (0,1)$ 
    \begin{align*}
      &\left| \E \left( \frac{\vol_n (H)}{n^2} \right) - 
      \int_{H}  p^2 (x) \ \mathrm{d} x \right|
      = O \left( \sigma_n^\alpha \right) .
    \end{align*}
    Furthermore there is a constant $\tilde{C}'>0$ such that 
    \begin{align*}
      &\Pr \left( \left| \frac{\vol_n (H)}{n^2} 
      - \E \left( \frac{\vol_n (H)}{n^2} \right) \right| \geq \eps \right) 
      \leq \exp \left( - \tilde{C}' n \eps^2 \sigma_n^{d} \right)
    \end{align*}
  \end{corollary}

  \begin{corollary}[$r$-graph with Gaussian weights and $\sigma_n/r_n \to 0$]
    \label{cor:volume_r_graph_sigma_to_r_to_zero}
    Let $G_n$ be the $r$-neighborhood graph with Gaussian weights and
    let $\sigma_n/r_n \to 0$ for $n \to \infty$. 
    Then we have for the bias term for sufficiently large $n$
    \begin{align*}
      &\left| \E \left( \frac{\vol_n (H)}{n^2} \right) - 
      \int_{H}  p^2 (x) \ \mathrm{d} x \right| 
      = O \left( r_n + \exp \left(- \frac{1}{4} 
      \frac{r_n^2}{\sigma_n^2} \right) \right) .
    \end{align*}
    and for the variance term for a suitable constant $\tilde{C}'>0$
    \begin{align*}
      &\Pr \left( \left| \frac{\vol_n (H)}{n^2} 
      - \E \left( \frac{\vol_n (H)}{n^2} \right) \right| \geq \eps \right) 
      \leq \exp \left( - \tilde{C}' n \eps^2 \sigma_n^{d} \right) .
    \end{align*}
  \end{corollary}

  \begin{corollary}[$r$-graph with Gaussian weights and $r_n/\sigma_n \to 0$]
    \label{cor:volume_r_graph_r_to_sigma_to_zero}
    Let $G_n$ be the $r$-neighborhood graph with Gaussian weights and     
    let $r_n/\sigma_n \to 0$ for $n \to \infty$. 
    Then we have for the bias term for sufficiently large $n$
    \begin{align*}
      \left| \E \left( \frac{\sigma_n^{d}}{n^2 r_n^d} \vol_n (H) \right)
      - \frac{\eta_d}{(2 \pi)^{d/2}} 
      \int_{H} p^2 (x) \ \mathrm{d} x \right|
      = O \left( r_n + \left( \frac{r_n}{\sigma_n} \right)^2 \right) .
    \end{align*}
    and for the variance term for a suitable constant $\tilde{C}>0$
    \begin{align*}
      &\Pr \left( \left| \frac{\sigma_n^{d}}{n^2 r_n^d} \vol_n (H)
      - \E \left( \frac{\sigma_n^{d}}{n^2 r_n^d} \vol_n (H) \right) \right| > \eps \right) 
      \leq 2 \exp \left(- \tilde{C} n \eps^2 r_n^d \right) .
    \end{align*}
  \end{corollary}

  \subsubsection{The main theorems for the $r$-graph and the complete weighted graph}
  \label{sec:proofs_main_theorems_r_graph}
  
  \begin{proof}{ of Theorem~\ref{thm:limit_ncut_cheegercut_r_graph}}
    As discussed in Section~\ref{sec:proofs_intuitively} we can study the
    convergence of the bias and variance terms of the $\cut$ and the volume 
    separately.

    For the {\em unweighted $r$-graph} we have with Corollary~\ref{cor:cut_unweighted} 
    that the bias term of the $\cut$ is in $O(r_n)$ and that for $\eps>0$ we can find
    a constant $\tilde{C}$ such that the probability that the variance term of the
    $\cut$ exceeds $\eps$ is bounded by $2 \exp ( -\tilde{C} n \sigma_n^{d+1} \eps^2)$.
    Thus the $\cut$-term converges almost surely to zero 
    for $r_n \to 0$ and $n r_n^{d+1} / \log n \to \infty$.
    It follows from Corollary~\ref{cor:volume_unweighted} that under these 
    conditions the $\vol$-term also converges to zero.
    The best convergence rate for the $\cut$-term is $\sqrt[d+3]{\log n / n}$, 
    which is achieved setting $r_n \sim \sqrt[d+3]{\log n / n}$.
    Setting $r_n$ in this way the convergence rate of the $\vol$-term is also
    $\sqrt[d+3]{\log n / n}$.

    For the $r$-graph with {\em Gaussian weights and $r_n/\sigma_n \to \infty$} 
    we have with Corollaries~\ref{cor:cut_r_graph_sigma_to_r_to_zero}
    and~\ref{cor:volume_r_graph_sigma_to_r_to_zero}
    that the bias term of both, the $\cut$ and the volume, is in
    $O ( r_n + \exp (-1/4 (r_n/\sigma_n)^2) )$. 
    Furthermore, we can find a constant $\tilde{C}>0$ such that the probability 
    that the variance term of the $\cut$ exceeds an $\eps>0$ is bounded by 
    $2 \exp ( -\tilde{C} n \sigma_n^{d+1} \eps^2)$.
    Similarly, the variance term of the volume would converge almost surely for
    $n \sigma_n^{d} / \log n \to \infty$.
    This implies almost sure convergence of $\Delta_n$ to zero under the condition
    $n \sigma_n^{d+1} / \log n \to \infty$ for $n \to \infty$.

    For the $r$-graph with {\em Gaussian weights and $r_n/\sigma_n \to 0$} 
    we have with Corollary~\ref{cor:cut_r_graph_gaussian_weights_r_to_sigma_to_zero}
    a rate of $O ( r_n + ( r_n / \sigma_n )^2 )$ for the bias term of the $\cut$. 
    Furthermore, the probability that the variance term 
    exceeds an $\eps>0$ is bounded by $2 \exp (- \tilde{C} n \eps^2 r_n^{d+1} )$
    with a constant $\tilde{C}$.
    Therefore, the $\cut$-term almost surely converges to zero under the conditions
    $r_n \to 0$ and $n r_n^{d+1} / \log n \to \infty$.
    Under these conditions with Corollary~\ref{cor:volume_r_graph_r_to_sigma_to_zero}
    the volume-term also converges to zero.
  \end{proof}

  \begin{proof}{ of Theorem~\ref{thm:limit_ncut_cheegercut_complete_graph}}
    As discussed in Section~\ref{sec:proofs_intuitively} we can study the
    convergence of the bias and variance terms of the $\cut$ and the volume 
    separately.
    
    With Corollaries~\ref{cor:cut_complete_graph} and~\ref{cor:volume_complete_graph}
    we have that the bias term of both, the $\cut$ and the volume is in 
    $O ( \sigma_n^\alpha )$ for any $\alpha \in (0,1)$.
    Furthermore, the probability that the variance term of the $\cut$ exceeds an $\eps>0$ 
    is bounded by $2 \exp ( - \tilde{C} \ n \ \sigma_n^{d+1} \eps^2 )$ 
    with a suitable constant $\tilde{C}$. For the variance term of the volume
    the exponent in this bound is only $d$.
    Consequently, we have almost sure convergence to zero under the condition 
    $n \sigma_n^{d+1} / \log n \to \infty$. 

    For any fixed $\alpha \in (0,1)$ the optimal convergence rate is achieved
    setting $\sigma_n=((\log n)/n)^{1/(d+1+2\alpha)}$. 
    Since the variance term has to converge for any $\alpha \in (0,1)$ we choose
    $\sigma_n=((\log n)/n)^{1/(d+3)}$ and achieve a convergence rate
    of $\sigma_n^\alpha$ for any $\alpha \in (0,1)$.
  \end{proof}

  \subsection{The integrals $\capintegral^{(q)} (r)$ and the size of the boundary strips}
  \label{sec:gaussian_weight_function}

  \begin{lemma}[Unit weights] 
    \label{pro:integration_unit_weight_function}
    Let $f_n \equiv 1$ be the unit weight function. 
    Then for any $r>0$
    \begin{align*}
      \capintegral^{(1)} (r) = \capintegral^{(2)} (r) = \frac{\eta_{d-1}}{d+1} r^{d+1} 
      \intertext{and}
      \ballintegral^{(1)} (r) = \ballintegral^{(2)} (r) = \eta_d r^{d} . 
    \end{align*}
  \end{lemma}

  \begin{lemma}[Gaussian weights and $r_n / \sigma_n \to 0$]
    \label{pro:gaussian_weights_r_to_sigma_to_zero}
    Let $f_n$ denote the Gaussian weight function with parameter $\sigma_n$
    and let $r_n>0$.
    Then we have for $q=1,2$ for the cap integral
    \begin{align*}
      \left|\frac{\sigma_n^{qd}}{r_n^{d+1}} \capintegral^{(q)} (r_n) 
      - \frac{\eta_{d-1}}{(d+1) (2 \pi)^{qd/2}} \right|
      \leq 2 \left( \frac{r_n}{\sigma_n} \right)^2      
    \end{align*}
    For the ball integral $\ballintegral^{(q)} (r_n)$ we have
    \begin{align*}
      \left| \frac{\sigma_n^{qd}}{r_n^d} \ballintegral^{(q)} (r_n)  
      - \frac{\eta_d}{(2 \pi)^{qd/2}} \right|
      \leq 3 \left( \frac{r_n}{\sigma_n} \right)^2 .
    \end{align*}
  \end{lemma}

  \begin{lemma}[Gaussian weights and $\sigma_n / r_n \to 0$]
    \label{pro:gaussian_weight_function}
    Let $f_n$ denote the Gaussian weight function with a parameter $\sigma_n$
    and let $r_n / \sigma_n \geq 4 d$. 
    Then we have $\capintegral^{(1)} (\infty)=\sigma_n/\sqrt{2 \pi}$ and
    \begin{align*}
      &\left| \frac{1}{\sigma_n} \capintegral^{(1)} (r_n) - \frac{1}{\sqrt{2 \pi}} \right|
      = O\left( \exp \left( - \frac{1}{4} \left( \frac{r_n}{\sigma_n} \right)^2 \right) \right)
    \end{align*}
    Furthermore, $\capintegral^{(2)} (\infty)=O(\sigma_n^{1-d})$ and 
    $\capintegral^{(2)} (\infty)-\capintegral^{(2)}(r_n) =O(\sigma_n^{1-d} \exp \left( - (r_n/\sigma_n)^2 /4\right) )$.  

    For the ball integral we have under the same conditions
    $\ballintegral^{(1)} (\infty)=1$ 
    \begin{align*}
      \left| \ballintegral^{(1)} (r_n) - 1 \right|
      =O\left( \exp \left( - \frac{1}{4} \left( \frac{r_n}{\sigma_n} \right)^2 \right) \right) .
    \end{align*}
    Furthermore, $\ballintegral^{(2)} (\infty) = O (\sigma_n^{-d})$ and 
    $\ballintegral^{(2)} (\infty)-\ballintegral^{(2)}(r_n)=O(\sigma_n^{-d} \exp \left( - (r_n/\sigma_n)^2 /4\right) )$. 
  \end{lemma}

  The following lemma is necessary to bound the influence of points close to 
  the boundary on the $\cut$ and the volume. The first statement 
  is used for the $\cut$, whereas the second statement is used for the volume.
  \begin{lemma} \label{lem:boundary_strip_to_zero}
    Let the general assumptions hold and let $(r_n)_{n \in \N}$ be a sequence with
    $r_n \to 0$ for $n \to \infty$. 
    Define $\mathcal{R}_n=\{x \in \R^d \; | \; \dist (x,\partial C) \leq 2 r_n \}$.
    Then
    $\lebesguevol_{d-1} ( S \cap \mathcal{R}_n ) = O (r_n)$.

    For $H=H^+$ or $H=H^-$ define $\bar{\mathcal{R}}_n = \{ x \in H \cap C \; | \; \dist(x,\partial(H \cap C)) \leq 2 r_n \}$.
    Then $\lebesguevol_d ( \bar{\mathcal{R}}_{n} ) = O (r_n)$.
  \end{lemma}

\end{document}